\documentclass[acmlarge]{acmart}
\pdfoutput=1%For ARXIV
\usepackage{graphicx}
\usepackage{subcaption}
\usepackage{epstopdf}
\usepackage{tabularx}
\usepackage{multirow} 
\usepackage{bigstrut}
\usepackage{makecell}
\usepackage{txfonts}
\UseRawInputEncoding
\usepackage{balance}
\usepackage[T1]{fontenc}
\usepackage{lmodern}
\usepackage{array}
\usepackage[normalem]{ulem}

\usepackage{bm}
\begin{document}
	
	\title{Survey on Evolutionary Deep Learning: Principles, Algorithms, Applications and Open Issues}
	
	
	\author{Nan Li}
	\affiliation{
		\institution{Northeastern University}
		\streetaddress{No.195, Chuangxin Road}
		\city{Shenyang}
		\state{Liaoning Province}
		\country{China}}
	\email{2010500@stu.neu.edu.cn}
	%Corresponding authors
	\author{Lianbo Ma}
	\authornote{Corresponding Authors:Lianbo Ma and Guo Yu}
	\affiliation{%
		\institution{Northeastern University}
		\streetaddress{No.195, Chuangxin Road}
		\city{Shenyang City}
		\state{Liaoning Province}
		\country{China}}
	\email{malb@swc.neu.edu.cn}
	
	\author{Guo Yu}
	\authornotemark[1]
	\affiliation{%
		\institution{East China University of Science and Technology}
		\streetaddress{Meilong Road 130}
		%\city{Xuhui Qu}
		\state{Shanghai}
		\country{China}}
	\email{guoyu@ecust.edu.cn}
	
	\author{Bing Xue}
	\affiliation{%
		\institution{Victoria University of Wellington}
		\city{Wellington}
		\country{New Zealand}}
	\email{bing.xue@ecs.vuw.ac.nz}

	\author{Mengjie Zhang}
	\affiliation{%
		\institution{Victoria University of Wellington}
		\city{Wellington}
		\country{New Zealand}}
	\email{mengjie.zhang@ecs.vuw.ac.nz}
	
	\author{Yaochu Jin}
	\affiliation{%
		\institution{Bielefeld University}
		\city{Bielefeld}
		\country{Germany}}
	\email{yaochu.jin@uni-bielefeld.de}

	\renewcommand{\shortauthors}{Nan Li et al.}
	\begin{abstract}
		Over recent years, there has been a rapid development of deep learning (DL) in both industry and academia fields. However, finding the optimal hyperparameters of a DL model often needs high computational cost and human expertise. To mitigate the above issue, evolutionary computation (EC) as a powerful heuristic search approach has shown significant merits in the automated design of DL models, so-called evolutionary deep learning (EDL). This paper aims to analyze EDL from the perspective of automated machine learning (AutoML). Specifically, we firstly illuminate EDL from machine learning and EC and regard EDL as an optimization problem. According to the DL pipeline, we systematically introduce EDL methods ranging from feature engineering, model generation, to model deployment with a new taxonomy (i.e., what and how to evolve/optimize), and focus on the discussions of solution representation and search paradigm in handling the optimization problem by EC. Finally, key applications, open issues and potentially promising lines of future research are suggested. This survey has reviewed recent developments of EDL and offers insightful guidelines for the development of EDL.
	\end{abstract}
	
	
	
	\maketitle
	
	\section{Introduction}
	
\begin{figure}[h]
		\centering
		\includegraphics[width=0.9\textwidth]{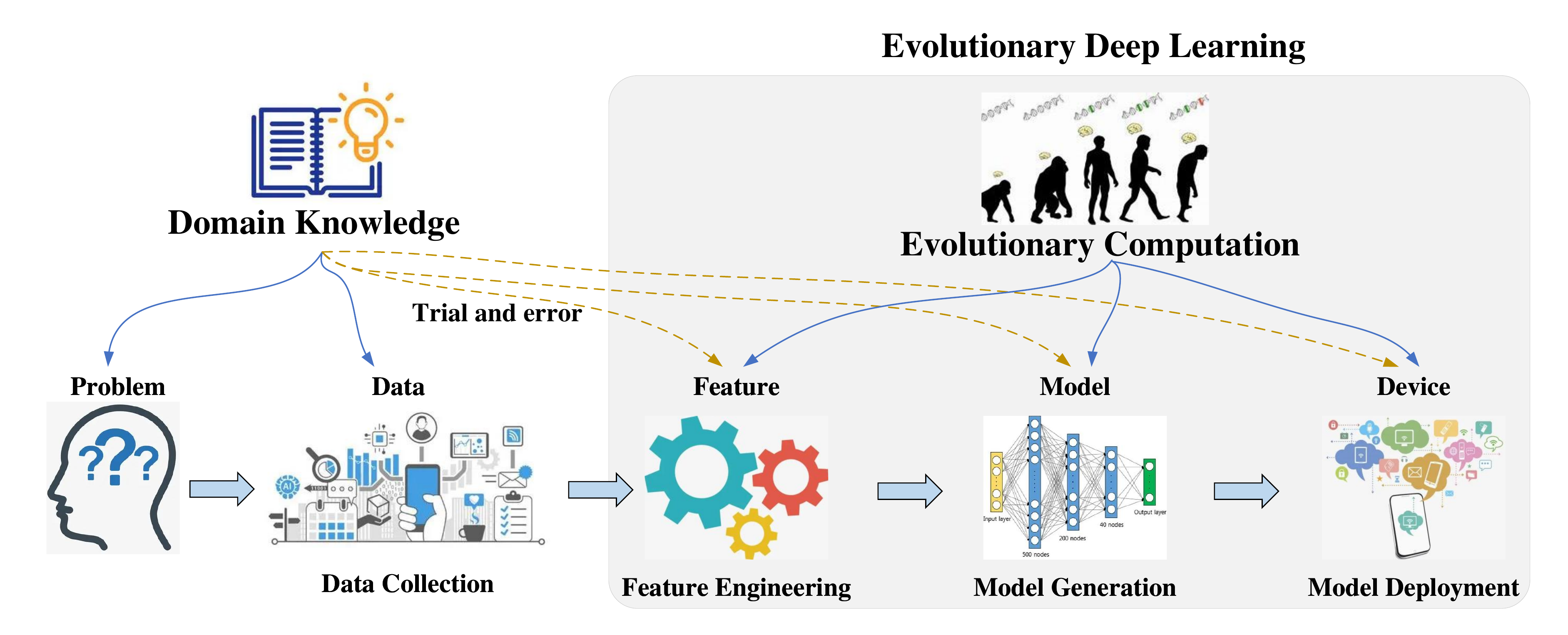}
		\caption{ An overview of DL, driven by domain knowledge or evolutionary computation, where the life of DL gets through problem, data collection, feature engineering, model generation and model deployment. }
		\label{fig1}
	\end{figure}

	Deep learning (DL) as a promising technology has been widely used in a variety of challenging tasks, such as image analysis \cite{A001r} and pattern recognition \cite{004r}. However, the practitioners of DL struggle to manually design deep models and find appropriate configurations by trial and error. An example is given in Fig. \ref{fig1}, where domain knowledge is fed to DL in different stages like feature engineering (FE) \cite{033r}, model generation \cite{031r} and model deployment \cite{042r,043r}. Unfortunately, the difficulty in the acquisition of expert knowledge makes DL undergo a great challenge in its development. 
	
	In contrast, the automatic design of deep neural networks (DNNs) tends to be prevalent in recent decades \cite{025r, 031r}. The main reason lies in the flexibility and computation efficiency of automated machine learning (AutoML) in FE \cite{033r}, parameter optimization (PO) \cite{531r}, hyperparameter optimization (HPO) \cite{035r}, neural architecture search (NAS) \cite{024r,025r,031r}, and model compression (MC) \cite{759r}. In this way, AutoML without manual intervention has attracted great attention and much progress has been made. 
	
\begin{figure}[h]
		\centering
		\includegraphics[width=0.9\textwidth]{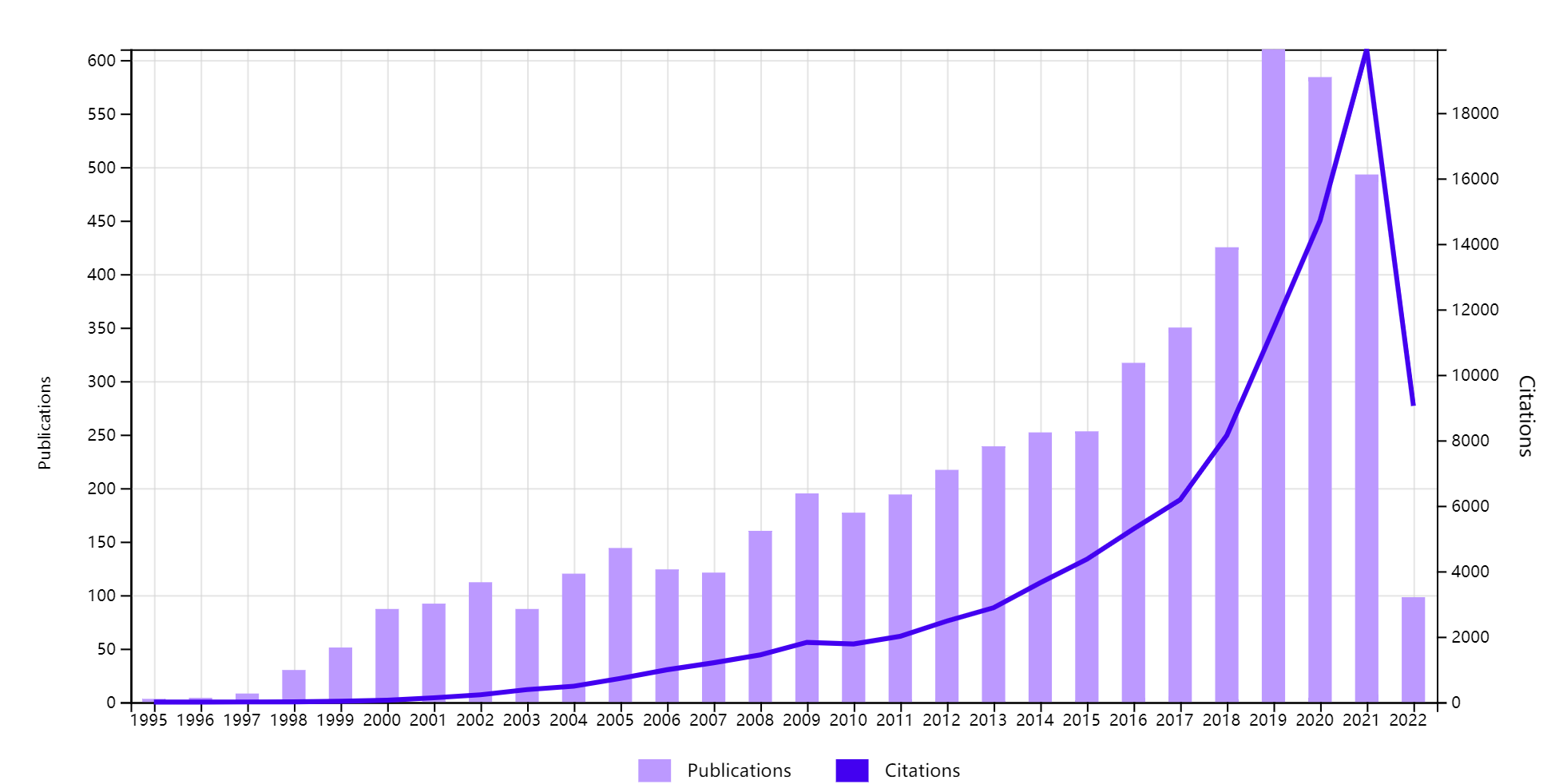}
		\caption{Total publications and citations referring to EC \& DL by years from Web of Science until July 2022.}
		\label{fig2}
	\end{figure}

	Evolutionary computation (EC) has been widely applied to automatic DL, owing to its flexibility and automatically evolving mechanism. In EC, a population of individuals are driven by the environmental selection to evolve towards the optimal solutions or front \cite{040r}. Nowadays, there are many automatic DL methods driven by EC, termed as evolutionary deep learning (EDL) \cite{041r, z01, z02, z03}. For example, a number of studies on EC have been carried out to the feature engineering \cite{033r}, model generation \cite{024r,031r}, and model deployments \cite{042r}, as shown in Fig. \ref{fig1}. Therefore, the integration of EC and DL has become a hot research topic in both academic and industrial communities. Moreover, in Fig. \ref{fig2}, the number of publications and citations referring to EC \& DL by years from Web of Science gradually increases until around 2012, whereas it sharply rises in the following decade. Hence, more and more researchers work on the area of EDL.

	\begin{table}[h]\scriptsize
		\centering
		\caption{Comparison between existing surveys and our work, where FE, PO, HPO, NAS, and MC indicate feature engineering, parameter optimization, hyperparameter optimization, neural architecture search, and model compression, respectively. ``\checkmark'' and ``-'' indicate the content is included or not in the paper, respectively.}
		\begin{tabular}{p{4.042em}p{4.042em}p{4.042em}p{4.042em}p{4.042em}p{4.042em}p{4.042em}}
			\hline
			Survey & Type  & FE    & PO    & HPO   & NAS   & MC \bigstrut\\
			\hline
			\cite{024r}  & AutoML & \checkmark      & -     & \checkmark     & \checkmark     & - \bigstrut[t]\\
			\cite{025r}  & AutoML & \checkmark     & -     & \checkmark     & \checkmark     & - \\
			\cite{049r}  & NAS   &-     & \checkmark     & \checkmark     & \checkmark     & - \\
			\cite{045r}  & NAS   & -     & -     & \checkmark     & \checkmark     & - \\
			\cite{468r}  & NAS   & -     & -     & \checkmark     & \checkmark     & - \\
			\cite{006r}  & NAS   & -     & -     & \checkmark     & \checkmark     & - \\
			\cite{041r}  & EDL   & -     & -     & -     & \checkmark     & - \\
			\cite{037r}  & EDL   & -     & \checkmark     & \checkmark     & \checkmark     & - \\
			\cite{031r}  & EDL   & -     & \checkmark     & \checkmark     & \checkmark     & - \\
			\cite{033r}  & EDL   & \checkmark     & -     & -     & -     & - \\
			\cite{044r}  & EDL   & \checkmark     & \checkmark     & -     & -     & - \\
			\cite{046r}  & EDL   & -     & \checkmark     & \checkmark     & \checkmark     & - \\
			\cite{047r}  & EDL   & \checkmark     & -     & \checkmark     & \checkmark     & - \\
			\cite{048r}  & EDL   & \checkmark     & \checkmark     & \checkmark     & \checkmark     & - \\
			\cite{050r}  & EDL   & -     & \checkmark     & \checkmark     &   \checkmark   & - \\
			\cite{051r}  & EDL   & -     & \checkmark     & \checkmark     &   \checkmark   & - \\
			Ours  & EDL   & \checkmark     & \checkmark     & \checkmark     & \checkmark     & \checkmark \bigstrut[b]\\
			\hline
		\end{tabular}%
		\label{table1}%
	\end{table}%

	In Table \ref{table1}, we have listed recent surveys on automatic DL. A large number of surveies concentrate on the optimization of DL models \cite{025r,041r, 031r,033r}, or NAS \cite{037r,049r}. Many others focus on specific optimization paradigms such as reinforcement learning (RL)  \cite{468r}, EC \cite{036r} and gradient \cite{006r}. However, very few of them have systematically analysed EDL and runs the gamut of FE, PO, HPO, NAS, and MC. To fill the gap, we aim to give a comprehensive review of EDL in detail. The main contributions of this work are as follows.

	\begin{itemize}
		\item Existing work on EDL is reviewed from the perspective of DL and EC to facilitate the understanding of readers from the communities of both ML and EC, and we also formulated EDL into an optimization problem from the perspective of EC. 
		
		\item The survey describes and discusses on EDL in terms of feature engineering, model generation, and model deployment from a novel taxonomy, where the solution representation and the search paradigms are emphasized and systematically discussed. To the best of our knowledge, few survey has investigated the evolutionary model deployment.
	
		\item On the basis of the comprehensive review of EDL approaches, a number of applications, open issues and trends of EDL are discussed, which will guide the development of EDL.
	\end{itemize}
	
	The rest of this paper is organized as follows. Section \ref{OEDL} presents an overview of EDL. In Section \ref{EDL-FE}, EC-driven feature engineering is presented. EC-driven model generation is discussed in Section \ref{EDL-MG}. Section \ref{EDL-MD} reviews EC-driven model compressions. After that, relevant applications, open issues and the trends of EDL are discussed in Section \ref{AOT}. Finally, a conclusion of the paper is drawn in Section \ref{Con}.

	\section{An Overview of Evolutionary Deep Learning}\label{OEDL}
	\subsection{Deep Learning} 
	DL can be described as a triplet $M$ = ($D$, $T$, $P$) \cite{024r}, where $D$ is the dataset used for the training of a deep model ($M$), and $T$ is the targeted task. $P$ indicates the performance of $M$. The aim of DL is to \textit{boost its performance over specific task $T$, which is measured by $P$ on dataset $D$}. In Fig. \ref{fig1}, we can see there are three fundamental processes of DL, i.e., feature engineering, model generation and model deployment. 
	
	\textbf{Feature engineering}: It aims to find a high-quality $D$ to improve the performance ($P$) of the deep model ($M$) on specific tasks ($T$). In practice, the feature space of $D$ may include redundant and noisy information, which harms the performance ($P$) of the model ($M$).  On Prostate dataset, the size of feature subset (65) selected in \cite{z21} is only 1\% of the total size of features (10509). 
	
	\textbf{Model generation}: It targets at optimizing/generating a model ($M$) with desirable performance ($P$) for specific task ($T$) on the given datasets ($D$) \cite{025r}.  Model generation can be further divided into parameter optimization, model architecture optimization, and joint optimization \cite{031r}. Parameter optimization is to search the best parameters (e.g., weights) for a predefined model. Architecture optimization is dedicated to finding the optimal network topology (e.g., number of layers and types of operations) of a deep model ($M$) \cite{127r}. Joint optimization involves in the above two optimization issues by automatically searching for a  powerful model ($M$) on the datasets ($D$) \cite{129r}.
	
	\textbf{Model deployment}: This process aims to deploy a deep model ($M$) to solve a deployment task $T$ with acceptable performance ($P$) on input data ($D$) within limited computational budgets. The key issue of model deployment is how to reduce the latency, storage, and energy consumption when the number of parameters of a deep model is large, e.g., Transformer-XL Large has 257M parameters \cite{007r}.

	\subsection{Evolutionary Computation}
	
	EC is a collection of stochastic population-based search methods inspired by evolution mechanisms such as natural selection and genetics, which does not need gradient information and is able to handle a black-box optimization problem without explicit mathematical formulations \cite{002r, 061r}. Owing to the above characteristics, EC has been widely employed to the automatic design of DL.
	
	\begin{figure}[h]
		\centering
		\includegraphics[width=0.6\textwidth]{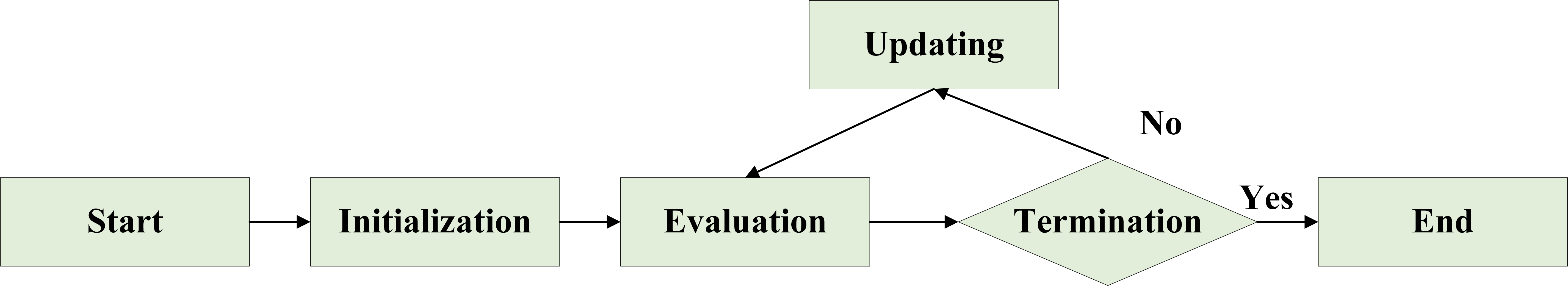}
		\caption{A general framework of EC.}
		\label{fig9}
	\end{figure}

	In principle, we can broadly divide EC methods into two categories: evolutionary algorithms (EA) and swarm intelligence (SI) \cite{031r}. Our work doesn't make an explicit distinction between EAs and SI since they comply with a general framework, as shown in Fig. \ref{fig9}, which consists of three main components.

	\begin{description}
		\item[Initialization] is performed to generate a population of individuals which are encoded according to the decision space (or search space and variable space) of the optimization problem, such as the feature set, model parameters and topological structure. 
		
		\item[Evaluation] aims to calculate the fitness of individuals. In fact, the evaluation of the individuals in EDL is a computationally expensive task \cite{077r}. For example, the work \cite{539r} used 3000 GPU days to find a desirable architecture. 
		
		\item[Updating] aims to generates a number of offspring solutions through various reproduction operations. For example, a new soultion is generated via velocity and position formula in particle swarm optimization (PSO) \cite{z21}. In terms of genetic algorithm (GA), some reproduction operators (e.g., crossover and mutation) are used to generate new individuals \cite{369r}.
	\end{description}
	%From Fig. \ref{fig9}, we can see that the optimization process involving selection, reproduction, and evaluation continues until the termination condition is satisfied. 

	\subsection{Evolutionary Deep Learning}
	
	\subsubsection{EDL from two perspectives}\label{ITP}
	\
	
	In contrast to traditional DL which heavily relies on expert or domain knowledge to build deep model, EDL is to automatically design the deep model through an evolutionary process \cite{036r,049r,045r, z03}.
	
	\textbf{From the perspective of DL}: Traditional DL needs a lot of expert knowledge in inventing and analysing a learning tool to a specific dataset or task. In contrast, EDL can be seen as a human-friendly learning tool that can automatically find appropriate deep models on given datasets or tasks \cite{024r}. In other words, \textit{EDL concentrates on how easy a learning tool can be used}.  
	
	\textbf{From the perspective of EC}: The configurations of a model is represented as an individual, and the performance as the objective to be optimized. EC plays an important role in the optimization driven by evolutionary mechanisms. Namely, \textit{EDL can be seen as an evolutionary optimization process to find the optimal configurations of the deep model with high performance}.

	From the above analysis, EDL not only aims to increase the adaptability of a deep model towards  learning tasks via the automatic construction approach (from the perspective of DL), but also tries to achieve the optimal model under the designed objectives or constraints (from the perspective of EC).

	\subsubsection{Definition and Framework of EDL}
	
	\

	According to the above discussion in Subsection \ref{ITP} and following \cite{024r}, we can define EDL as follows.
	\begin{equation}\label{Eq1}
	\begin{array}{l}
	\begin{array}{*{20}{c}}
	{\mathop {{\rm{Max}}}\limits_{config.} }&{{\textrm{Learning tools' performance,}}}
	\end{array}\\
	\begin{array}{*{20}{c}}
	{s.t.}&{\left\{ {\begin{array}{*{20}{c}}
			{{\textrm{No assistance from humans}}}\\
			{{\textrm{Limited computational budgets.}}}
			\end{array}} \right.}
	\end{array}
	\end{array}
	\end{equation}
	where $config.$ indicates the configurations which form the decision space of an optimization problem. The problem is to maximize the objective (i.e., learning tools' performance $P$) of tasks $T$ on datasets $D$ under the constraints of no assistance from humans and limited computational resources. Accordingly, three aspects are taken into account in the design of EDL.
	\begin{description}
		\item[Desirable generalization performance:] EDL should have desirable generalization performance across given datasets and tasks.
		\item[High search efficiency:] EDL is able to find optimal or desirable configuration within a limited computational budges (e.g., hardware, latency, energy consumption) under different designed objectives (e.g., high accuracy, small model size).
		\item[Without human assistance:] EDL is able to automatically configure without human intervention.
	\end{description}
	
	Following the EC framework described in Fig. \ref{fig9}, we present a general framework of EDL as follows. 
	\begin{description}
		\item [Step 1 Initialization:] A population of individuals are initialized according to the designed encoding scheme.
		\item [Step 2 Evaluation:]  Each individual is evaluated according to the objectives (e.g., high accuracy, small model size) or constraints (e.g., energy consumption).
		\item [Step 3 Updating:]  A required number of new solutions are generated from previous generation via various updating operations.
		\item [Step 4 Termination condition:] Go to Step 2 if the predefined termination condition is unsatisfied; Otherwise, go to Step 5.
		\item [Step 5 Output:] Output the solution with the best performance.
	\end{description}

	\subsubsection{Taxonomy of EDL Approaches}
	
	\ 
	
\begin{figure}[htbp]
		\centering
		\includegraphics[width=0.95\textwidth]{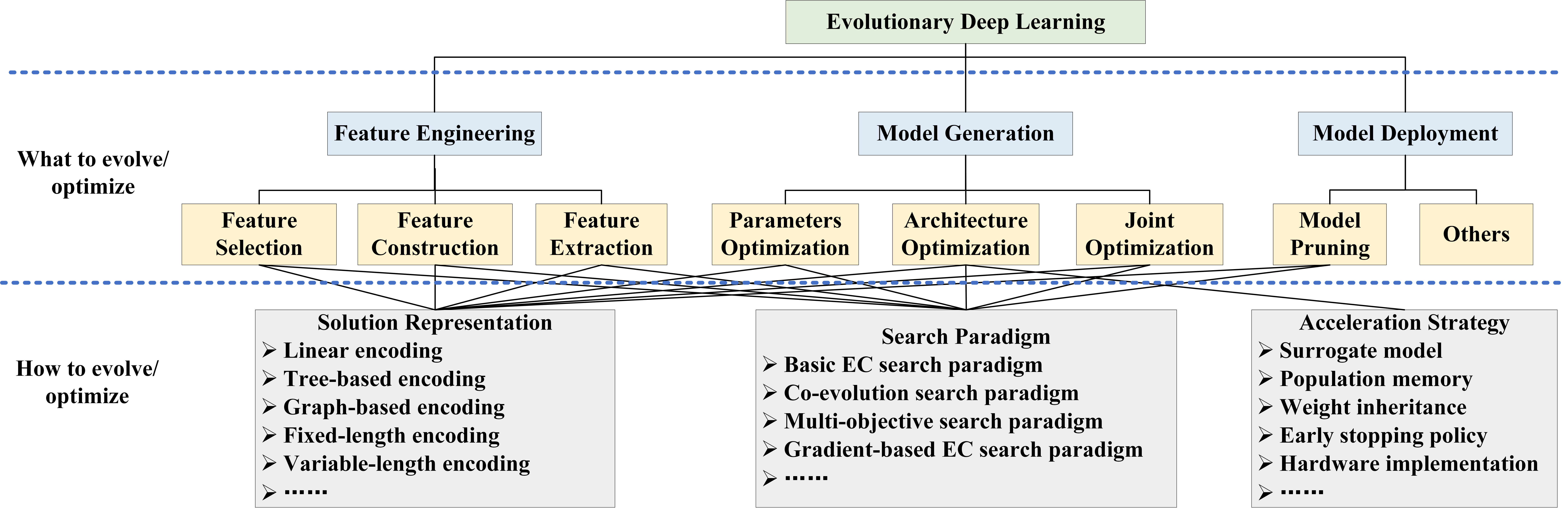}
		\caption{A taxonomy of EDL approaches.}
		\label{fig10}
	\end{figure}

	In this section, a novel taxonomy of EDL approaches is proposed according to ``what to evolve/optimize'' and ``how to evolve/optimize'', as shown in Fig. \ref{fig10}. 
	
	\textbf{``What to evolve/optimize''}: We may be concerned about ``what EDL can do'' or ``what kinds of problems EDL can tackle''. In feature engineering, there are three key issues to be resolved, including the feature selection, feature construction and feature extraction \cite{024r}. In model generation, parameter optimization, architecture optimization, and joint optimization become the critical issues \cite{031r}, while model deployment is involved with the issues of model pruning and other compression technologies.
	
	\textbf{``How to evolve/optimize''}: The answer to the question is designing appropriate solution representation and search paradigm for EC, and acceleration strategies for NAS. The representation schemes are designed for the encoding of individuals, search paradigms for the achievement of optimal configurations, acceleration strategies for the reduction of time or resources consumption. 
	
	According to the above taxonomy, we will elaborately introduce EDL in feature engineering, model generation and model deployment in Sections \ref{EDL-FE}, \ref{EDL-MG} and \ref{EDL-MD}, respectively.

	\section{Feature Engineering}\label{EDL-FE}
	
	Feature engineering is adopted to pre-process given raw data by filtering out the irrelevant features of the data or creating the new features based on original features \cite{033r}. Various EC-based techniques have been proposed to reduce data dimensionality, speed up learning process, or improve model performance \cite{033r}. The common techniques can be categorized into feature selection \cite{172r}, feature construction \cite{367r} and feature extraction \cite{434r}.

	\subsection{Feature Selection}
	\subsubsection{Problem Formulation} 
	\ 
	
	Feature selection aims to automatically select a representative subset of features where there are no irrelevant or redundant features. However, the search space grows exponentially with the increase of features. If a dataset has $n$ features, then there are $2^n$ solutions in the search space. In addition, the interactions between features may seriously impact the feature selection performance \cite{033r}. In the followings, we will review existing work on solution representations and search paradigms in EC for feature selection.

	\subsubsection{Solution Representations}
	\ 
	
	Generally, there are three different categories of solution representations.
	
	\textbf{Linear encoding}: This encoding uses vectors or strings to store feature information. For example, in \cite{165r}, a fixed-length binary vector was used to express whether a feature is selected or not, where ``1'' indicates a corresponding feature is selected, and ``0'' is the opposite. In \cite{166r}, a binary index was used to indicate the corresponding feature. 
	
	\textbf{Tree-based encoding}: In canonical genetic programming (GP), all leaf nodes/terminal nodes represent the selected features and non-terminal nodes represent functions (e.g., arithmetic or logic operators) \cite{113r}.  For automatic classification on high-dimensional data, Krawiec et al. \cite{113r} proposed a tree-based encoding to select a subset of highly discriminative features, where each feature consisted of sibling leaf nodes and their paternal function node. On the basis of the tree-based encoding, Muni et al. \cite{188r} proposed a multi-tree GP mothed for online feature selection. 
	
	\textbf{Graph-based encoding}: In \cite{190r}, the feature space of the high-dimensional data is represented by a graph and each node of the graph represents a feature. A feature subset is composed of visited nodes of the graph, i.e., the path of node composition or subgraph. Yu et al. \cite{193r} converted feature selection to the optimal path problem in a directed graph, where the value of the node was ``1'' or ``0'' to indicate whether the feature was selected or not.
	
	\subsubsection{Search Paradigms}
	\
	
	In feature selection, representative types of search paradigms are introduced as follows.
	
	\textbf{Basic EC search paradigm}: In feature selection, typical evolutionary search methods have been widely used,  such as GA \cite{194r, 165r}, GP \cite{113r, 147r}, PSO \cite{170r, 172r}, ant colony optimization (ACO) \cite{163r, 205r}, and artificial bee colony (ABC) \cite{181r}. Besides, some other studies \cite{163r} combined ACO with DE to seek optimal feature subsets, where the solutions searched by the ACO were fed into the DE to further explore the optimal solution. In \cite{194r}, a family of feature selection methods based on different variants of GA were developed to improve the accuracy of content-based image retrieval systems.
	
	\textbf{Co-evolution search paradigm}: In co-evolution search paradigm for feature selection,  at least two populations are simultaneously evolved and interacted toward the optimal subset of features \cite{195r, 197r}. For example, a divide-and-conquer strategy was developed in \cite{195r} to manage two subpopulations. One subpopulation was to conduct an evolution process of classifier design, while the other one was to search for an optimal subset of features.   
	
	\textbf{Multi-objective search paradigm}: This type of search paradigms are driven by two or more conflicting objectives \cite{091r, 200r, 204r}, such as the maximization of the accuracy of a classifier and minimization of the size of a feature subset. On the basis of the above two conflicting objectives, Xue et al. \cite{091r} designed a multi-objective PSO algorithm for feature selection and obtained a set of Pareto non-dominated candidate solutions for feature selection after the multi-objective search.
		
	\subsubsection{Summary}
	\ 
	
	GA and GP are widely applied to feature selection. GA early serves for low-dimensional (i.e., $\leq$1000) datasets \cite{033r, 202r}. Recently, many GA-based  approaches have been proposed to solve high-dimensional feature selection \cite{204r}. Nevertheless, GP is commonly applied to large-scale/high-dimensional feature selection since it is flexible in feature representation \cite{033r}. Especially,  GP outperforms GA on some small but high-dimensional datasets, e.g., Brain Tumor-2 \cite{208r} with 10367 features but only 50 samples. In addition, PSO has been proved with faster convergence rate to an optimal feature subset than GAs and GP \cite{209r}. The graph representation of ACO outperforms GA and GP on flexibility, but the challenge of ACO is how to design appropriate graph encoding for large-scale scenarios \cite{041r,033r}.

	\subsection{Feature Construction}
	
	\subsubsection{Problem Formulation}
	
	\ 
	
	Feature construction is to create new high-level features from the original features \cite{365r} via appropriate function operators (e.g., conjunction and average) \cite{064r, z04}, so that the high-level features are more easily discriminative than the original ones. Feature construction is a complicated combinatorial optimization problem, where search space increases exponentially along with the total number of original features and the function operators. In the following subsections, we will describe the EC-based feature construction methods in terms of both solution representations and search paradigms.
	
	\subsubsection{Solution Representations}
	
	\ 
	
	Existing EC-based approaches for feature construction can be categorized into three groups.

	\textbf{Linear encoding}: The study \cite{064r} used $n$-bit ($n$ is the total number of original features) binary vector to represent each particle, where ``0'' indicated the corresponding feature not applied to build the new high-level feature while ``1'' was in the opposite. On the basis of the encoding, a local search was performed to select candidate operators from a predefined function set to construct a new high-level feature.
	
	\textbf{Tree-based encoding}: Tree-based encoding is natural for feature construction, where leaf nodes represent the feature information and internal nodes represent operators. Many studies \cite{365r, 367r} have demonstrated the effectiveness of tree encoding in feature construction. For example, Bhanu et al. \cite{367r} designed a GP-based coevolutionary feature construction procedure to improve the discriminative  ability of classifiers. In \cite{365r}, an individual in EC was represented by a multi-tree encoding with multiple high-level features. 
	
	\textbf{Graph-based encoding}: In this encoding, the nodes and edges represent features and operators (e.g., ``+'', ``-'', ``*'', ``/''), respectively. Teller et al. \cite{368r} applied an arbitrary directed graph to represent all features and operators, where each possible high-level feature can be represented as a subgraph of this directed graph. For linear GP, features and operations form a many-to-many directed acyclic graph, in which each feature is loaded into predefined registers and register's value can be used in multiple operators \cite{z05}. However, graph encoding becomes inefficient on high-dimensional feature sets since the complexity of graph traversal exacerbates the difficulty of feature construction. 
	
	\subsubsection{Search Paradigms}
	
	\ 
	
	There are four categories of search paradigms for feature construction in existing work .
	
	\textbf{Basic EC search paradigm}: Existing studies include but are not limited to GA \cite{369r}, and GP \cite{365r, 374r, 384r}. For example, the work \cite{374r} designed GP-based feature construction to reduce the feature (input) dimensions of a classifier. Especially, GP has been also widely used to construct new features, where each individual following the form of a GP tree usually represents a constructed high-level feature \cite{371r}. 
	
	\textbf{Co-evolution search paradigm}: It can be decomposed to feature construction subproblem and classifier design subproblem, and each subproblem is solved with a standalone subpopulation by an EC-based method  \cite{367r, 378r}.  For example, the study \cite{378r} decomposed feature construction into two subproblems (i.e., feature construction, and object detection), where the feature construction was solved by evolving a population of pixel (i.e., feature) and the object detection was optimized using object detection algorithm (ODA) \cite{378r}.
	
	\textbf{Multi-features construction search paradigm}: Unlike early methods \cite{365r, 369r, 370r, 279r} only constructing one high-level feature in a single search process, this sort of paradigms are able to create multiple high-level features. For example, Ahmed et al. \cite{380r} employed Fisher criterion together with  $p$-value measure as the discriminant information between classes, based on which multiple features were constructed through multiple GP trees.	
	
	\textbf{Multi-objective evolutionary search paradigm}: In this search paradigm, the number of features and classification accuracy are commonly taken into account as the objective functions for multi-objective evolutionary optimization \cite{381r, 382r}. Especially, Hammami et al. \cite{381r} constructed a set of high-level features by optimizing a multi-objective optimization problem (MOP) with three objectives (i.e., the number of features, the mutual information, and classification accuracy) with Pareto dominance relationship. 
	
	\subsubsection{Summary}
	
	\ 
	
	GP-based approaches are popular in feature construction due to the flexible representation of features and operations. In addition, the hybrid of evolutionary algorithms also attracts much attention for feature construction. However, there is still plenty of room for the improvement of efficiency in constructing features in high-dimensional or large-scale scenarios, where a large number of computational resources are needed \cite{384r, 279r}. Notably, feature construction often requires more computational overhead than feature selection, since feature construction commonly performs after the feature selection and the quality of the selected features may influence the performance of feature construction.
	
	%% Categorization of EC-based approaches for feature selection
	
	\subsection{Feature Extraction}
	
	\subsubsection{Problem Formulation}
	
	\

	Feature extraction is to reduce the feature dimensions by altering the original features/data via some transformation functions \cite{025r}. Traditional extractors include principal component analysis (PCA) \cite{411r} and linear discriminant analysis (LDA) \cite{412r}. However, they cannot keep somewhat important information after the transformation \cite{411r} and it is tedious to tune their hyperparameters (e.g., number of retained features) to find the best extraction. Thus, automatically finding high-quality map functions by EC-based approaches to achieve informative feature set tends to be popular.
	
	\subsubsection{Solution Representations}
	
	\ 
	
	There are two typical ways for solution representation in  EC-driven feature extraction.
	
	\textbf{Linear encoding}: In this encoding,  map functions \cite{413r, 415r} or function parameters \cite{420r} are encoded as a linear format. For example, Wissam et al. \cite{415r} predefined three sets of track functions (i.e., trace functions, diametric functions, and circus functions) for feature extraction, and the optimal combination between the functions were obtained by an EC-based method. In \cite{420r}, the hyperparameters of map functions were encoded by some linear vectors which were constructed by a number of optimal projection basis vectors obtained via EC. 
	
	\textbf{Tree-based encoding}: In tree-based encoding, leaf nodes represent original features or constants, while the non-leaf nodes are some operators for feature extraction including common arithmetic, logical operators (i.e., ``+'', ``/'', ``$\cup$'') or other transformation operators (e.g., uLBP, and SobelY). In EC-driven feature extraction, an individual represents a feature extractor or map function \cite{434r,427r}. Especially, an EC-based framework was developed in \cite{427r} to search for features and sequences of operations by use of tree-based encoding.
	
	\subsubsection{Search Paradigms}
	
	\ 
	
	In this section, some common search paradigms for feature extraction are introduced. 
	
	\textbf{Basic EC search paradigm}: EC has been successfully utilized in various feature extraction tasks \cite{435r, 436r}. For example, Zhao et al. \cite{421r} introduced bagging concept to an evolutionary algorithm for feature extraction. The work in \cite{422r} developed an evolutionary discriminant feature extraction (EDFE) algorithm by combining GA with subspace analysis, which can reduce the complexity of the search space and improve the classification performance.

	\textbf{Co-evolution search paradigm}: In feature extraction, finding the optimal extractor is an optimization problem, which can be decomposed into a series of subproblems \cite{433r, 439r}. For example, Hajati et al. \cite{439r} proposed a co-evolutionary method for feature extraction. Specifically, a subpopulation was evolved to optimize the classifier-related subproblem (i.e., classifier construction), and the other subpopulation made use of genetic information from the first population for the optimization of a feature-related subproblem (i.e., feature extraction).

	\textbf{Multi-objective search paradigm}: In multi-objective feature extraction, the model accuracy, computational time, complexity, and robustness are often taken into account as the objectives \cite{427r, 442r}. Cano et al. \cite{442r} proposed a Pareto-based multi-objective GP algorithm for feature extraction and data visualization, where the objectives were to minimize the complexity of data transformation (i.e., tree size) and maximize the recognition performance (i.e., accuracy).
	
	\subsubsection{Summary}
	
	\

	In existing studies, many efficient searching and balancing strategies, driven by EC approaches to achieve satisfactory solutions at significantly-reduced computation overheads, have been developed in recent years \cite{427r, 442r, 445r, 444r}. However, the performance of extractors may be limited with existing encoding methods and predefined operation sets. Therefore, it is essential to develop efficient algorithms, operation control strategies and representation for high-dimensional feature extraction.

	\section{Model Generation}\label{EDL-MG}
	
	Model generation is to search for optimal models with desirable learning capability on given tasks \cite{024r, 025r}. In this section, we introduce corresponding evolutionary  parameter optimization, architecture optimization, and joint optimization from solution representation to search paradigms. Readers interested in other model generation approaches (e.g., RL-based and gradient-based approaches) can refer to the reviews \cite{045r, 468r}.
	
	\subsection{Model Parameter Optimization}
	
	\subsubsection{Problem Formulation}
	
	\ 
	
	Model parameter optimization targets at searching for the best parameter set (i.e.,  weights $W^*$) for a predefined architecture ($A$). The loss function $L$ (e.g., the cross-entropy loss function) measures the performance of the model with optimized parameters (i.e., $W$ in Eq. \ref{Eq5}) on given datasets. The general model parameter optimization can be formulated as
	\begin{equation}\label{Eq5}
	\begin{aligned}
	\begin{matrix}
	{{W}^{*}} $=$ \underset{W}{\arg \min} L\left(W,A\right)
	\end{matrix}  \\
	\end{aligned}
	\end{equation}
	where $W$ is usually large-scale (millions of model parameters) and highly non-convex. 
	
	\subsubsection{Solution Representations}
	
	\ 
	
	There are two typical EC-based representation schemes for model parameter optimization, including direct encoding and indirect encoding \cite{025r}.

	\textbf{Direct encoding}: The model parameters are directly represented via a vector or matrix, in which each element represents a specific parameter \cite{103r,472r}. For example, a chromosome with 64 real numbers was used to directly represent the network corresponding weights, where the first 63 real numbers were used to encode three convolution masks of size 1 $\times$ 21. The last real number was the random seed of a generator for the initialization of a fully connected network \cite{103r}. This encoding approach may require a huge computational overhead to represent and optimize the large-scale weights.

	\textbf{Indirect encoding}: This encoding approach represents only a subset of the model parameters via a deterministic transformation \cite{504r, 373r}. In \cite{373r}, the weight information was encoded as a set of Fourier coefficients in the frequency domain to reduce dimensionality of representation by ignoring high-frequency coefficients. Although this method is able to speed up the search process, the loss of parameter the information may occur due to the incomplete information representation, which maydegrades the model performance.
	
	\subsubsection{Search Paradigms}
	
	\

	EC-based methods for model parameter optimization can be divided into two categories according to whether or not method combines with the gradient approach, i.e., pure EC and gradient-based EC.

	\textbf{Pure EC} paradigms optimize model parameters only via evolutionary search, including the 
	basic EC search paradigm and co-evolution search paradigm.
	\begin{itemize}
		\item
		\textbf{Basic EC search paradigm}: In addition to GA \cite{471r, 472r}, some heuristic algorithms like PSO \cite{481r}, ABC \cite{490r} and ACO \cite{491r} are also commonly utilized for model parameter optimization. For example, Karaboga et al. \cite{490r} adopted ABC to find a set of weights for a feed-forward neural network (FNN) on targeted tasks.
		
		\item
		\textbf{Co-evolution search paradigm}: Co-evolution search is conducted on the subproblems of the original optimization problem (e.g., synapse-based and neuron-based problems \cite{492r, 493r}). For example, Chandra et al. \cite{492r} regarded a single hidden layer as a subcomponent in the initialization phase, which will be merged with the individuals with the best fitness from different sub-populations to constitute new neural networks during the co-evolution optimization process. 
		
	\end{itemize}
	
	\begin{figure}[h]
		\centering
		\begin{subfigure}{0.325\linewidth}
			\centering
			\includegraphics[width=1\linewidth]{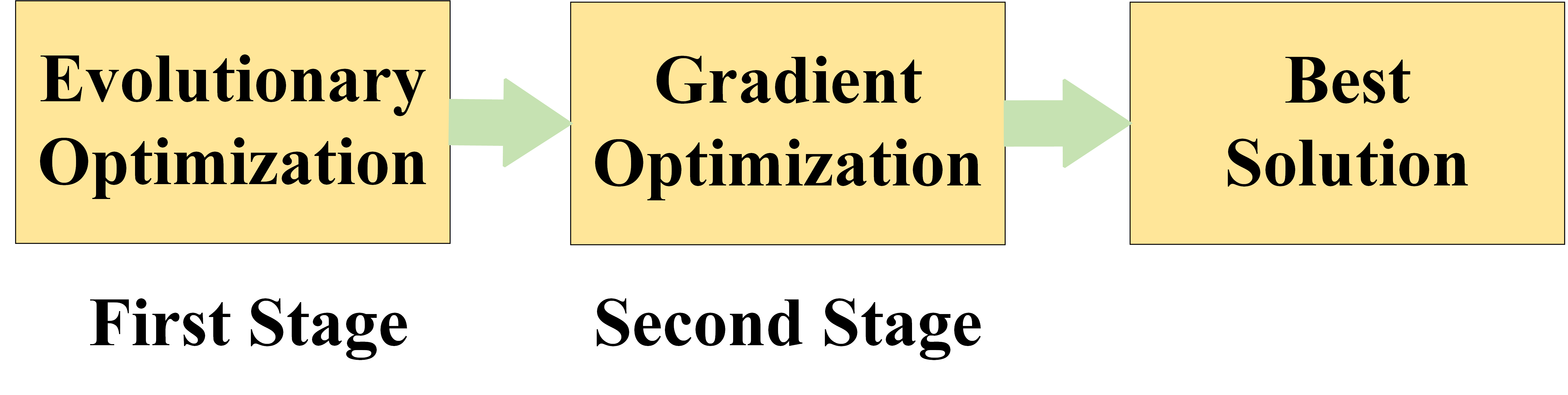}
			\caption{}
			\label{fig18a}%文中引用该图片代号
		\end{subfigure}
		\centering
		\begin{subfigure}{0.325\linewidth}
			\centering
			\includegraphics[width=1\linewidth]{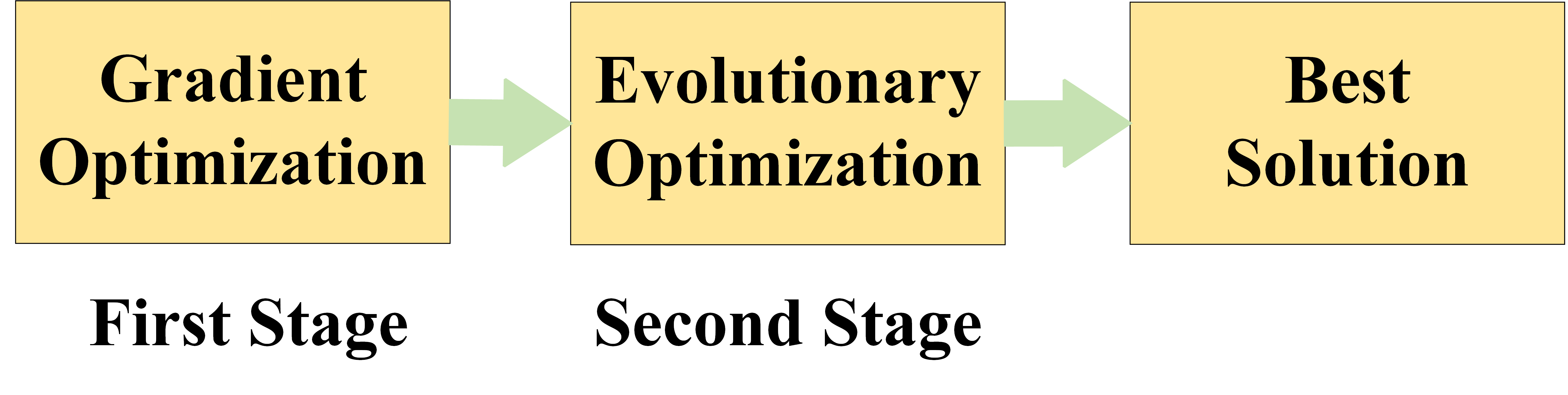}
			\caption{}
			\label{fig18b}%文中引用该图片代号
		\end{subfigure}
		\centering
		\begin{subfigure}{0.325\linewidth}
			\centering
			\includegraphics[width=0.9\linewidth]{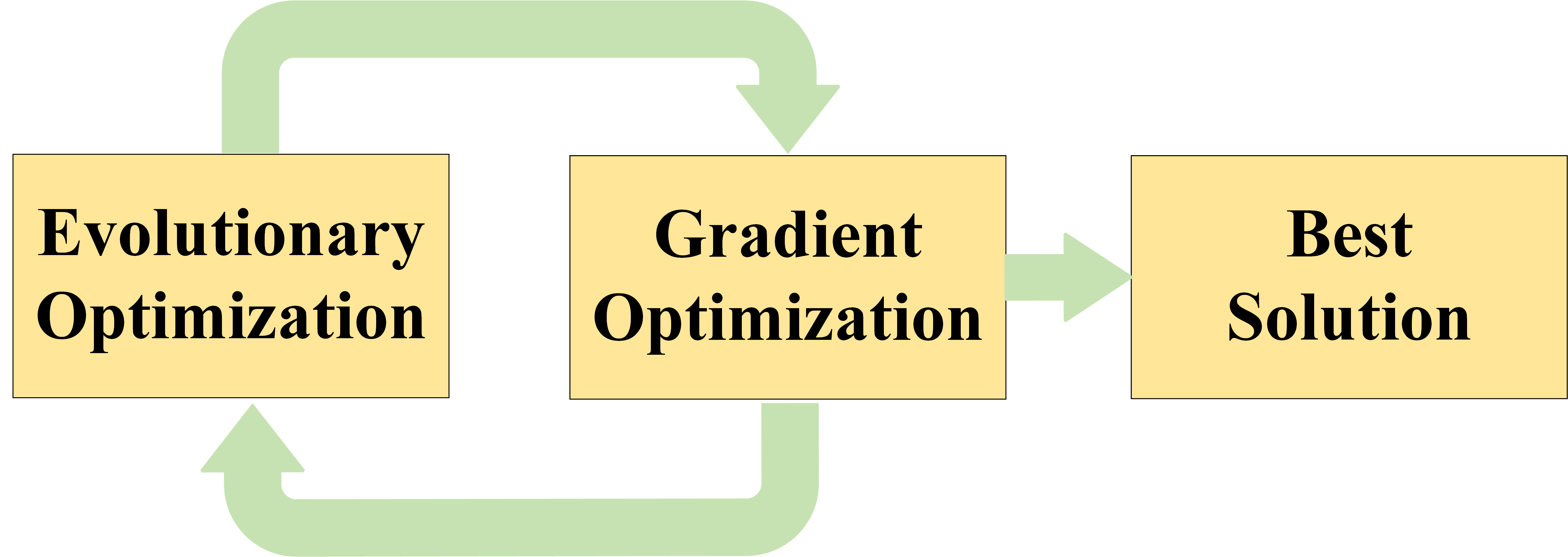}
			\caption{}
			\label{fig18c}%文中引用该图片代号
		\end{subfigure}
		\caption{ Three hybrid ways of gradient-based ECs.}
		\label{fig18}
	\end{figure}
	
	\textbf{Gradient-based EC} combine basic EC with the gradient-based method to enhance the exploitation ability in optimizing model parameters. According to the execution order, there are three hybrid ways.
	
	\begin{itemize}
		\item
		The first hybridization approach is shown in Fig. (\ref{fig18a}), where the EC is used to identify the optimal parameters for model, then the parameters are further optimized using gradient-based method to find the final optimal solution \cite{z06, z07}. For example, a genetic adaptive momentum estimation algorithm (GADAM) was proposed in \cite{531r} by incorporating Adam and GA into a unified learning scheme, where Adam was an adaptive moment estimation method with first-order gradient.
		
		\item
		The second hybridization approach is given in Fig. (\ref{fig18b}), where the gradient-based method is used to produce a set of parameters for the initialization of the population used in EC \cite{528r}. For example, the study \cite{532r} firstly trained a RL agent through a gradient-based method, then the parameters of the RL were used as the initial population to feed the EC. As a result, the parameters will be further optimized by the EC.  
		
		\item
		The third approach is presented in Fig. (\ref{fig18c}), which iteratively applies EC and gradient-based method during the optimization to find the optimal parameters. Following this framework, when the method is used and which method is chosen are varying in different studies \cite{513r, 514r}.
	\end{itemize}
	
	\subsubsection{Summary}
	
	\ 
	
	In model parameter optimization, direct encoding is straightforward and able to keep more information than the indirect encoding. Compared to gradient-based methods easily trapped into local optima, EC shows more powerful ability in global search. Here, several scenarios are introduced as follows, where EC is applied to model parameter optimization.
	
	\textbf{Small-scale scenario}: \cite{471r} shows that pure EC approaches outperform gradient-based methods in search effectiveness on some small-scale problems, where the models are with small numbers of parameters or simple architectures (e.g., FNN).
	
	\textbf{Large-scale scenario}: The performance of pure EC approaches might not be promising in large-scale learning models, while a better way is to utilize the  hybridization of EC and gradient-based methods. Such hybrid methods can alleviate the issue of getting trapped in local optima and increase the effectiveness of subsequent exploitation \cite{513r}.
	
	In addition to above scenarios, EC-based method can be used to train the DNN, when the exact gradient information of the loss function is difficult to be acquired \cite{z08}. For example, the rewards of policy network are sparse or deceptive in deep reinforcement learning (DRL) so that the gradient information is unattainable. The work \cite{535r} introduced the novelty search (NS) and the quality diversity (QD) to the evolution strategies (ES) for the policy network.

	\subsection{Model Architecture Optimization}
	
	\subsubsection{Problem Formulation}
	
	\ 
	
	Model architecture optimization, also termed as NAS, is to search promising network architectures with good performance such as model accuracy on given tasks. The model architecture optimization can be formulated as follows.
	\begin{equation}\label{Eq6}
	\left\{ \begin{matrix}
	{{A}^{*}}$=$\underset{W,A}{\mathop{\arg \min }}\,L\left(W,A\right)  \\
	\begin{matrix}
	s.t. & A\in \mathcal{A}  \\
	\end{matrix}  \\
	\end{matrix} \right.
	\end{equation}
	\noindent where $A^*$ indicates the architecture from the search space ($\mathcal{A}$) with the best performance under the parameters $W$, and $L$ is used to measure the performance of architectures on given tasks. Thereby, this optimization is a bi-level optimization problem \cite{031r,037r}, where the model architecture optimization is subject to the model parameter optimization \cite{032r}. Since the current NAS works are mainly focused on CNN, we will discuss the solution representations, the search paradigms, and acceleration strategies of CNN. Due to the page limit, the design of the search space of NAS are not introduced here, but interested readers can check these surveys \cite{037r, 045r} which have details about search space design. 
	
	\subsubsection{Solution Representations}
	
	\ 
	
	According to varying lengths of encodings, we can classify the encoding strategies into fixed-length and variable-length encodings. 
	
	\textbf{Fixed-length encoding}: The length of each individual is fixed during the evolution. For example, a fixed-length vector is designed to represent the model architecture of CNN \cite{137r}, where a subset of elements in the vector represents an architectural units (e.g., convolutional, pooling or fully-connected layer) of a CNN. Such encoding may be easily adapted to evolutionary operations (e.g., crossover and mutation) of EC \cite{137r}, but it has to specify an appropriate maximal length, which is usually unknown in advance and needs to predefine based on domain expertise.

	\textbf{Variable-length encoding}: Different from the fixed-length approach, the variable-length encoding strategy does not require a prior knowledge about the optimal depth of model architecture and actually could be a way to reduce the complexity of the search space. The flexible design of this encoding may encode more detailed information about the architecture into a solution vector, and the optimal length of the solution is automatically found during the search process \cite{037r}. In \cite{152r}, the entire variational autoencoder (VAE) was divided into four blocks, including h-block, $\mu$-block, $\sigma$-block and t-block, while the variable-length chromosomes consisted of different quantities and types of layers. Notably, variable-length encoding it is not straightforward to apply standard genetic operators (e.g., crossover).
	
	Since the neural network architectures are composed of basic units and connections between them, so that both of them are to be encoded, as suggested in \cite{037r}.
	
	{\textbf{1) Encoding hyperparameters of basic units.}}
	In CNNs, there are many hyperparameters to be specified for each unit (e.g., layer, block or cell), such as feature map size, type of convolution layer, and filter size \cite{036r}. In \cite{104r}, DenseBlock only had to set two hyperparameters (e.g., block type and specific parameter of internal unit) to configure the block can be seen as a microcosm of a complete CNN model. The parameterization of a cell is more flexible than that of a block since it can be configured via a combination of different primitive layers \cite{029r}.
	
	{\textbf{2)	Encoding connections between units.}}
	In general, there are two kinds of model architectures according to the connection patterns of basic units: linear topological architectures and non-linear topological architectures \cite{610r}. The linear pattern of architecture consists of sequential basic units, and the non-linear pattern allows for skip or loop connections in the architecture \cite{037r}.

	\begin{itemize}
		\item
		\textbf{Linear topological architecture}: The linear topology widely appears in the construction of layer-wise and  block-wise search spaces. Due to the simplicity of linear topology, basic units can be stacked one by one by a linear piecing method. In this way, the skeleton of an architecture can be built up effectively \cite{104r, 152r} regardless of the complexity of the internal of basic units.
		
		\item
		\textbf{Non-linear topological architecture}: Compared to the linear architecture, the non-linear topological architecture receives much more attention due to its flexibility to construct well-performing architectures \cite{126r, z10, 137r}, such as macro structures composed of basic units, and micro structures within basic units. There are two typical encoding approaches for non-linear topological architectures. The one is to use adjacent matrix to represent the connections in non-linear architectures, where ``1'' of the matrix  denotes the existence of the connection between two units and ``0'' goes the opposite. In \cite{662r}, skip connections are represented by a matrix where constraints can be set in place to guarantee valid encoding and avoid recurrent edges while performing skip connections. Note that adjacent matrix has a limitation that the number of basic units needs to be fixed in advance \cite{674r}. Another one is to utilize an ordered pair to represent a directed acyclic graph, and then encode the connections between unites. The ordered pair can be formulated as $G$ = ($V$, $E$) where $V$ is a set of vertices and $E$ is a directed edge in the acyclic graph, and it has been applied in \cite{673r} to encode the connections.
		
	\end{itemize}
	
	\subsubsection{Search Paradigms}
	
	\

	In this section, the commonly used EC-based search paradigms for NAS are introduced.
	
	\textbf{Basic EC search paradigm}: Many basic EC algorithms have been widely applied in existing NAS methods, such as GA \cite{674r} and PSO \cite{663r}. A general framework of EC is presented in 
	Fig. \ref{fig9}. 
	
	\textbf{Incremental search paradigm}: A model architecture can be built in an incremental way where model elements (e.g., layers and connections) are gradually added to the model during the evolutionary process \cite{016r, 117r, 644r}. This way allows to find parts of architecture at different optimization stages, which reduces the computational burden on acquiring a complete model at once \cite{016r}. For example, Wang et al. \cite{117r} used an incremental approach to stack blocks for building architectures, which improved the capacity of the final architecture via a progressive process.
	
	\textbf{Co-evolution search paradigm}: An architecture optimization problem is decomposed into the optimizations of a blueprint and its components \cite{648r, 649r}. Specifically, the blueprint plays a role in specifying the topological connection patterns of its components, and an optimal architecture is acquired by cooperatively optimizing the blueprint and its components. For example, O'Neill et al. \cite{648r} proposed a co-evolution search paradigm for NAS, where the candidate blueprints and components were sampled from two populations, and then combined to form new architectures.

	\textbf{Multi-objective search paradigm}: This paradigm targets at searching for a set of Pareto optimal architectures based on multiple criteria, and finding the final solutions according to some practical considerations, such as computational environment \cite{599r, 611r}. This paradigm becomes popular in practical applications, since many objectives are required to be considered such as the accuracy, inference time, model size, and energy consumption. In \cite{599r}, NSGA-II and RL were used to explore model architectures with respect to the model accuracy, and model complexity (e.g., the number of model parameters and multiply-adds operators). 
	
	\subsubsection{Acceleration Strategies}
	
	\ 
	
\begin{figure}[h]
		\centering
		\includegraphics[width=0.6\textwidth]{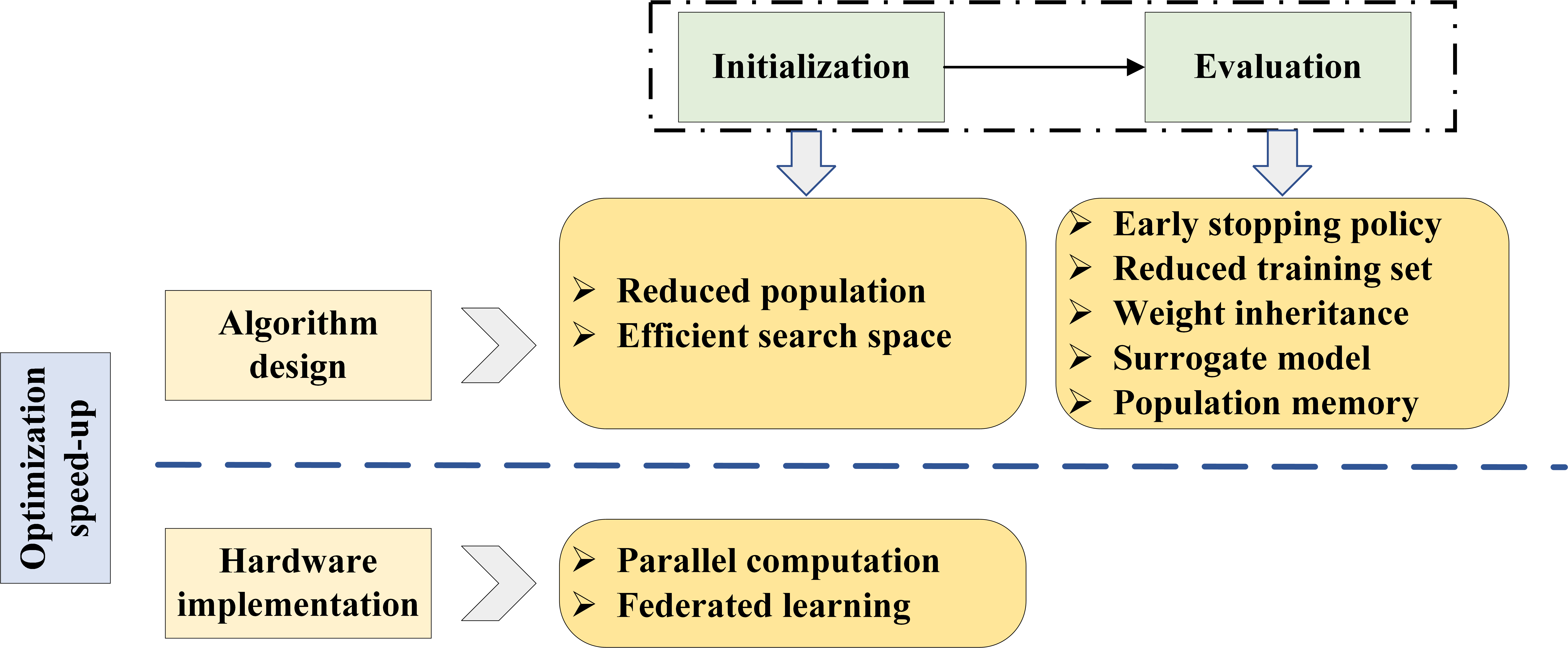}
		\caption{Overview of acceleration strategies.}
		\label{fig28}
	\end{figure}

	NAS is a high computational overhead task, mainly due to the large search space and highly time-consuming evaluation \cite{024r}. To overcome this challenge, various acceleration strategies \cite{663r,664r} have been developed to accelerate the optimization. In this section, we summarize the speed-up strategies from the aspects of algorithm design to the hardware implementation, as shown in Fig. \ref{fig28}.
	
	From the algorithm design point of view, we summarized a number of acceleration strategies from population initialization to evaluations.
	\begin{itemize}
		\item \textbf{Initialization}: 
		
		\begin{itemize}
			\item {\em{Reduced population}}: The simplest way of acceleration during the initialization stage is to set the population with a small size. In other words, less evaluations are required with a smaller size of population since the evaluation of a candidate architecture is time-consuming \cite{024r}. As a result, some studies \cite{664r,675r} use small population with fixed size to speed up their evolution, like CARS (size = 32) \cite{610r}. In contrast, some other studies use dynamic sizes of populations during the optimization. In \cite{598r}, the population size is dynamically changed to reach a balance between algorithmic efficiency and population diversity. 
			
			\item {\em{Efficient search space}}: Another way is to design efficient search space to speed up the search process. For example, an architecture constructed on the basis of cell-wise search space \cite{037r} is composed of many similar structures of cells and only representative cells need to be optimized, which contributes to significant computational speed-up.
		\end{itemize}

		\item \textbf{Evaluations}: 
		
		\begin{enumerate}
			\item {\textbf{Early stopping policy:}} A relatively small number of training epochs are used to reduce the training cost (i.e., early stopping policy) since the training time is reduced \cite{548r, 549r,663r}.
			
			\item {\textbf{Reduced training set:}} Some methods are designed to reduce the size of the training set to improve training efficiency at the expense of a little accuracy \cite{661r,577r}. Besides, low-resolution data (e.g., ImageNet 32\time32) \cite{676r} is also commonly used as the training set to accelerate the search process for the optimal architecture. 
			
			\item {\textbf{Weight inheritance \footnote{In \cite{z11,z12,z13}, Rumelhart/Hinton/LeCun used the term "weight sharing" to mean that different network connections/links share the same set of weights, and pointed out that "weight sharing" is the core of shared weight NNs/CNNs. More recent \cite{032r, 781r} use of this term refers to "weight/parameter replications" or "weight inheritance".}:}}
			\begin{itemize}
				\item {\em{Supernet-based inheritance}}: uses an over-parameterized and pre-trained supernet to encode all candidate architectures (i.e., subnets). In other words, the subnets share weights of the identical structures from the supernet, and they are directly evaluated on the validation dataset to obtain their model accuracy \cite{558r, 559r, 605r, 661r}. 
				
				\item {\em{Parent-based inheritance}}: inherits weights from previously-trained networks (i.e., parental networks) instead of a supernet, since offspring individuals retain some identical parts of their parental architectures \cite{539r, 651r, 658r, 545r} \cite{539r}. As a result, offspring architectures can inherit the weights of the identical parts and no longer need to be trained from scratch.
			\end{itemize}
			
			\item {\textbf{Surrogate model}}: Since the evaluation of an architecture is time-consuming \cite{024r, 037r}, cheap surrogate models have been introduced in NAS as performance predictors to reduce the computational time \cite{032r}.   
			
			\begin{itemize}
				
				\item {\em{Online performance predictors}}: They are trained online on the datasets sampled from past several epochs \cite{037r}, including a sequence of data pairs with different training epochs and their corresponding performance of these epochs \cite{032r}. After that, they will be used for the performance prediction on new architectures. To reduce the true evaluations of architectures, some performance predictors directly predict whether a candidate architecture can be survived into next iteration through a trained ranking or classification method, such as classification-wise NAS \cite{677r}. 
				
				\item{\em{Offline performance predictors}}: They are essentially a sort of regression models  mapping the architectures to specific performance. End-to-end predictors can be trained in an offline manner, so that they are able to predict the performance of architectures during the entire search process. Consequently, they can significantly reduce the computational burden \cite{123r,016r,678r}. 
				
			\end{itemize}
			
			\item \textbf{Population memory}: 
			Population memory is used to store elite individuals from different generations during the optimization \cite{036r,540r}. When a new individual is generated, it does not need to be evaluated again if it is the same as an individual in the memory. In other words, the performance of individuals sharing the same architectures are the same and can be acquired via the population memory instead of training from scratch. This mechanism relies on the fact that similar or same individuals may repeatedly appear in different generations. 
			
		\end{enumerate}
		
	\end{itemize}
	
	According to the above introduction, we can conclude that many of them improve the search efficiency at the expense of sub-optimiality. For example, a small population cannot well cover a multi-objective optimal front. Parameter sharing may lead to the biased search due to much similarities among the individuals. Highly accurate surrogates need a large number of training data, which are commonly time-consuming. Population memory heavily relies on the random emergence of similar or same individuals.  
	
	\textbf{Hardware implementation}: Importantly, a powerful hardware platform can significantly speed up the search process under the reasonable utilization of computing resources (e.g., cloud computing \cite{679r} and volunteer computers \cite{680r}). Parallel computation is a powerful tool to decompose large search problems into small sub-problems, which can be simultaneously optimized by several cheaper hardware\cite{684r, 685r}. For example, Lorenzo et al. \cite{682r} proposed a parallel PSO algorithm to search for optimal architecture of CNN. The security of the computing device also becomes an important consideration. For this reason, an emerging decentralized privacy-preserving framework is applied to NAS, which unites multiple local clients to collaboratively learn a shared global model trained on the parameters or gradients of the local models, instead of the raw data. For example, Zhu et al. \cite{612r} firstly proposed a real-time federated NAS that can not only optimize the model architecture but also reduce the computational payload. Specifically, the decentralized system is able to accelerate the algorithm efficiency of federated NAS. Besides, data encryption is employed on the transmitted data (parameters or gradients of the local models) between the clients and the server to ensure the privacy even though all of the training are performed in local. Accordingly, federated NAS is highly efficient and secure, which may become a new hot research topic.  
	
	\begin{table}[htbp]\scriptsize
		\centering
		\caption{Different acceleration strategies}
		\begin{tabular}{l|l|l|l|p{16.055em}}
			\hline
			\multicolumn{1}{c|}{\multirow{8}[16]{*}{Algorithm design}} & Initialization & \multicolumn{2}{l|}{Reduced population} &\cite{664r},\cite{598r},\cite{577r},\cite{610r} \bigstrut\\
			\cline{2-5}          & \multirow{6}[14]{*}{Evaluation} & \multicolumn{2}{l|}{Early stopping policy} &\cite{663r},\cite{548r},\cite{587r},\cite{660r},\cite{146r},\cite{664r},\cite{115r} \bigstrut\\
			\cline{3-5}          &       & \multicolumn{2}{l|}{Reduced training set} &\cite{661r},\cite{577r},\cite{117r} \bigstrut\\
			\cline{3-5}          &       & \multirow{2}[4]{*}{Weight inheritance} & Supernet-based sharing &\cite{661r},\cite{558r},\cite{559r},\cite{605r} \bigstrut\\
			\cline{4-5}          &       &       & Parent-based sharing &\cite{651r},\cite{658r},\cite{545r},\cite{548r},\cite{660r},\cite{088r},\cite{065r} \bigstrut\\
			\cline{3-5}          &       & \multirow{2}[4]{*}{Surrogate model} & Online performance predictors &\cite{032r},\cite{016r}, \cite{677r}, \cite{686r} \bigstrut\\
			\cline{4-5}          &       &       & Offline performance predictors &\cite{123r},\cite{687r},  \cite{151r}\bigstrut\\
			\cline{3-5}          &       & \multicolumn{2}{l|}{Population memory} &\cite{036r},\cite{077r},\cite{665r},\cite{608r} \bigstrut\\
			\hline
			\multicolumn{4}{l|}{Hardware implementation} &\cite{679r},\cite{684r},\cite{685r},\cite{612r} \bigstrut\\
			\hline
		\end{tabular}
		\label{table8}%
	\end{table}%
	
	Table \ref{table8} lists the common acceleration strategies to improve the algorithm efficiency. It is noted that multiple strategies can be utilized together to improve computational speed-up. For example, Lu et al. \cite{032r} employed supernet and learning curve performance predictor in NAS, while Liu et al. \cite{577r} leveraged a small populations size and a small dataset to reduce the time overhead of evaluation.
	
	\subsubsection{Summary}
	
	\ 
	
	Most NAS methods are based on basic EC search paradigms on an entire-structured search space, which are introduced above. However, there are also some other automatic search techniques such as RL-based \cite{468r}, Bayesian-based \cite{692r}, and gradient-based \cite{694r} methods, for architecture search.
	
	RL-based methods can be regarded as an incremental search, where a policy function is learned by using a reward-prediction error to drive the generation of incremental architecture. Due to the large-scale of state space and action space, RL-based methods require immense computational resources. In addition, there are a large number of hyper-parameters (e.g., discount factor) in RL-based NAS. Besides, they transform a multi-objective optimization problem into a single-objective problem via a priori or expert knowledge, so they are unable to find a Pareto optimal set to the target tasks. 
	
	Bayesian-based methods are a common tool for hyperparametric optimization problems with low dimensions. In comparison to EC-based methods, they are much more efficient on the condition that a proper distance function has to be designed to evaluate the similarities between two subnets. However, the computational cost of Gaussian process grows exponentially and its accuracy decreases, when the dimensionality of the problem increases.
	
	Gradient-based methods, taking a NAS problem as a continuous differentiable problem instead of a discrete one, are able to efficiently search architectures with proper weight parameters. Unfortunately, their GPU costs are usually very high due to a large number of parameters to be updated in gradient-based algorithms \cite{694r}. 
	
	In contrast, EC-based methods benefit from less hyperparameters to be optimized and no distance functions to be designed. In addition, EC-based methods can be applied to NAS with multiple objectives and constraints. Although there many acceleration strategies in EC-based methods, they still suffer from high computational overheads.

	\subsection{Joint Optimization}
	
	\subsubsection{Problem Formulation}
	
	\ 
	
	The independent optimization of architecture or parameters is difficult to achieve the optimal model on give tasks. Hence, joint optimization methods have been developed to search for the optimal configuration of architecture ($A^*$), and parameters ($W^*$, associated weights). The optimization problem can be defined in Eq. \ref{Eq7}. 
	
	\begin{equation}\label{Eq7}
	\begin{aligned}
	\begin{matrix}
	\left({{W}^{*}},{{A}^{*}}\right) $=$ \underset{W,A}{\arg \min} L\left(W,A\right)
	\end{matrix}  \\
	\end{aligned}
	\end{equation}
	where $L$ is the loss function. 
	
	In the followings, we will introduce the joint optimization regarding the solution representations and search paradigms, and then discuss the pros and cons of EC-based methods in comparison to others.
	
	\subsubsection{Solution Representations}
	
	\ 
	
	There are three typical classes of encoding schemes used for joint optimization.
	
	\textbf{Linear encoding}: This is a simple but effective encoding strategy, which has been widely used in many studies to build architecture with high performance \cite{696r,701r}. In \cite{701r}, a variable-length binary vector was used to represent weights and structure of neural networks, where the weights utilize direct encoding.
	
	\textbf{Tree-based encoding}: In this encoding, the topology and weights of an architecture can be represented by a tree structure with a number of nodes and edges \cite{727r, 728r}. In \cite{703r}, the mechanism of Reverse Encoding Tree (RET) was developed to ensure the robustness of a deep model, where the topological information of an architecture was represented by a combination of nodes and the weight information was recorded on the edges.

	\textbf{Graph-based encoding}: In this encoding, the nodes of a graph represent neurons or other network units, and the edges are used to record the weight information \cite{711r,755r}. For example, a graph incidence matrix was developed in \cite{515r} to encode a neural network. The size of the matrix was set to ($N_i$ + $N_h$ + $N_o$) $\times$ ($N_i$ + $N_h$ + $N_o$), where $N_i$, $N_h$ and $N_o$ indicate the numbers of input, hidden, and output nodes, respectively. In the graph incidence matrix, real numbers represented the weight and biases, and ``0'' meant that there was no connection between two nodes.
	
	\subsubsection{Search Paradigms}
	
	\ 
	
	There are a number of effective search paradigms for joint optimization, and the EC-based search paradigms are in the spotlight.
	
	\textbf{Basic EC search paradigm}: Some basic EC search methods have been employed to handle joint optimization problems \cite{302r, 515r, 035r, 705r,707r}. In \cite{515r},  an architecture and its corresponding weights were simultaneously optimized by an EC-based method using linear and graph encodings. In neuro-evolution of augmenting topologies (NEAT) \cite{035r}, the architecture of a small network is evolved by an incremental mechanism, while the weights are optimized by an EC-based method. NEAT is able to ensure the lowest dimensional search space over all generations. Some representative studies on NEAT are presented in \cite{705r,707r}.
	
	\textbf{Multi-objective search paradigm}: Multi-objective optimization on model design has been developed in many studies (e.g., artificial neural network \cite{708r} and recurrent neural network \cite{709r}). For example, Smith et al. \cite{709r} built a bi-objective optimization (i.e., the minimizations of the mean squared error (MSE) on a training dataset and the number of connections in the network) to search for optimal weights and connections of network architectures. The chromosome of an individual was composed of two parts, where the one with Boolean type represented the structure of a network, and the other with real values represented the weights.
	
	\subsubsection{Summary}
	
	\ 
	
	Direct encoding is used to be prevalent in the joint optimization of small-scale neural networks \cite{515r,717r}. However, with the increase of the scale of neural networks, direct coding of high-dimensional vector or matrix of weights is not realistic. Therefore, recent studies are more on indirect encoding. For example, a complex mapping with acceptable accuracy loss is designed in \cite{735r, 756r} to construct weight vectors with arbitrary size.   
	
	EC-based approaches with capability of searching the optimal solution have been developed to configure a DL model for the specific task. However, they often encounter a prohibitive computational cost, which is even higher than that of model architecture optimization. Hence, designing efficient EC-based approaches for architecture and parameter search deserves much investigation.

	\section{Model Deployment}\label{EDL-MD}
	
	The large-scale DNNs are not straightforward to be deployed into devices (e.g., smartphones) with limited computation and storage resources (e.g., battery capacity and memory size). To solve this issue, various model compression approaches have been proposed to reduce the model size and inference time, such as pruning, model distillation, and quantization \cite{042r}. However, they need much expert knowledge and a lot of efforts on the manual compression of neural network models. In contrast, EC-based approaches are automation approaches and has been recently introduced to achieve automated model compression. We have observed that most of them concentrate on the area of model pruning. 
	
	\subsection{Model Pruning}

	\subsubsection{Problem Formulation}
	
	\ 
	
	DNN is commonly an over-parameterized model, which has redundant and non-informative components (e.g., weights, channels and filters). To address this issue, researchers have designed various pruning approaches (e.g., channel pruning \cite{765r}) to obtain a lightweight deep network model with high accuracy. Model pruning can be formulated as
	
	\begin{equation}\label{Eq8}
	\begin{aligned}
	& Los{{s}_{A_{s}^{*}}}\approx Los{{s}_{A}} \\ 
	& \begin{matrix}
	\text{s}\text{.t}\text{.} & {A}_{{s}}^{*}$=$\underset{{C}}{\mathop{\text{pruning}}}\,\left({A, C}\right)  \\
	\end{matrix} \\ 
	\end{aligned}
	\end{equation}
	
	\noindent where $C$ represents redundant and non-informative units, $A$ and $A{_s}^{*}$ represent original model and lightweight model, respectively, $Los{{s}_{A_{s}^{*}}}$ and $Los{{s}_{A}}$ represent loss of $A{_s}^{*}$ and $A$. 
	
	This study aims to introduce EC-based methods for model pruning, and readers interested in traditional pruning methods such as weight-based pruning, neuron-based pruning, filter-based pruning, layer-based pruning, and channel-based pruning may refer to the surveys \cite{042r} to get more details. 
	
	\subsubsection{Solution Representations}
	
	\

	For model pruning, binary encoding is one of the most popular approaches among these solution representations, where each element corresponds to the network component (e.g., channel). In \cite{811r}, the network pruning task was formulated as a binary programming problem, where a binary variable was directly associated with each convolution filter to determine whether or not the filter took effect. Although binary representation is straightforward, the length of the representation becomes large when the model complexity (i.e., the number of units) improves, and the overhead of exploration will also increase.
	
	To address the above issue, some efficient solution representations (i.e., indirect encoding) have been developed. For example, Liu et al. \cite{813r} used $N$ digits to record the number of compressed layers. The first digit represented the number of compressed layers, and following digits recorded the selected compression operator index of each layer. This way can significantly improve the search efficiency. In \cite{768r}, encoding vectors are used to represent the number of channels in each layer for original networks. Then a meta-network is constructed to generate the weights according to network encoding vectors. By stochastically fed with different structure encoding, the meta-network gradually learns to generate weights for various pruned structures.
	
	\subsubsection{Search Paradigms}
	
	\ 
	
	The search paradigms in model pruning studies can be categorized into two main groups.
	
	\textbf{Basic EC search paradigm}: A number of studies introduce single-objective EC search paradigm for model pruning \cite{814r, 769r}. For example, Wu et al. \cite{814r} first analysed the pruning sensitivity on weights via differential evolution (DE), and then the model was compressed by iteratively performing the weight pruning process according to the weight sensitivity. In addition, this method adopted a recovery strategy to increase the pruned model performance during the fine-tuning phase.
	
	\textbf{Multi-objective search paradigm}: Recently, this sort of search paradigm has been adopted to model pruning, which is able to provide users with a set of Pareto lightweight models. For example, Zhou et al. \cite{772r} considered two objectives (i.e., minimizing convolutional filters and maximizing the model performance) for biomedical image segmentation. During the model pruning, a classical multi-objective optimization algorithm (NSGA-II \cite{773r}) was used find the optimal set of non-dominated solutions, where the optimization was based on a binary string encoding (each bit represents a filter).

	In Table \ref{table11}, we have summarized these two categories of search paradigms as well as their corresponding ways of encoding.
	
	\begin{table}[htbp]\scriptsize
		\centering
		\caption{Different search paradigms and solutions representations for model pruning}
		\begin{tabular}{m{8.445em}|m{29.89em}|m{10.89em}}
			\hline
			\multicolumn{1}{r|}{} & \multicolumn{1}{l|}{Direct encoding} & \multicolumn{1}{l}{Indirect encoding} \bigstrut\\
			\hline
			Basic EC search paradigm & \cite{818r},\cite{812r},\cite{769r},\cite{814r},\cite{820r},\cite{805r},\cite{821r},\cite{771r},\cite{823r},\cite{824r},\cite{826r},\cite{831r},\cite{832r} & \cite{768r} \bigstrut\\
			\hline
			Multi-objective  search paradigm & \cite{772r},\cite{827r},\cite{829r},\cite{837r},\cite{838r},\cite{839r},\cite{840r} & \cite{068r},\cite{828r},\cite{607r} \bigstrut\\
			\hline
		\end{tabular}%
		\label{table11}%
	\end{table}%

	\subsection{Other EC-based Model Deployment Methods}
	Different from model pruning, there are several other EC-based model compression methods for model deployment. In the followings, some typical methods are introduced, including knowledge distillation, low-rank factorization, and EC for hybrid techniques. 
	
	\subsubsection{Knowledge Distillation}
	
	\ 
	
	Knowledge distillation (KD) \cite{842r} aims to get a small light network but with good generalization capability. The basic idea is to transfer the knowledge learned from a big cumbersome network (or teacher network) with good generalization ability to a small but light network (or student network). 
	
	However, knowledge distillation may be seriously influenced when there is a big gap in the learning capability between the teacher and student networks. In other words, if the difference is large, the student network may not be able to learn knowledge from the teacher network. Recently, several EC-based approaches have been proposed to mitigate the above issue of knowledge distillation. For example, Wu et al. \cite{799r} proposed an evolutionary embedding learning (EEL) paradigm to learn a fast accurate student network via massive knowledge distillation. Their experimental results show that the EEL is able to narrow the performance between the teacher and student networks on given tasks. Zhang et al. \cite{800r} developed an evolutionary knowledge distillation method to improve the effectiveness of knowledge transfer. In this method, an evolutionary teacher was learned online and consistently transfers intermediate knowledge to the student network to narrow the gap of the learning capability between them.
	
	\subsubsection{Low-rank Factorization}
	
	\ 
	
	DNNs often involve in a huge number of weights, which may impact the inference speed and seriously increase the storage overhead of the DNN. The weights can be viewed as a matrix $W$ with $m$ $\times$ $n$ dimensions. The low-rank approach is commonly applied to the weight matrix ($W$) after the DNN is fully trained. For example, singular value decomposition \cite{801r} is a typical low-rank factorization method, where $W$ is decomposed as follows.
	\begin{equation}\label{Eq9}
	\begin{aligned}
	\begin{matrix}
	W $=$ USV^T
	\end{matrix}  \\
	\end{aligned}
	\end{equation}
	\noindent where $U$ $\in$ $R^{m \times m}$, $V^T$ $\in$ $R^{n \times n}$ are orthogonal matrices and  $S$ $\in$ $R^{m \times n}$ is a diagonal matrix. 
	
	Notably, most of the existing low-rank factorization methods rely on domain expertise and experience for the selection of hyperparameters (e.g., the rank and sparsity of weight matrix) to get an appropriate compression results without serious performance degradation \cite{844r, 845r, 846r}. 
	
	Accordingly, EC-based methods have been introduced to solve the above challenge \cite{811r, 802r}. For example, Huang et al. \cite{802r} presented a multi-objective evolution approach to automatically optimize rank and sparsity for weight matrix without human intervention, where two objectives were taken into account including the minimization of the model classification error rate and maximization of the model compression rate. They therefore generated a set of approximately compressed models with different compression rates to mitigate the expensive training process.
	
	\subsubsection{EC for Jointly Optimization}
	
	\ 
	
	Many compression techniques (e.g., quantization) can be easily applied on top of other techniques (e.g., pruning and low-rank factorization). For example, pruning first and then quantification can obtain a lightweight model with faster inference. Similarly, EC can optimize more than one model compression method at the same time.  In the followings, we will briefly review such works.
	
	Phan et al. \cite{803r} designed an efficient 1-Bit CNNs, which combined quantization with a compact model. Specifically, they firstly created a number of strong baseline binary networks (BNNs), which had abundant random group combinations at each convolutional layer. Then, they adopted evolutionary search to seek an optimal group convolution combination with accuracy above threshold. Finally, the obtained binary models werr trained from scratch to achieve the final lightweight network. Different from  \cite{803r}, Polino et al. \cite{804r} jointly utilized weight quantization and distillation to compress large networks (i.e., teacher network) into small networks (i.e., student network), where the latency and model error were regarded as the objectives during the optimization.
	
	Recently, Zhou et al. \cite{805r} developed an evolutionary algorithm-based method for shallowing DNNs at block levels (ESNB). In ESNB, a prior knowledge was extracted from the original model to guide the population initialization. Then, an evolutionary  multi-objective optimization mothed was performed to minimize the number of blocks and the accuracy drop (i.e., loss). After that, knowledge distillation was employed to compensate for the performance degradation via matching output of the pruned model with the softened and hardened output of the original model.
	
	\subsubsection{Summary}
	
	\ 
	
	There is still a big room for the improvement on addressing the huge computational overhead of evolutionary model deployment. Acceleration strategies may be able to alleviate the issue. Besides, there is a high coupling between model deployment and model generation since the performance of the compressed network is strongly dependent on the performance of the original network. The black-box nature of model also hampers deployment in security-critical tasks (e.g., medicine and finance).Consequently, it is promising and challenging to take the model compression, NAS, and interpretability as a single optimization problem and handle it with acceptable time consumption.

	\section{Applications, Open Issues, and Trends} \label{AOT}
	
	\subsection{Applications}
	
	EDL algorithms have been widely used in various real-world applications. In practical, great development has been achieved in computer vision (CV), natural language processing (NLP) and other practical applications (e.g., crisis prediction and disease prediction).  
	
	\subsubsection{Computer Vision}
	
	\ 
	
	CV is an important domain of computer science, playing an important role in identifying useful information (e.g., objects and classifications) for specific tasks (e.g., image segmentation \cite{772r} and object detection \cite{454r}) on images or videos. In the early days,  manually designed models for computer vision achieved good performance on public datasets at the expense of extensive time and labour. With the development of EDL, many new structures have been developed by computer programming and they show better performance than these manually designed models, especially on the widely used benchmark datasets for image classification, such as  CIFAR-10, CIFAR-100 \cite{Krizhevsky2009LearningML}, and ImageNet \cite{Deng2009ImageNetAL}.  For example, the state-of-the-art NAT-M4 \cite{032r} with a small model size  achieves Top-1 accuracy of 80.5\% on ImageNet. Image-to-image processing \cite{037r} (e.g., super-resolution, image inpainting, and image restoration) also received extensive attention from researchers \cite{075r, 570r, 885r}.  Ho et al. \cite{583r} employed NAS techniques to improve image denoising, inpainting, and super-resolution on the foundation of deep image prior \cite{583r}. In addition to the above applications, EDL also has great potential in other areas of CV, such as object detection \cite{774r},  video/picture understanding \cite{701r}, and image segmentation  \cite{598r}.  
	
	\subsubsection{Natural Language Processing}
	
	\ 
	
	Natural language processing (NLP) driven by computer science and computational linguistics, aims to understand, analyze, and extract knowledge on text and speech recoginition \cite{775r}. Many effective NLP models (e.g., GPT-2 \cite{776r} and BERT \cite{777r}) narrow the chasm between human communication and computer understanding using sophisticated mechanisms. Recently, EC-inspired NLP models have been proposed such as language model \cite{904r}, entity recognition \cite{779r}, text classification \cite{890r, z14, z15, z16}, and keyword spotting \cite{893r}. Satapathy et al. \cite{778r} introduced evolutionary multi-objective (i.e., inference time and accuracy) optimization in an English translation system. Sikdar et al. \cite{779r} employed DE in feature selection for named entity recognition (NER).

	\subsubsection{Other Applications}
	
	\

	In addition to CV and NLP, EDL also shows strong ability on handling other practical applications, such as medical analysis \cite{210r, 780r}, financial prediction \cite{769r}, signal processing \cite{z19, z20}, and industrial prediction \cite{267r, 633r}. In particular, Zhu et al. \cite{780r} presented a Markov blanket-embedded genetic algorithm for feature selection to improve gene selection. In \cite{769r}, financial bankruptcy analysis was handled by an evolutionary pruning neural network. The work in \cite{z19} designed a feature selection method based on ACO to classify electromyography signals. For remote sensing imagery, a suitable model was found by multi-objective neural evolution architecture search \cite{633r}, where architecture complexity and performance error of searched network were two conflicting objectives.

	\subsection{Open Issues}
	
	EDL is a hot research topic in both fields of machine learning and evolutionary computation. There are a large number of publications on various top conferences and journals, such as ICCV, CVPR, GECCO, TPAMI, TEVC, TCYB, and TNNLS (see the reference list). Yet some challenges remain to be resolved.
	
	\textbf{Acceleration strategies}: Many EDL approaches suffer from low efficiency due to the expensive evaluations. So various acceleration strategies, such as surrogate model \cite{123r}, supernet \cite{581r}, and early stop \cite{663r} have been designed. However, the improvements of the accuracy are at the expense of sacrifice a bit of model accuracy. Taking the supernet-based inheritance as an example \cite{781r}, we cannot guarantee that every subnet receives a reliable evaluation due to the catastrophic forgetting \cite{783r} and weight coupling \cite{782r}. Therefore, how to balance the efficiency and accuracy needs further investigation.

	\textbf{Effectiveness}: There is a debate on whether EDL has many advantages over other search paradigms (e.g., random search and RL). Some studies argue that many popular search paradigms (e.g., EC-based methods and RL-based methods) have no big difference from the random search methods in their performance, and some random search methods even outperform EC-based methods in some scenarios \cite{785r}. On the contrary,  EC-based approaches have also been proved to be more effective than random search methods in many studies  \cite{122r, 581r,582r}. Thus, a unified platform is essential to measure the effectiveness of different search models, under the consistent search space and hyperparameters configuration \cite{z17}. In addition, elaborate experiments are required to justify the effects of different genetic operators (e.g., crossover operator) to the evolutionary process of EDL.
	
	\textbf{Large-scale datasets}: There is an issue for the studies of EDL on large-scale datasets. It is noted that many studies of EDL are tested on small- and medium-scale datasets such as CIFAR-10 and CIFAR-100 (including 60000 32$\times$32 images), and especially the accuracy on CIFAR-10 reaches up to 98\% \cite{032r}. Although large-scale datasets are ubiquitous and essential in various domains like gene analysis \cite{193r}  and ImageNet \cite{Deng2009ImageNetAL}, computational costs are unaffordable for many researchers as pointed in some statistical reports \cite{025r,037r}. Therefore, the sensitivity of the EDL methods to different scales of datasets is necessary \cite{031r} and how to economically and efficiently verify EDL methods on large-scale datasets also deserves much investigations.  
	
	\textbf{End-to-end EDL}: Originally, AutoML aims to simultaneously optimize feature engineering, model generation and model deployment as a whole. However, there is a strong correlation between them where the performance of next phase heavily relies on the results of the previous phase \cite{813r}. As a result, most studies only focus on parts of the EDL pipeline (Fig. \ref{fig1}). For instance, TPOT \cite{789r} is designed on top of Pytorch for building classification tasks, which however only supports a multi-layer perception machine (i.e., model generation). There are many partially accomplished end-to-end for EDL, such as ModelArts (model generation), Google's Cloud (NAS), and Feature Labs (feature engineering) \cite{024r}, to name but a few. The main reason is that the optimization of the whole EDL pipeline may need huge computational cost not only on the exploration of the large-scale search space but also on handling highly-coupled relation between different parts of EDL. Consequently, finding an optimal solution of the complete EDL pipeline is essential but challenging.
	
	\subsection{Challenges and Future Trends} 
	
	\ 
	
	Although remarkable progress has been made in EDL, there are still many promising lines of research.
	
	\textbf{Fair comparisons}: Unfair comparisons of different EDL methods are easily encountered with the following reasons. Firstly, uniform benchmarks are essential. In feature engineering, no uniform benchmark is for the fair comparison of different algorithms due to different downstream prediction models and feature sets. Secondly, there is no uniform criterion for different methods in handling NAS and model compression by using different tricks (e.g., cutout \cite{790r} and ScheduledDropPath \cite{027r}), which may influence the performance of the final architecture. Thirdly, a fair platform for EDL is essential. There are some fair benchmarks but only for specific tasks, such as BenchENAS \cite{z17}, NAS-Bench-101 \cite{Ying2019NASBench101TR}, NAS-Bench-201 \cite{786r}, NAS-Bench-301 \cite{793r}, and HW-NAS-Bench \cite{062r}. 
	
	\textbf{Interpretability}: EDL is known as a black-box optimization, and there is a lack of theoretical analysis to explain its superiority \cite{z18}. For example, it is difficult to explain why EC-based method tends to select features contribute to the performance of the classification model in feature engineering. As a result, the development of EDL in some sensitive domains such as financial and medical fields is slow. To overcome this issue, Evans et al. \cite{795r} used visualization to expound how the evolved convolution filter served and indirectly explained the search process of the model. Nevertheless, some studies argue that the explanation for these occurrences is usually post-hoc and lacks trustworthy mathematical deduction \cite{037r, 044r}. Thus, the interpretability of EDL is an interesting and promising research direction.

	\textbf{Exploring more scenarios}: There is still plenty of room for the improvement of the performance of EDL on both benchmarks and real-world applications. Although EDL methods outperform manually designed models on various image benchmarks (CIFAR-10 and ImageNet), the state-of-the-art EDL methods \cite{798r} lost their advantages on NLP in comparison with human-designed models (e.g.,  GPT-2 \cite{776r}, Transformer-XL \cite{796r}). In comparison with the benchmarks, it is more difficult to handle real-world tasks, which inevitably contain noise (e.g., mislabeling and inadequate or imbalance data) or may have small-scale datasets (leading to overfitting). Hence, some techniques such as unsupervised and self-supervised learning may be incorporated into EDL to mitigate these types of issues. 
	
	\section{Conclusions}\label{Con}
	
	With the development of machine leaning and evolutionary computation, many EDL approaches have been proposed to automatically optimize the parameters or architectures of deep models following the EC optimization framework. EDL approaches show competitive performance in robust and search capability, in comparison with the manually designed approaches. Therefore, EDL has become a hot research topic. 
	
	In this survey, we first introduced EDL from the perspective of DL and EC to facilitate the understanding of readers from the communities of ML and EC. Then we formulated EDL as a complex optimization problem, and provided a comprehensive survey of EC techniques in solving EDL optimization problems in terms of feature engineering, model generation to model deployment to form a new taxonomy (i.e., what, where and how to evolve/optimize in EDL). Specifiically, we discussed the solution representations and search paradigms of EDL at different stages of its pipeline in detail. Then the pros and cons of EC-based approaches in comparison to non-EC based ones are discussed.  Subsequently, various applications are summarized to show the potential ability of EDL in handling real-world problems. 
	
	Although EDL approaches have achieved great progress in AutoML, there are still a number of challenging issues to be resolved. For example, effective acceleration strategies are essential to reduce the expensive optimization process. Another issue is to handle large-scale datasets and how to perform fair comparisons between different EDL approaches or non-EC based methods. More investigations are required to theoretically analyse or interpret the search ability of EDL. In addition, a lot of efforts are required on the improving the performance of EDL on both benchmarks (e.g., large-scale and small-scale data) and real-world applications. Lastly, the development of end-to-end EDL is challenging but deserves much efforts.

	\bibliographystyle{ACM-Reference-Format}
	\bibliography{sample-base}


\begin{thebibliography}{264}

%%% the Web address.

\ifx \showCODEN    \undefined \def \showCODEN     #1{\unskip}     \fi
\ifx \showDOI      \undefined \def \showDOI       #1{#1}\fi
\ifx \showISBNx    \undefined \def \showISBNx     #1{\unskip}     \fi
\ifx \showISBNxiii \undefined \def \showISBNxiii  #1{\unskip}     \fi
\ifx \showISSN     \undefined \def \showISSN      #1{\unskip}     \fi
\ifx \showLCCN     \undefined \def \showLCCN      #1{\unskip}     \fi
\ifx \shownote     \undefined \def \shownote      #1{#1}          \fi
\ifx \showarticletitle \undefined \def \showarticletitle #1{#1}   \fi
\ifx \showURL      \undefined \def \showURL       {\relax}        \fi
% invisible to TeX
\providecommand\bibfield[2]{#2}
\providecommand\bibinfo[2]{#2}
\providecommand\natexlab[1]{#1}
\providecommand\showeprint[2][]{arXiv:#2}

\bibitem[Abdi and Williams(2010)]%
        {411r}
\bibfield{author}{\bibinfo{person}{Herv{\'e} Abdi} {and}
  \bibinfo{person}{Lynne~J Williams}.} \bibinfo{year}{2010}\natexlab{}.
\newblock \showarticletitle{Principal Component Analysis}.
\newblock \bibinfo{journal}{\emph{Comput. Stat.}} \bibinfo{volume}{2},
  \bibinfo{number}{4} (\bibinfo{year}{2010}), \bibinfo{pages}{433--459}.
\newblock
\showISSN{1939-5108}


\bibitem[Ahmed et~al\mbox{.}(2019)]%
        {548r}
\bibfield{author}{\bibinfo{person}{Amr Ahmed}, \bibinfo{person}{Saad~Mohamed
  Darwish}, {and} \bibinfo{person}{Mohamed~M. El-Sherbiny}.}
  \bibinfo{year}{2019}\natexlab{}.
\newblock \showarticletitle{A Novel Automatic {CNN} Architecture Design
  Approach Based on Genetic Algorithm}. In \bibinfo{booktitle}{\emph{Int. Conf.
  Adv. Intell. Syst. Inform.}} \bibinfo{pages}{473--482}.
\newblock
\showISSN{2194-5357}


\bibitem[Ahmed et~al\mbox{.}(2014)]%
        {380r}
\bibfield{author}{\bibinfo{person}{Soha Ahmed}, \bibinfo{person}{Mengjie
  Zhang}, \bibinfo{person}{Lifeng Peng}, {and} \bibinfo{person}{Bing Xue}.}
  \bibinfo{year}{2014}\natexlab{}.
\newblock \showarticletitle{Multiple Feature Construction for Effective
  Biomarker Identification and Classification Using Genetic Programming}. In
  \bibinfo{booktitle}{\emph{Proc. Genetic Evol. Comput. Conf.}}
  \bibinfo{pages}{249--256}.
\newblock


\bibitem[Al-kazemi and Mohan(2002)]%
        {481r}
\bibfield{author}{\bibinfo{person}{Buthainah Al-kazemi} {and}
  \bibinfo{person}{Chilukuri~Krishna Mohan}.} \bibinfo{year}{2002}\natexlab{}.
\newblock \showarticletitle{Training Feedforward Neural Networks using
  Nulti-phase Particle Swarm Optimization}. In \bibinfo{booktitle}{\emph{Proc.
  Int. Conf. Neural Inf. Process.}}, Vol.~\bibinfo{volume}{5}.
  \bibinfo{pages}{2615--2619}.
\newblock


\bibitem[Al-Sahaf et~al\mbox{.}(2019)]%
        {044r}
\bibfield{author}{\bibinfo{person}{Harith Al-Sahaf}, \bibinfo{person}{Ying Bi},
  \bibinfo{person}{Qi Chen}, \bibinfo{person}{Andrew Lensen},
  \bibinfo{person}{Yi Mei}, \bibinfo{person}{Yanan Sun}, \bibinfo{person}{Binh
  Tran}, \bibinfo{person}{Bing Xue}, {and} \bibinfo{person}{Mengjie Zhang}.}
  \bibinfo{year}{2019}\natexlab{}.
\newblock \showarticletitle{A Survey on Evolutionary Machine Learning}.
\newblock \bibinfo{journal}{\emph{J. R. Soc. N. Z.}} \bibinfo{volume}{49},
  \bibinfo{number}{2} (\bibinfo{year}{2019}), \bibinfo{pages}{205--228}.
\newblock
\showISSN{0303-6758}


\bibitem[Albukhanajer et~al\mbox{.}(2015)]%
        {415r}
\bibfield{author}{\bibinfo{person}{Wissam~A. Albukhanajer},
  \bibinfo{person}{Johann~A. Briffa}, {and} \bibinfo{person}{Yaochu Jin}.}
  \bibinfo{year}{2015}\natexlab{}.
\newblock \showarticletitle{Evolutionary Multiobjective Image Feature
  Extraction in the Presence of Noise}.
\newblock \bibinfo{journal}{\emph{IEEE Trans. Cybern.}} \bibinfo{volume}{45},
  \bibinfo{number}{9} (\bibinfo{year}{2015}), \bibinfo{pages}{1757--1768}.
\newblock


\bibitem[Alexandropoulos and Aridas(2019)]%
        {047r}
\bibfield{author}{\bibinfo{person}{Stamatios-Aggelos~N Alexandropoulos} {and}
  \bibinfo{person}{Christos~K Aridas}.} \bibinfo{year}{2019}\natexlab{}.
\newblock \showarticletitle{Multi-objective Evolutionary Optimization
  Algorithms for Machine Learning: A Recent Survey}.
\newblock \bibinfo{journal}{\emph{Approximation and Optimization}}
  (\bibinfo{year}{2019}), \bibinfo{pages}{35--55}.
\newblock
\showISSN{1931-6828}


\bibitem[Aljarah et~al\mbox{.}(2018)]%
        {696r}
\bibfield{author}{\bibinfo{person}{Ibrahim Aljarah}, \bibinfo{person}{Hossam
  Faris}, {and} \bibinfo{person}{Seyed~Mohammad Mirjalili}.}
  \bibinfo{year}{2018}\natexlab{}.
\newblock \showarticletitle{Optimizing Connection Weights in Neural Networks
  Using the Whale Optimization Algorithm}.
\newblock \bibinfo{journal}{\emph{Soft Comput.}} \bibinfo{volume}{22},
  \bibinfo{number}{1} (\bibinfo{year}{2018}), \bibinfo{pages}{1--15}.
\newblock
\showISSN{1868-8071}


\bibitem[Andersen et~al\mbox{.}(2021)]%
        {z14}
\bibfield{author}{\bibinfo{person}{Hayden Andersen}, \bibinfo{person}{Sean
  Stevenson}, \bibinfo{person}{Tuan Ha}, \bibinfo{person}{Xiaoying Gao}, {and}
  \bibinfo{person}{Bing Xue}.} \bibinfo{year}{2021}\natexlab{}.
\newblock \showarticletitle{Evolving Neural Networks for Text Classification
  Using Genetic Algorithm-based Approaches}. In \bibinfo{booktitle}{\emph{Proc.
  IEEE Congr. Evol. Comput.}} \bibinfo{pages}{1241--1248}.
\newblock


\bibitem[Assun{\c{c}}{\~a}o et~al\mbox{.}(2019a)]%
        {146r}
\bibfield{author}{\bibinfo{person}{Filipe Assun{\c{c}}{\~a}o},
  \bibinfo{person}{Joao Correia}, {and} \bibinfo{person}{R{\'u}ben
  Concei{\c{c}}{\~a}o}.} \bibinfo{year}{2019}\natexlab{a}.
\newblock \showarticletitle{Automatic Design of Artificial Neural Networks for
  Gamma-Ray Detection}.
\newblock \bibinfo{journal}{\emph{IEEE Access}}  \bibinfo{volume}{7}
  (\bibinfo{year}{2019}), \bibinfo{pages}{110531--110540}.
\newblock
\showISSN{2169-3536}


\bibitem[Assun{\c{c}}{\~a}o et~al\mbox{.}(2018)]%
        {675r}
\bibfield{author}{\bibinfo{person}{Filipe Assun{\c{c}}{\~a}o},
  \bibinfo{person}{Nuno Louren{\c{c}}o}, \bibinfo{person}{P. Machado}, {and}
  \bibinfo{person}{Bernardete Ribeiro}.} \bibinfo{year}{2018}\natexlab{}.
\newblock \showarticletitle{Evolving the Topology of Large Scale Deep Neural
  Networks}. In \bibinfo{booktitle}{\emph{Proc. Eur. Conf. Genetic Program}}.
  \bibinfo{pages}{19--34}.
\newblock
\showISSN{0302-9743}


\bibitem[Assun{\c{c}}{\~a}o et~al\mbox{.}(2019b)]%
        {664r}
\bibfield{author}{\bibinfo{person}{Filipe Assun{\c{c}}{\~a}o},
  \bibinfo{person}{Nuno Louren{\c{c}}o}, \bibinfo{person}{Penousal Machado},
  {and} \bibinfo{person}{Bernardete Ribeiro}.}
  \bibinfo{year}{2019}\natexlab{b}.
\newblock \showarticletitle{Fast denser: Efficient Deep Neuroevolution}. In
  \bibinfo{booktitle}{\emph{Proc. Eur. Conf. Genetic Program}}.
  \bibinfo{pages}{197--212}.
\newblock


\bibitem[Atre et~al\mbox{.}(2021)]%
        {680r}
\bibfield{author}{\bibinfo{person}{Medha Atre}, \bibinfo{person}{Birendra Jha},
  {and} \bibinfo{person}{Ashwini Rao}.} \bibinfo{year}{2021}\natexlab{}.
\newblock \showarticletitle{Distributed Deep Learning Using Volunteer
  Computing-Like Paradigm}. In \bibinfo{booktitle}{\emph{Proc. Int. Parallel
  and Distrib. Process. Symp.}} \bibinfo{pages}{933--942}.
\newblock


\bibitem[Barman and Kwon(2020)]%
        {705r}
\bibfield{author}{\bibinfo{person}{Shohag Barman} {and}
  \bibinfo{person}{Yung-Keun Kwon}.} \bibinfo{year}{2020}\natexlab{}.
\newblock \showarticletitle{A Neuro-Evolution Approach to Infer A Boolean
  Network From Time-Series Gene Expressions}.
\newblock \bibinfo{journal}{\emph{Bioinformatics}} \bibinfo{volume}{36},
  \bibinfo{number}{2} (\bibinfo{year}{2020}), \bibinfo{pages}{i762--i769}.
\newblock
\showISSN{1367-4803}


\bibitem[Behjat and Chidambaran(2019)]%
        {707r}
\bibfield{author}{\bibinfo{person}{Amir Behjat} {and} \bibinfo{person}{Sharat
  Chidambaran}.} \bibinfo{year}{2019}\natexlab{}.
\newblock \showarticletitle{Adaptive Genomic Evolution of Neural Network
  Topologies ({AGENT}) for State-to-Action Mapping in Autonomous Agents}. In
  \bibinfo{booktitle}{\emph{Proc. Int. Conf. Robot. Autom.}}
  \bibinfo{pages}{9638--9644}.
\newblock


\bibitem[Bhanu and Krawiec(2002)]%
        {367r}
\bibfield{author}{\bibinfo{person}{Bir Bhanu} {and} \bibinfo{person}{Krzysztof
  Krawiec}.} \bibinfo{year}{2002}\natexlab{}.
\newblock \showarticletitle{Coevolutionary Construction of Features for
  Transformation of Representation in Machine Learning}. In
  \bibinfo{booktitle}{\emph{Proc. Genetic Evol. Comput. Conf.}}
  \bibinfo{pages}{249--254}.
\newblock


\bibitem[Bi et~al\mbox{.}(2018)]%
        {436r}
\bibfield{author}{\bibinfo{person}{Ying Bi}, \bibinfo{person}{Bing Xue}, {and}
  \bibinfo{person}{Mengjie Zhang}.} \bibinfo{year}{2018}\natexlab{}.
\newblock \showarticletitle{An Automatic Feature Extraction Approach to Image
  Classification Using Genetic Programming}. In \bibinfo{booktitle}{\emph{Proc.
  Int. Conf. Appl. Evol. Comput.}} \bibinfo{pages}{421--438}.
\newblock


\bibitem[Cano et~al\mbox{.}(2017)]%
        {442r}
\bibfield{author}{\bibinfo{person}{Alberto Cano},
  \bibinfo{person}{Sebasti{\'a}n Ventura}, {and} \bibinfo{person}{Krzysztof~J.
  Cios}.} \bibinfo{year}{2017}\natexlab{}.
\newblock \showarticletitle{Multi-Objective Genetic Programming for Feature
  Extraction and Data Visualization}.
\newblock \bibinfo{journal}{\emph{Soft Comput.}} \bibinfo{volume}{21},
  \bibinfo{number}{8} (\bibinfo{year}{2017}), \bibinfo{pages}{2069--2089}.
\newblock
\showISSN{0950-7051}


\bibitem[Castelli et~al\mbox{.}(2011)]%
        {382r}
\bibfield{author}{\bibinfo{person}{Mauro Castelli}, \bibinfo{person}{Luca
  Manzoni}, {and} \bibinfo{person}{Leonardo Vanneschi}.}
  \bibinfo{year}{2011}\natexlab{}.
\newblock \showarticletitle{Multi Objective Genetic Programming for Feature
  Construction in Classification Problems}. In \bibinfo{booktitle}{\emph{Proc.
  Int. Conf. Learn. Intell. Optim.}} \bibinfo{pages}{503--506}.
\newblock


\bibitem[Chai et~al\mbox{.}(2022)]%
        {755r}
\bibfield{author}{\bibinfo{person}{Zheng-Yi Chai}, \bibinfo{person}{ChuanHua
  Yang}, {and} \bibinfo{person}{Ya-Lun Li}.} \bibinfo{year}{2022}\natexlab{}.
\newblock \showarticletitle{Communication Efficiency Optimization in Federated
  Learning Based on Multi-Objective Evolutionary Algorithm}.
\newblock \bibinfo{journal}{\emph{Evol. Intell.}} (\bibinfo{year}{2022}),
  \bibinfo{pages}{1--12}.
\newblock
\showISSN{1864-5909}


\bibitem[Chandra(2015)]%
        {493r}
\bibfield{author}{\bibinfo{person}{Rohitash Chandra}.}
  \bibinfo{year}{2015}\natexlab{}.
\newblock \showarticletitle{Competition and Collaboration in Cooperative
  Coevolution of Elman Recurrent Neural Networks for Time-Series Prediction}.
\newblock \bibinfo{journal}{\emph{IEEE Trans. Neural Netw. Learn. Syst.}}
  \bibinfo{volume}{26}, \bibinfo{number}{12} (\bibinfo{year}{2015}),
  \bibinfo{pages}{3123--3136}.
\newblock


\bibitem[Chandra and Zhang(2012)]%
        {492r}
\bibfield{author}{\bibinfo{person}{Rohitash Chandra} {and}
  \bibinfo{person}{Mengjie Zhang}.} \bibinfo{year}{2012}\natexlab{}.
\newblock \showarticletitle{Cooperative Coevolution of Elman Recurrent Neural
  Networks for Chaotic Time Series Prediction}.
\newblock \bibinfo{journal}{\emph{Neurocomputing}}  \bibinfo{volume}{86}
  (\bibinfo{year}{2012}), \bibinfo{pages}{116--123}.
\newblock
\showISSN{0925-2312}


\bibitem[Chen et~al\mbox{.}(2022)]%
        {208r}
\bibfield{author}{\bibinfo{person}{Ke Chen}, \bibinfo{person}{Bing Xue},
  \bibinfo{person}{Mengjie Zhang}, {and} \bibinfo{person}{Fengyu Zhou}.}
  \bibinfo{year}{2022}\natexlab{}.
\newblock \showarticletitle{Evolutionary Multitasking for Feature Selection in
  High-Dimensional Classification via Particle Swarm Optimization}.
\newblock \bibinfo{journal}{\emph{IEEE Trans. Evol. Comput.}}
  \bibinfo{volume}{26}, \bibinfo{number}{3} (\bibinfo{year}{2022}),
  \bibinfo{pages}{446--460}.
\newblock


\bibitem[Chen et~al\mbox{.}(2015)]%
        {z07}
\bibfield{author}{\bibinfo{person}{Qi Chen}, \bibinfo{person}{Bing Xue}, {and}
  \bibinfo{person}{Mengjie Zhang}.} \bibinfo{year}{2015}\natexlab{}.
\newblock \showarticletitle{Generalisation and Domain Adaptation in {GP} with
  Gradient Descent for Symbolic Regression}. In \bibinfo{booktitle}{\emph{Proc.
  IEEE Congr. Evol. Comput.}} \bibinfo{pages}{1137--1144}.
\newblock


\bibitem[Chen et~al\mbox{.}(2019a)]%
        {823r}
\bibfield{author}{\bibinfo{person}{Shuxin Chen}, \bibinfo{person}{Lin Lin},
  \bibinfo{person}{Zixun Zhang}, {and} \bibinfo{person}{Mitsuo Gen}.}
  \bibinfo{year}{2019}\natexlab{a}.
\newblock \showarticletitle{Evolutionary NetArchitecture Search for Deep Neural
  Networks Pruning}. In \bibinfo{booktitle}{\emph{Proc. Aust. Conf. Artif.
  Intell.}} \bibinfo{pages}{189--196}.
\newblock


\bibitem[Chen et~al\mbox{.}(2020)]%
        {152r}
\bibfield{author}{\bibinfo{person}{Xiangru Chen}, \bibinfo{person}{Yanan Sun},
  \bibinfo{person}{Mengjie Zhang}, {and} \bibinfo{person}{Dezhong Peng}.}
  \bibinfo{year}{2020}\natexlab{}.
\newblock \showarticletitle{Evolving Deep Convolutional Variational
  Autoencoders for Image Classification}.
\newblock \bibinfo{journal}{\emph{IEEE Trans. Evol. Comput.}}
  \bibinfo{volume}{25}, \bibinfo{number}{5} (\bibinfo{year}{2020}),
  \bibinfo{pages}{815--829}.
\newblock
\showISSN{1089-778X}


\bibitem[Chen et~al\mbox{.}(2019b)]%
        {065r}
\bibfield{author}{\bibinfo{person}{Yukang Chen}, \bibinfo{person}{Gaofeng
  Meng}, \bibinfo{person}{Qian Zhang}, \bibinfo{person}{Shiming Xiang}, {and}
  \bibinfo{person}{Chang Huang}.} \bibinfo{year}{2019}\natexlab{b}.
\newblock \showarticletitle{{RENAS}: Reinforced Evolutionary Neural
  Architecture Search}. In \bibinfo{booktitle}{\emph{Proc. IEEE Conf. Comput.
  Vis. Pattern Recognit.}} \bibinfo{pages}{4787--4796}.
\newblock


\bibitem[Cheng et~al\mbox{.}(2021)]%
        {204r}
\bibfield{author}{\bibinfo{person}{Fan Cheng}, \bibinfo{person}{Feixiang Chu},
  \bibinfo{person}{Yi Xu}, {and} \bibinfo{person}{Lei Zhang}.}
  \bibinfo{year}{2021}\natexlab{}.
\newblock \showarticletitle{A Steering-Matrix-Based Multiobjective Evolutionary
  Algorithm for High-Dimensional Feature Selection}.
\newblock \bibinfo{journal}{\emph{IEEE Trans. Cybern.}} \bibinfo{volume}{52},
  \bibinfo{number}{9} (\bibinfo{year}{2021}), \bibinfo{pages}{9695--9708}.
\newblock


\bibitem[Cheng et~al\mbox{.}(2017)]%
        {043r}
\bibfield{author}{\bibinfo{person}{Yu Cheng}, \bibinfo{person}{Duo Wang},
  \bibinfo{person}{Pan Zhou}, {and} \bibinfo{person}{Tao Zhang}.}
  \bibinfo{year}{2017}\natexlab{}.
\newblock \showarticletitle{A Survey of Model Compression and Acceleration for
  Deep Neural networks}.
\newblock \bibinfo{journal}{\emph{arXiv preprint arXiv:1710.09282}}
  (\bibinfo{year}{2017}).
\newblock


\bibitem[Chiba et~al\mbox{.}(2019)]%
        {679r}
\bibfield{author}{\bibinfo{person}{Zouhair Chiba}, \bibinfo{person}{Noreddine
  Abghour}, \bibinfo{person}{Khalid Moussaid}, \bibinfo{person}{Amina~El Omri},
  {and} \bibinfo{person}{Mohamed Rida}.} \bibinfo{year}{2019}\natexlab{}.
\newblock \showarticletitle{Intelligent Approach to Build a Deep Neural Network
  Based {IDS} for Cloud Environment Using Combination of Machine Learning
  Algorithms}.
\newblock \bibinfo{journal}{\emph{Comput. \& Sec.}}  \bibinfo{volume}{86}
  (\bibinfo{year}{2019}), \bibinfo{pages}{291--317}.
\newblock
\showISSN{0167-4048}


\bibitem[Choudhary et~al\mbox{.}(2020)]%
        {042r}
\bibfield{author}{\bibinfo{person}{Tejalal Choudhary}, \bibinfo{person}{Vipul
  Mishra}, \bibinfo{person}{Anurag Goswami}, {and} \bibinfo{person}{Jagannathan
  Sarangapani}.} \bibinfo{year}{2020}\natexlab{}.
\newblock \showarticletitle{A Comprehensive Survey on Model Compression and
  Acceleration}.
\newblock \bibinfo{journal}{\emph{Artif. Intell. Rev.}} \bibinfo{volume}{53},
  \bibinfo{number}{7} (\bibinfo{year}{2020}), \bibinfo{pages}{5113--5155}.
\newblock
\showISSN{1573-7462}


\bibitem[Chrabaszcz et~al\mbox{.}(2017)]%
        {676r}
\bibfield{author}{\bibinfo{person}{Patryk Chrabaszcz}, \bibinfo{person}{Ilya
  Loshchilov}, {and} \bibinfo{person}{Frank Hutter}.}
  \bibinfo{year}{2017}\natexlab{}.
\newblock \showarticletitle{A Downsampled variant of image{N}et as an
  alternative to the Cifar datasets}.
\newblock \bibinfo{journal}{\emph{arXiv preprint arXiv:1707.08819}}
  (\bibinfo{year}{2017}).
\newblock


\bibitem[Chu et~al\mbox{.}(2020)]%
        {608r}
\bibfield{author}{\bibinfo{person}{Xiangxiang Chu}, \bibinfo{person}{Bo Zhang},
  \bibinfo{person}{Ruijun Xu}, {and} \bibinfo{person}{Hailong Ma}.}
  \bibinfo{year}{2020}\natexlab{}.
\newblock \showarticletitle{Multi-Objective Reinforced Evolution in Mobile
  Neural Architecture Search}. In \bibinfo{booktitle}{\emph{Proc. Eur. Conf.
  Comput. Vis.}} \bibinfo{pages}{99--113}.
\newblock
\showISSN{0302-9743}


\bibitem[Conti et~al\mbox{.}(2018)]%
        {535r}
\bibfield{author}{\bibinfo{person}{Edoardo Conti}, \bibinfo{person}{Vashisht
  Madhavan}, \bibinfo{person}{Felipe~Petroski Such}, \bibinfo{person}{Joel
  Lehman}, \bibinfo{person}{Kenneth~O. Stanley}, {and} \bibinfo{person}{Jeff
  Clune}.} \bibinfo{year}{2018}\natexlab{}.
\newblock \showarticletitle{Improving Exploration in Evolution Strategies for
  Deep Reinforcement Learning via a Population of Novelty-Seeking Agents}. In
  \bibinfo{booktitle}{\emph{Proc. Adv. Neural Inf. Process. Syst.}},
  Vol.~\bibinfo{volume}{31}. \bibinfo{pages}{5032--5043}.
\newblock


\bibitem[Cui et~al\mbox{.}(2018)]%
        {514r}
\bibfield{author}{\bibinfo{person}{Xiaodong Cui}, \bibinfo{person}{Wei Zhang},
  \bibinfo{person}{Zolt{\'a}n T{\"u}ske}, {and} \bibinfo{person}{Michael
  Picheny}.} \bibinfo{year}{2018}\natexlab{}.
\newblock \showarticletitle{Evolutionary Stochastic Gradient Descent for
  Optimization of Deep Neural Networks}.
\newblock \bibinfo{journal}{\emph{Proc. Adv. Neural Inf. Process. Syst.}}
  \bibinfo{volume}{31} (\bibinfo{year}{2018}), \bibinfo{pages}{6051--6061}.
\newblock


\bibitem[Da~Silva and Neto(2011)]%
        {194r}
\bibfield{author}{\bibinfo{person}{S{\'e}rgio~Francisco Da~Silva} {and}
  \bibinfo{person}{Jo{\~a}o do ES~Batista Neto}.}
  \bibinfo{year}{2011}\natexlab{}.
\newblock \showarticletitle{Improving The Ranking Quality of Medical Image
  Retrieval Using A Genetic Feature Selection Method}.
\newblock \bibinfo{journal}{\emph{Decis. Support. Syst.}} \bibinfo{volume}{51},
  \bibinfo{number}{4} (\bibinfo{year}{2011}), \bibinfo{pages}{810--820}.
\newblock
\showISSN{0167-9236}


\bibitem[Dahal and Zhan(2020)]%
        {559r}
\bibfield{author}{\bibinfo{person}{Binay Dahal} {and}
  \bibinfo{person}{Justin~Zhijun Zhan}.} \bibinfo{year}{2020}\natexlab{}.
\newblock \showarticletitle{Effective Mutation and Recombination for Evolving
  Convolutional Networks}. In \bibinfo{booktitle}{\emph{Proc. Adv. Neural Inf.
  Process. Syst.}} \bibinfo{pages}{1--6}.
\newblock


\bibitem[Dai et~al\mbox{.}(2019)]%
        {796r}
\bibfield{author}{\bibinfo{person}{Zihang Dai}, \bibinfo{person}{Zhilin Yang},
  \bibinfo{person}{Yiming Yang}, \bibinfo{person}{Jaime~G Carbonell},
  \bibinfo{person}{Quoc Le}, {and} \bibinfo{person}{Ruslan Salakhutdinov}.}
  \bibinfo{year}{2019}\natexlab{}.
\newblock \showarticletitle{Transformer-{XL}: Attentive Language Models beyond
  a Fixed-Length Context}. In \bibinfo{booktitle}{\emph{Proc. Assoc. Comput.
  Linguist.}} \bibinfo{pages}{2978--2988}.
\newblock


\bibitem[D'Ambrosio and Stanley(2007)]%
        {756r}
\bibfield{author}{\bibinfo{person}{David~B. D'Ambrosio} {and}
  \bibinfo{person}{Kenneth~O. Stanley}.} \bibinfo{year}{2007}\natexlab{}.
\newblock \showarticletitle{A Novel Generative Encoding for Exploiting Neural
  Network Sensor and Output Geometry}. In \bibinfo{booktitle}{\emph{Proc.
  Genetic Evol. Comput. Conf.}} \bibinfo{pages}{974--981}.
\newblock


\bibitem[Darwish et~al\mbox{.}(2020)]%
        {050r}
\bibfield{author}{\bibinfo{person}{Ashraf Darwish}, \bibinfo{person}{Aboul~Ella
  Hassanien}, {and} \bibinfo{person}{Swagatam Das}.}
  \bibinfo{year}{2020}\natexlab{}.
\newblock \showarticletitle{A Survey of Swarm And Evolutionary Computing
  Approaches for Deep Learning}.
\newblock \bibinfo{journal}{\emph{Artif. Intell. Rev.}} \bibinfo{volume}{53},
  \bibinfo{number}{3} (\bibinfo{year}{2020}), \bibinfo{pages}{1767--1812}.
\newblock
\showISSN{2210-6502}


\bibitem[Deb et~al\mbox{.}(2002)]%
        {773r}
\bibfield{author}{\bibinfo{person}{Kalyanmoy Deb}, \bibinfo{person}{Samir
  Agrawal}, \bibinfo{person}{Amrit Pratap}, {and} \bibinfo{person}{T.
  Meyarivan}.} \bibinfo{year}{2002}\natexlab{}.
\newblock \showarticletitle{A Fast and Elitist Multiobjective Genetic
  Algorithm: {NSGA-II}}.
\newblock \bibinfo{journal}{\emph{IEEE Trans. Evol. Comput.}}
  \bibinfo{volume}{6}, \bibinfo{number}{2} (\bibinfo{year}{2002}),
  \bibinfo{pages}{182--197}.
\newblock
\showISSN{1089-778X}


\bibitem[Demirkir and Sankur(2006)]%
        {774r}
\bibfield{author}{\bibinfo{person}{Cem Demirkir} {and}
  \bibinfo{person}{B{\"u}lent Sankur}.} \bibinfo{year}{2006}\natexlab{}.
\newblock \showarticletitle{Object Detection Using Haar Feature Selection
  Optimization}. In \bibinfo{booktitle}{\emph{Proc. IEEE Signal Process.
  Commun. Appl.}} \bibinfo{pages}{1--4}.
\newblock


\bibitem[Deng et~al\mbox{.}(2009)]%
        {Deng2009ImageNetAL}
\bibfield{author}{\bibinfo{person}{Jia Deng}, \bibinfo{person}{Wei Dong},
  \bibinfo{person}{Richard Socher}, \bibinfo{person}{Li-Jia Li},
  \bibinfo{person}{K. Li}, {and} \bibinfo{person}{Li Fei-Fei}.}
  \bibinfo{year}{2009}\natexlab{}.
\newblock \showarticletitle{Image{N}et: A Large-scale Hierarchical Image
  Database}.
\newblock \bibinfo{journal}{\emph{Proc. IEEE Conf. Comput. Vis. Pattern
  Recognit.}} (\bibinfo{year}{2009}), \bibinfo{pages}{248--255}.
\newblock


\bibitem[Devlin et~al\mbox{.}(2018)]%
        {777r}
\bibfield{author}{\bibinfo{person}{Jacob Devlin}, \bibinfo{person}{Ming-Wei
  Chang}, \bibinfo{person}{Kenton Lee}, {and} \bibinfo{person}{Kristina
  Toutanova}.} \bibinfo{year}{2018}\natexlab{}.
\newblock \showarticletitle{Bert: Pre-training of Deep Bidirectional
  Transformers for Language Understanding}.
\newblock \bibinfo{journal}{\emph{arXiv preprint arXiv:1810.04805}}
  (\bibinfo{year}{2018}).
\newblock


\bibitem[DeVries and Taylor(2017)]%
        {790r}
\bibfield{author}{\bibinfo{person}{Terrance DeVries} {and}
  \bibinfo{person}{Graham~W Taylor}.} \bibinfo{year}{2017}\natexlab{}.
\newblock \showarticletitle{Improved Regularization of Convolutional Neural
  Networks with Cutout}.
\newblock \bibinfo{journal}{\emph{arXiv preprint arXiv:1708.04552}}
  (\bibinfo{year}{2017}).
\newblock


\bibitem[Dong and Yang(2020)]%
        {786r}
\bibfield{author}{\bibinfo{person}{Xuanyi Dong} {and} \bibinfo{person}{Yi
  Yang}.} \bibinfo{year}{2020}\natexlab{}.
\newblock \showarticletitle{{NAS}-Bench-201: Extending the Scope of
  Reproducible Neural Architecture Search}. In \bibinfo{booktitle}{\emph{Proc.
  Int. Conf. Learn. Represent.}}
  \bibinfo{pages}{https://arxiv.org/abs/2001.00326}.
\newblock


\bibitem[Dowdell and Zhang(2020)]%
        {007r}
\bibfield{author}{\bibinfo{person}{Thomas Dowdell} {and}
  \bibinfo{person}{Hongyu Zhang}.} \bibinfo{year}{2020}\natexlab{}.
\newblock \showarticletitle{Language Modelling for Source Code with
  {T}ransformer-{XL}}.
\newblock \bibinfo{journal}{\emph{arXiv preprint arXiv:2007.15813}}
  (\bibinfo{year}{2020}).
\newblock


\bibitem[Elsken et~al\mbox{.}(2017)]%
        {651r}
\bibfield{author}{\bibinfo{person}{Thomas Elsken}, \bibinfo{person}{Jan-Hendrik
  Metzen}, {and} \bibinfo{person}{Frank Hutter}.}
  \bibinfo{year}{2017}\natexlab{}.
\newblock \showarticletitle{Simple and Efficient Architecture Search for
  Convolutional Neural Networks}.
\newblock \bibinfo{journal}{\emph{arXiv preprint arXiv:1711.04528}}
  (\bibinfo{year}{2017}).
\newblock


\bibitem[Elsken et~al\mbox{.}(2019)]%
        {605r}
\bibfield{author}{\bibinfo{person}{Thomas Elsken}, \bibinfo{person}{Jan~Hendrik
  Metzen}, {and} \bibinfo{person}{Frank Hutter}.}
  \bibinfo{year}{2019}\natexlab{}.
\newblock \showarticletitle{Efficient Multi-Objective Neural Architecture
  Search via Lamarckian Evolution}. In \bibinfo{booktitle}{\emph{Proc. Int.
  Conf. Learn. Represent.}} \bibinfo{pages}{https://arxiv.org/abs/1804.09081}.
\newblock


\bibitem[Erguzel et~al\mbox{.}(2014)]%
        {z19}
\bibfield{author}{\bibinfo{person}{Turker~Tekin Erguzel},
  \bibinfo{person}{Serhat Ozekes}, \bibinfo{person}{Selahattin Gultekin}, {and}
  \bibinfo{person}{Nevzat Tarhan}.} \bibinfo{year}{2014}\natexlab{}.
\newblock \showarticletitle{Ant Colony optimization Based Feature Selection
  Method for {QEEG} data classification}.
\newblock \bibinfo{journal}{\emph{Psychiatry Investig.}} \bibinfo{volume}{11},
  \bibinfo{number}{3} (\bibinfo{year}{2014}), \bibinfo{pages}{243}.
\newblock


\bibitem[Est{\'e}vez and Caballero(1998)]%
        {165r}
\bibfield{author}{\bibinfo{person}{Pablo~A Est{\'e}vez} {and}
  \bibinfo{person}{Rodrigo~E Caballero}.} \bibinfo{year}{1998}\natexlab{}.
\newblock \showarticletitle{A Niching Genetic Algorithm for Selecting Features
  for Neural Network Classifiers}. In \bibinfo{booktitle}{\emph{Proc. Int.
  Conf. Artif. Neural Netw.}} \bibinfo{pages}{311--316}.
\newblock
\showISSN{1431-6854}


\bibitem[Evans et~al\mbox{.}(2018a)]%
        {z01}
\bibfield{author}{\bibinfo{person}{Benjamin Evans}, \bibinfo{person}{Harith
  Al-Sahaf}, \bibinfo{person}{Bing Xue}, {and} \bibinfo{person}{Mengjie
  Zhang}.} \bibinfo{year}{2018}\natexlab{a}.
\newblock \showarticletitle{Evolutionary Deep Learning: A Genetic Programming
  Approach to Image Classification}. In \bibinfo{booktitle}{\emph{Proc. IEEE
  Congr. Evol. Comput.}} \bibinfo{pages}{1538--1545}.
\newblock


\bibitem[Evans et~al\mbox{.}(2018b)]%
        {795r}
\bibfield{author}{\bibinfo{person}{Benjamin~Patrick Evans},
  \bibinfo{person}{Harith Al-Sahaf}, \bibinfo{person}{Bing Xue}, {and}
  \bibinfo{person}{Mengjie Zhang}.} \bibinfo{year}{2018}\natexlab{b}.
\newblock \showarticletitle{Evolutionary Deep Learning: A Genetic Programming
  Approach to Image Classification}. In \bibinfo{booktitle}{\emph{Proc. IEEE
  Congr. Evol. Comput.}} \bibinfo{pages}{1--6}.
\newblock


\bibitem[Fahrudin et~al\mbox{.}(2016)]%
        {302r}
\bibfield{author}{\bibinfo{person}{Tresna~Maulana Fahrudin},
  \bibinfo{person}{Iwan Syarif}, {and} \bibinfo{person}{Ali~Ridho Barakbah}.}
  \bibinfo{year}{2016}\natexlab{}.
\newblock \showarticletitle{Ant Colony Algorithm for Feature Selection on
  Microarray Datasets}. In \bibinfo{booktitle}{\emph{International Electronics
  Symposium}}. \bibinfo{pages}{351--356}.
\newblock


\bibitem[Fan et~al\mbox{.}(2020)]%
        {598r}
\bibfield{author}{\bibinfo{person}{Zhun Fan}, \bibinfo{person}{Jiahong Wei},
  \bibinfo{person}{Guijie Zhu}, \bibinfo{person}{Jiajie Mo}, {and}
  \bibinfo{person}{Wenji Li}.} \bibinfo{year}{2020}\natexlab{}.
\newblock \showarticletitle{Evolutionary Neural Architecture Search for Retinal
  Vessel Segmentation}.
\newblock \bibinfo{journal}{\emph{arXiv preprint arXiv:2001.06678}}
  (\bibinfo{year}{2020}).
\newblock


\bibitem[Fernandes and Yen(2021)]%
        {826r}
\bibfield{author}{\bibinfo{person}{Francisco~Erivaldo Fernandes} {and}
  \bibinfo{person}{Gary~G. Yen}.} \bibinfo{year}{2021}\natexlab{}.
\newblock \showarticletitle{Automatic Searching and Pruning of Deep Neural
  Networks for Medical Imaging Diagnostic}.
\newblock \bibinfo{journal}{\emph{IEEE Trans. Neural Netw. Learn. Syst.}}
  \bibinfo{volume}{32}, \bibinfo{number}{12} (\bibinfo{year}{2021}),
  \bibinfo{pages}{5664--5674}.
\newblock


\bibitem[Fogelberg and Zhang(2005)]%
        {z05}
\bibfield{author}{\bibinfo{person}{Christopher Fogelberg} {and}
  \bibinfo{person}{Mengjie Zhang}.} \bibinfo{year}{2005}\natexlab{}.
\newblock \showarticletitle{Linear Genetic Programming for Multi-class Object
  Classification}. In \bibinfo{booktitle}{\emph{Proc. Aust. Joint Conf. Artif.l
  Intell.}} \bibinfo{pages}{369--379}.
\newblock


\bibitem[Fortuna and Frasca(2021)]%
        {801r}
\bibfield{author}{\bibinfo{person}{Luigi Fortuna} {and} \bibinfo{person}{Mattia
  Frasca}.} \bibinfo{year}{2021}\natexlab{}.
\newblock \showarticletitle{Singular Value Decomposition}.
\newblock \bibinfo{journal}{\emph{Optim. Rob. Control}} (\bibinfo{year}{2021}),
  \bibinfo{pages}{51--58}.
\newblock


\bibitem[Frachon et~al\mbox{.}(2019)]%
        {660r}
\bibfield{author}{\bibinfo{person}{Luc Frachon}, \bibinfo{person}{Wei Pang},
  {and} \bibinfo{person}{George~M Coghill}.} \bibinfo{year}{2019}\natexlab{}.
\newblock \showarticletitle{Immunecs: Neural Committee Search by an Artificial
  Immune System}.
\newblock \bibinfo{journal}{\emph{arXiv preprint arXiv:1911.07729}}
  (\bibinfo{year}{2019}).
\newblock


\bibitem[Freitas(2003)]%
        {051r}
\bibfield{author}{\bibinfo{person}{Alex~A Freitas}.}
  \bibinfo{year}{2003}\natexlab{}.
\newblock \showarticletitle{A Survey of Evolutionary Algorithms for Data Mining
  and Knowledge Discovery}.
\newblock In \bibinfo{booktitle}{\emph{Adv. Evol. Comput.}}
  \bibinfo{publisher}{Springer}, \bibinfo{pages}{819--845}.
\newblock


\bibitem[Fujino et~al\mbox{.}(2017)]%
        {540r}
\bibfield{author}{\bibinfo{person}{Saya Fujino}, \bibinfo{person}{Naoki Mori},
  {and} \bibinfo{person}{Keinosuke Matsumoto}.}
  \bibinfo{year}{2017}\natexlab{}.
\newblock \showarticletitle{Deep Convolutional Networks for Human Sketches By
  Means of The Evolutionary Deep Learning}. In \bibinfo{booktitle}{\emph{Proc.
  Int. Conf. Soft Comput. Intell. Syst.}} \bibinfo{pages}{1--5}.
\newblock


\bibitem[Garc{\'i}a et~al\mbox{.}(2011)]%
        {371r}
\bibfield{author}{\bibinfo{person}{David Garc{\'i}a},
  \bibinfo{person}{Antonio~Gonz{\'a}lez Mu{\~n}oz}, {and}
  \bibinfo{person}{Ra{\'u}l P{\'e}rez}.} \bibinfo{year}{2011}\natexlab{}.
\newblock \showarticletitle{A Two-Step Approach of Feature Construction for A
  Genetic Learning Algorithm}.
\newblock \bibinfo{journal}{\emph{Proc. Int. Conf. Fuzzy Syst.}}
  (\bibinfo{year}{2011}), \bibinfo{pages}{1255--1262}.
\newblock
\showISSN{0022-0000}


\bibitem[Gerum et~al\mbox{.}(2020)]%
        {821r}
\bibfield{author}{\bibinfo{person}{Richard~C Gerum}, \bibinfo{person}{Andr{\'e}
  Erpenbeck}, \bibinfo{person}{Patrick Krauss}, {and} \bibinfo{person}{Achim
  Schilling}.} \bibinfo{year}{2020}\natexlab{}.
\newblock \showarticletitle{Sparsity Through Evolutionary Pruning Prevents
  Neuronal Networks From Overfitting}.
\newblock \bibinfo{journal}{\emph{Neural Netw.}}  \bibinfo{volume}{128}
  (\bibinfo{year}{2020}), \bibinfo{pages}{305--312}.
\newblock


\bibitem[Golubski and Feuring(1999)]%
        {728r}
\bibfield{author}{\bibinfo{person}{Wolfgang Golubski} {and}
  \bibinfo{person}{Thomas Feuring}.} \bibinfo{year}{1999}\natexlab{}.
\newblock \showarticletitle{Evolving Neural Network Structures by Means of
  Genetic Programming}. In \bibinfo{booktitle}{\emph{Proc. Eur. Conf. Genetic
  Program}}. \bibinfo{pages}{211--220}.
\newblock
\showISSN{0302-9743}


\bibitem[Gou et~al\mbox{.}(2021)]%
        {842r}
\bibfield{author}{\bibinfo{person}{Jianping Gou}, \bibinfo{person}{Baosheng
  Yu}, \bibinfo{person}{Stephen~J Maybank}, {and} \bibinfo{person}{Dacheng
  Tao}.} \bibinfo{year}{2021}\natexlab{}.
\newblock \showarticletitle{Knowledge Distillation: A Survey}.
\newblock \bibinfo{journal}{\emph{Int. J. Comput. Vis.}} \bibinfo{volume}{129},
  \bibinfo{number}{6} (\bibinfo{year}{2021}), \bibinfo{pages}{1789--1819}.
\newblock


\bibitem[Guo et~al\mbox{.}(2020)]%
        {581r}
\bibfield{author}{\bibinfo{person}{Zichao Guo}, \bibinfo{person}{Xiangyu
  Zhang}, \bibinfo{person}{Haoyuan Mu}, \bibinfo{person}{Wen Heng},
  \bibinfo{person}{Zechun Liu}, \bibinfo{person}{Yichen Wei}, {and}
  \bibinfo{person}{Jian Sun}.} \bibinfo{year}{2020}\natexlab{}.
\newblock \showarticletitle{Single Path One-Shot Neural Architecture Search
  with Uniform Sampling}. In \bibinfo{booktitle}{\emph{Proc. Eur. Conf. Comput.
  Vis.}} \bibinfo{pages}{544--560}.
\newblock
\showISSN{0302-9743}


\bibitem[Hajati et~al\mbox{.}(2010)]%
        {439r}
\bibfield{author}{\bibinfo{person}{Farshid Hajati}, \bibinfo{person}{Caro
  Lucas}, {and} \bibinfo{person}{Yongsheng Gao}.}
  \bibinfo{year}{2010}\natexlab{}.
\newblock \showarticletitle{Face Localization Using an Effective
  Co-evolutionary Genetic Algorithm}.
\newblock \bibinfo{journal}{\emph{Proc. Int. Conf. Digit. Image Comput.: Tech.
  and Appl.}} (\bibinfo{year}{2010}), \bibinfo{pages}{522--527}.
\newblock


\bibitem[Hammami et~al\mbox{.}(2018)]%
        {381r}
\bibfield{author}{\bibinfo{person}{Marwa Hammami}, \bibinfo{person}{Slim
  Bechikh}, {and} \bibinfo{person}{Chih-Cheng Hung}.}
  \bibinfo{year}{2018}\natexlab{}.
\newblock \showarticletitle{A Multi-Objective Hybrid Filter-Wrapper
  Evolutionary Approach for Feature Construction on High-Dimensional Data}. In
  \bibinfo{booktitle}{\emph{Proc. IEEE Congr. Evol. Comput.}}
  \bibinfo{pages}{1--8}.
\newblock


\bibitem[Han and Cho(2006)]%
        {711r}
\bibfield{author}{\bibinfo{person}{Sang-Jun Han} {and}
  \bibinfo{person}{Sung-Bae Cho}.} \bibinfo{year}{2006}\natexlab{}.
\newblock \showarticletitle{Evolutionary Neural Networks for Anomaly Detection
  Based on the Behavior of a Program}.
\newblock \bibinfo{journal}{\emph{IEEE Trans. Syst. Man Cybern.}}
  \bibinfo{volume}{36}, \bibinfo{number}{3} (\bibinfo{year}{2006}),
  \bibinfo{pages}{559--570}.
\newblock


\bibitem[Hancer et~al\mbox{.}(2015)]%
        {200r}
\bibfield{author}{\bibinfo{person}{Emrah Hancer}, \bibinfo{person}{Bing Xue},
  \bibinfo{person}{Mengjie Zhang}, {and} \bibinfo{person}{Dervis Karaboga}.}
  \bibinfo{year}{2015}\natexlab{}.
\newblock \showarticletitle{A Multi-objective Artificial Bee Colony Approach to
  Feature Selection Using Fuzzy Mutual Information}. In
  \bibinfo{booktitle}{\emph{Proc. IEEE Congr. Evol. Comput.}}
  \bibinfo{pages}{2420--2427}.
\newblock


\bibitem[He et~al\mbox{.}(2021)]%
        {025r}
\bibfield{author}{\bibinfo{person}{Xin He}, \bibinfo{person}{Kaiyong Zhao},
  {and} \bibinfo{person}{Xiaowen Chu}.} \bibinfo{year}{2021}\natexlab{}.
\newblock \showarticletitle{Auto{ML}: A Survey of the State-of-the-Art}.
\newblock \bibinfo{journal}{\emph{Knowl-Based Syst}}  \bibinfo{volume}{212}
  (\bibinfo{year}{2021}), \bibinfo{pages}{106622}.
\newblock
\showISSN{0950-7051}


\bibitem[He et~al\mbox{.}(2017)]%
        {765r}
\bibfield{author}{\bibinfo{person}{Yihui He}, \bibinfo{person}{Xiangyu Zhang},
  {and} \bibinfo{person}{Jian Sun}.} \bibinfo{year}{2017}\natexlab{}.
\newblock \showarticletitle{Channel Pruning for Accelerating Very Deep Neural
  Networks}. In \bibinfo{booktitle}{\emph{Proc. IEEE Int. Conf. Comput. Vis.}}
  \bibinfo{pages}{1398--1406}.
\newblock


\bibitem[Ho et~al\mbox{.}(2021)]%
        {583r}
\bibfield{author}{\bibinfo{person}{Kary Ho}, \bibinfo{person}{Andrew Gilbert},
  \bibinfo{person}{Hailin Jin}, {and} \bibinfo{person}{John~P. Collomosse}.}
  \bibinfo{year}{2021}\natexlab{}.
\newblock \showarticletitle{Neural Architecture Search for Deep Image Prior}.
\newblock \bibinfo{journal}{\emph{Comput. \& Graph.}}  \bibinfo{volume}{98}
  (\bibinfo{year}{2021}), \bibinfo{pages}{188--196}.
\newblock
\showISSN{0097-8493}


\bibitem[Hong and Cho(2006)]%
        {166r}
\bibfield{author}{\bibinfo{person}{Jin-Hyuk Hong} {and}
  \bibinfo{person}{Sung-Bae Cho}.} \bibinfo{year}{2006}\natexlab{}.
\newblock \showarticletitle{Efficient Huge-Scale Feature Selection With
  Speciated Genetic Algorithm}.
\newblock \bibinfo{journal}{\emph{Pattern Recognit. Lett.}}
  \bibinfo{volume}{27}, \bibinfo{number}{2} (\bibinfo{year}{2006}),
  \bibinfo{pages}{143--150}.
\newblock
\showISSN{0167-8655}


\bibitem[Hong et~al\mbox{.}(2020)]%
        {837r}
\bibfield{author}{\bibinfo{person}{Wenjing Hong}, \bibinfo{person}{Peng Yang},
  \bibinfo{person}{Yiwen Wang}, {and} \bibinfo{person}{Ke Tang}.}
  \bibinfo{year}{2020}\natexlab{}.
\newblock \showarticletitle{Multi-objective Magnitude-Based Pruning for
  Latency-Aware Deep Neural Network Compression}. In
  \bibinfo{booktitle}{\emph{Proc. Int. Conf. on Parallel Probl. Solving Nat.}}
  \bibinfo{pages}{470--483}.
\newblock
\showISSN{0302-9743}


\bibitem[Hosni et~al\mbox{.}(2020)]%
        {202r}
\bibfield{author}{\bibinfo{person}{Mohamed Hosni}, \bibinfo{person}{Gin{\'e}s
  Garc{\'\i}a-Mateos}, {and} \bibinfo{person}{Juan Carrillo-de Gea}.}
  \bibinfo{year}{2020}\natexlab{}.
\newblock \showarticletitle{A Mapping Study of Ensemble Classification Methods
  in Lung Cancer Decision Support Systems}.
\newblock \bibinfo{journal}{\emph{Med. \& Biol. Eng. \& Comput.}}
  \bibinfo{volume}{58}, \bibinfo{number}{10} (\bibinfo{year}{2020}),
  \bibinfo{pages}{2177--2193}.
\newblock
\showISSN{0140-0118}


\bibitem[Hsu et~al\mbox{.}(2021)]%
        {846r}
\bibfield{author}{\bibinfo{person}{Yen-Chang Hsu}, \bibinfo{person}{Ting Hua},
  \bibinfo{person}{Sungen Chang}, \bibinfo{person}{Qian Lou},
  \bibinfo{person}{Yilin Shen}, {and} \bibinfo{person}{Hongxia Jin}.}
  \bibinfo{year}{2021}\natexlab{}.
\newblock \showarticletitle{Language Model Compression with Weighted Low-rank
  Factorization}. In \bibinfo{booktitle}{\emph{Proc. Int. Conf. Learn.
  Represent.}} \bibinfo{pages}{https://arxiv.org/abs/2207.00112}.
\newblock


\bibitem[Hu et~al\mbox{.}(2021b)]%
        {759r}
\bibfield{author}{\bibinfo{person}{Bin Hu}, \bibinfo{person}{Tianming Zhao},
  \bibinfo{person}{Yucheng Xie}, \bibinfo{person}{Yan Wang}, {and}
  \bibinfo{person}{Xiaonan Guo}.} \bibinfo{year}{2021}\natexlab{b}.
\newblock \showarticletitle{{MIXP}: Efficient Deep Neural Networks Pruning for
  Further {FLOP}s Compression via Neuron Bond}. In
  \bibinfo{booktitle}{\emph{Proc. Int. Joint Conf. Neural Netw.}}
  \bibinfo{pages}{1--8}.
\newblock


\bibitem[Hu et~al\mbox{.}(2021a)]%
        {782r}
\bibfield{author}{\bibinfo{person}{Yiming Hu}, \bibinfo{person}{Xingang Wang},
  \bibinfo{person}{Lujun Li}, {and} \bibinfo{person}{Qingyi Gu}.}
  \bibinfo{year}{2021}\natexlab{a}.
\newblock \showarticletitle{Improving One-Shot NAS with Shrinking-and-Expanding
  Supernet}.
\newblock \bibinfo{journal}{\emph{Pattern Recognit.}}  \bibinfo{volume}{118}
  (\bibinfo{year}{2021}), \bibinfo{pages}{108025}.
\newblock
\showISSN{0031-3203}


\bibitem[Huang et~al\mbox{.}(2012)]%
        {z20}
\bibfield{author}{\bibinfo{person}{Hu Huang}, \bibinfo{person}{Hong-Bo Xie},
  \bibinfo{person}{Jing-Yi Guo}, {and} \bibinfo{person}{Hui-Juan Chen}.}
  \bibinfo{year}{2012}\natexlab{}.
\newblock \showarticletitle{Ant Colony Optimization-based Feature Selection
  Method for Surface Electromyography Signals Classification}.
\newblock \bibinfo{journal}{\emph{Comput. Biol. Med.}} \bibinfo{volume}{42},
  \bibinfo{number}{1} (\bibinfo{year}{2012}), \bibinfo{pages}{30--38}.
\newblock


\bibitem[Huang et~al\mbox{.}(2020)]%
        {802r}
\bibfield{author}{\bibinfo{person}{Junhao Huang}, \bibinfo{person}{Weize Sun},
  {and} \bibinfo{person}{Lei Huang}.} \bibinfo{year}{2020}\natexlab{}.
\newblock \showarticletitle{Deep Neural Networks Compression Learning Based on
  Multiobjective Evolutionary Algorithms}.
\newblock \bibinfo{journal}{\emph{Neurocomputing}}  \bibinfo{volume}{378}
  (\bibinfo{year}{2020}), \bibinfo{pages}{260--269}.
\newblock
\showISSN{0925-2312}


\bibitem[Ijjina and Chalavadi(2016)]%
        {103r}
\bibfield{author}{\bibinfo{person}{Earnest~Paul Ijjina} {and}
  \bibinfo{person}{Krishna~Mohan Chalavadi}.} \bibinfo{year}{2016}\natexlab{}.
\newblock \showarticletitle{Human Action Recognition Using Genetic Algorithms
  and Convolutional Neural Networks}.
\newblock \bibinfo{journal}{\emph{Pattern Recognit.}}  \bibinfo{volume}{59}
  (\bibinfo{year}{2016}), \bibinfo{pages}{199--212}.
\newblock
\showISSN{0031-3203}


\bibitem[Irwin-Harris et~al\mbox{.}(2019)]%
        {673r}
\bibfield{author}{\bibinfo{person}{William Irwin-Harris},
  \bibinfo{person}{Yanan Sun}, \bibinfo{person}{Bing Xue}, {and}
  \bibinfo{person}{Mengjie Zhang}.} \bibinfo{year}{2019}\natexlab{}.
\newblock \showarticletitle{A Graph-Based Encoding for Evolutionary
  Convolutional Neural Network Architecture Design}. In
  \bibinfo{booktitle}{\emph{Proc. IEEE Congr. Evol. Comput.}}
  \bibinfo{pages}{546--553}.
\newblock


\bibitem[Izenman(2013)]%
        {412r}
\bibfield{author}{\bibinfo{person}{Alan~Julian Izenman}.}
  \bibinfo{year}{2013}\natexlab{}.
\newblock \showarticletitle{Linear Discriminant Analysis}.
\newblock In \bibinfo{booktitle}{\emph{Modern Multivariate Statistical
  Techniques}}. \bibinfo{publisher}{Springer}, \bibinfo{pages}{237--280}.
\newblock
\showISSN{1431-875X}


\bibitem[Ja{\^a}fra et~al\mbox{.}(2019)]%
        {468r}
\bibfield{author}{\bibinfo{person}{Yesmina Ja{\^a}fra},
  \bibinfo{person}{Jean~Luc Laurent}, \bibinfo{person}{Aline Deruyver}, {and}
  \bibinfo{person}{Mohamed~Saber Naceur}.} \bibinfo{year}{2019}\natexlab{}.
\newblock \showarticletitle{Reinforcement Learning for Neural Architecture
  Search: A Review}.
\newblock \bibinfo{journal}{\emph{Image Vis. Comput.}}  \bibinfo{volume}{89}
  (\bibinfo{year}{2019}), \bibinfo{pages}{57--66}.
\newblock
\showISSN{0262-8856}


\bibitem[Jin et~al\mbox{.}(2018)]%
        {685r}
\bibfield{author}{\bibinfo{person}{Haifeng Jin}, \bibinfo{person}{Qingquan
  Song}, {and} \bibinfo{person}{Xia Hu}.} \bibinfo{year}{2018}\natexlab{}.
\newblock \showarticletitle{Auto-keras: Efficient Neural Architecture Search
  with Network Morphism}.
\newblock \bibinfo{journal}{\emph{arXiv preprint arXiv:1806.10282}}
  (\bibinfo{year}{2018}).
\newblock


\bibitem[Jin et~al\mbox{.}(2019)]%
        {684r}
\bibfield{author}{\bibinfo{person}{Haifeng Jin}, \bibinfo{person}{Qingquan
  Song}, {and} \bibinfo{person}{Xia Hu}.} \bibinfo{year}{2019}\natexlab{}.
\newblock \showarticletitle{Auto-{K}eras: An Efficient Neural Architecture
  Search System}. In \bibinfo{booktitle}{\emph{Proc. ACM SIGKDD Int. Conf. on
  Knowl. Discov. \& Data Min.}} \bibinfo{pages}{1946--1956}.
\newblock


\bibitem[Jin(2006)]%
        {040r}
\bibfield{author}{\bibinfo{person}{Yaochu Jin}.}
  \bibinfo{year}{2006}\natexlab{}.
\newblock \bibinfo{booktitle}{\emph{Multi-Objective Machine Learning}}.
\newblock \bibinfo{publisher}{Springer Science}.
\newblock


\bibitem[Jones et~al\mbox{.}(2019)]%
        {582r}
\bibfield{author}{\bibinfo{person}{David~T Jones}, \bibinfo{person}{Anja
  Schroeder}, {and} \bibinfo{person}{Geoff~S. Nitschke}.}
  \bibinfo{year}{2019}\natexlab{}.
\newblock \showarticletitle{Evolutionary Deep Learning to Identify Galaxies in
  the Zone of Avoidance}.
\newblock \bibinfo{journal}{\emph{arXiv preprint arXiv:1903.07461}}
  (\bibinfo{year}{2019}).
\newblock


\bibitem[Junior and Yen(2021a)]%
        {812r}
\bibfield{author}{\bibinfo{person}{Francisco Erivaldo~Fernandes Junior} {and}
  \bibinfo{person}{Gary~G. Yen}.} \bibinfo{year}{2021}\natexlab{a}.
\newblock \showarticletitle{Pruning Deep Convolutional Neural Networks
  Architectures with Evolution Strategy}.
\newblock \bibinfo{journal}{\emph{Inf. Sci.}}  \bibinfo{volume}{552}
  (\bibinfo{year}{2021}), \bibinfo{pages}{29--47}.
\newblock
\showISSN{0020-0255}


\bibitem[Junior and Yen(2021b)]%
        {820r}
\bibfield{author}{\bibinfo{person}{Francisco Erivaldo~Fernandes Junior} {and}
  \bibinfo{person}{Gary~G. Yen}.} \bibinfo{year}{2021}\natexlab{b}.
\newblock \showarticletitle{Pruning of Generative Adversarial Neural Networks
  for Medical Imaging Diagnostics with Evolution Strategy}.
\newblock \bibinfo{journal}{\emph{Inf. Sci.}}  \bibinfo{volume}{558}
  (\bibinfo{year}{2021}), \bibinfo{pages}{91--102}.
\newblock
\showISSN{0020-0255}


\bibitem[Karaboga et~al\mbox{.}(2007)]%
        {490r}
\bibfield{author}{\bibinfo{person}{Dervis Karaboga}, \bibinfo{person}{Bahriye
  Akay}, {and} \bibinfo{person}{Celal {\"O}zt{\"u}rk}.}
  \bibinfo{year}{2007}\natexlab{}.
\newblock \showarticletitle{Artificial Bee Colony ({ABC}) Optimization
  Algorithm for Training Feed-Forward Neural Networks}. In
  \bibinfo{booktitle}{\emph{Proc. Int. Conf. Modeling Decis. Artif. Intell.}}
  \bibinfo{pages}{318--329}.
\newblock


\bibitem[Karegowda and Manjunath(2011)]%
        {472r}
\bibfield{author}{\bibinfo{person}{Asha~Gowda Karegowda} {and}
  \bibinfo{person}{A.~S. Manjunath}.} \bibinfo{year}{2011}\natexlab{}.
\newblock \showarticletitle{Application of Genetic Algorithm Optimized Neural
  Network Connection Weights for Medical Diagnosis of PIMA Indians Diabetes}.
\newblock \bibinfo{journal}{\emph{Int. J. Soft Comput.}} \bibinfo{volume}{2},
  \bibinfo{number}{2} (\bibinfo{year}{2011}), \bibinfo{pages}{15--23}.
\newblock
\showISSN{2229-7103}


\bibitem[Khadka and Tumer(2018)]%
        {532r}
\bibfield{author}{\bibinfo{person}{Shauharda Khadka} {and}
  \bibinfo{person}{Kagan Tumer}.} \bibinfo{year}{2018}\natexlab{}.
\newblock \showarticletitle{Evolution-Guided Policy Gradient in Reinforcement
  Learning}. In \bibinfo{booktitle}{\emph{Proc. Adv. Neural Inf. Process.
  Syst.}}, Vol.~\bibinfo{volume}{31}. \bibinfo{pages}{1196--1208}.
\newblock


\bibitem[Khushaba et~al\mbox{.}(2008)]%
        {163r}
\bibfield{author}{\bibinfo{person}{Rami~N Khushaba}, \bibinfo{person}{Ahmed
  Al-Ani}, \bibinfo{person}{Akram AlSukker}, {and} \bibinfo{person}{Adel
  Al-Jumaily}.} \bibinfo{year}{2008}\natexlab{}.
\newblock \showarticletitle{A Combined Ant Colony and Differential Evolution
  Feature Selection Algorithm}. In \bibinfo{booktitle}{\emph{Proc. Int. Conf.
  Ant Colony Optim. Swarm Intell.}} \bibinfo{pages}{1--12}.
\newblock
\showISSN{1051-4651}


\bibitem[Kitano(1990)]%
        {674r}
\bibfield{author}{\bibinfo{person}{Hiroaki Kitano}.}
  \bibinfo{year}{1990}\natexlab{}.
\newblock \showarticletitle{Designing Neural Networks Using Genetic Algorithms
  with Graph Generation System}.
\newblock \bibinfo{journal}{\emph{Complex Syst.}} \bibinfo{volume}{4},
  \bibinfo{number}{4} (\bibinfo{year}{1990}), \bibinfo{pages}{225--238}.
\newblock
\showISSN{0167-2789}


\bibitem[Kotani and Kato(2004)]%
        {433r}
\bibfield{author}{\bibinfo{person}{Manabu Kotani} {and}
  \bibinfo{person}{Daisuke Kato}.} \bibinfo{year}{2004}\natexlab{}.
\newblock \showarticletitle{Feature Extraction Using Coevolutionary Genetic
  Programming}. In \bibinfo{booktitle}{\emph{Proc. IEEE Congr. Evol. Comput.}}
  \bibinfo{pages}{614--619}.
\newblock


\bibitem[Koutn{\'i}k et~al\mbox{.}(2010)]%
        {373r}
\bibfield{author}{\bibinfo{person}{Jan Koutn{\'i}k},
  \bibinfo{person}{Faustino~J. Gomez}, {and} \bibinfo{person}{J{\"u}rgen
  Schmidhuber}.} \bibinfo{year}{2010}\natexlab{}.
\newblock \showarticletitle{Evolving Neural Networks in Compressed Weight
  Space}. In \bibinfo{booktitle}{\emph{Proc. Genetic Evol. Comput. Conf.}}
  \bibinfo{pages}{619--626}.
\newblock


\bibitem[Koutn{\'i}k et~al\mbox{.}(2014)]%
        {735r}
\bibfield{author}{\bibinfo{person}{Jan Koutn{\'i}k},
  \bibinfo{person}{J{\"u}rgen Schmidhuber}, {and} \bibinfo{person}{Faustino~J.
  Gomez}.} \bibinfo{year}{2014}\natexlab{}.
\newblock \showarticletitle{Evolving Deep Unsupervised Convolutional Networks
  for Vision-Based Reinforcement Learning}. In \bibinfo{booktitle}{\emph{Proc.
  Genetic Evol. Comput. Conf.}} \bibinfo{pages}{541--548}.
\newblock


\bibitem[Krawiec(2002)]%
        {113r}
\bibfield{author}{\bibinfo{person}{Krzysztof Krawiec}.}
  \bibinfo{year}{2002}\natexlab{}.
\newblock \showarticletitle{Genetic Programming-Based Construction of Features
  for Machine Learning and Knowledge Discovery Tasks}.
\newblock \bibinfo{journal}{\emph{Genet. Program Evolvable Mach.}}
  \bibinfo{volume}{3}, \bibinfo{number}{4} (\bibinfo{year}{2002}),
  \bibinfo{pages}{329--343}.
\newblock


\bibitem[Krizhevsky et~al\mbox{.}(2009)]%
        {Krizhevsky2009LearningML}
\bibfield{author}{\bibinfo{person}{Alex Krizhevsky}, \bibinfo{person}{Geoffrey
  Hinton}, {et~al\mbox{.}}} \bibinfo{year}{2009}\natexlab{}.
\newblock \showarticletitle{Learning Multiple Layers of Features From Tiny
  Images}.
\newblock  (\bibinfo{year}{2009}).
\newblock


\bibitem[Krizhevsky et~al\mbox{.}(2012)]%
        {A001r}
\bibfield{author}{\bibinfo{person}{Alex Krizhevsky}, \bibinfo{person}{Ilya
  Sutskever}, {and} \bibinfo{person}{Geoffrey~E Hinton}.}
  \bibinfo{year}{2012}\natexlab{}.
\newblock \showarticletitle{Image{N}et Classification with Deep Convolutional
  Neural Networks}. In \bibinfo{booktitle}{\emph{Proc. Adv. Neural Inf.
  Process. Syst.}}, Vol.~\bibinfo{volume}{25}. \bibinfo{pages}{1097--1105}.
\newblock


\bibitem[Kwasigroch et~al\mbox{.}(2019)]%
        {658r}
\bibfield{author}{\bibinfo{person}{Arkadiusz Kwasigroch},
  \bibinfo{person}{Michał Grochowski}, {and} \bibinfo{person}{Mateusz
  Mikolajczyk}.} \bibinfo{year}{2019}\natexlab{}.
\newblock \showarticletitle{Deep Neural Network Architecture Search using
  Network Morphism}. In \bibinfo{booktitle}{\emph{Proc. Int. Conf. Methods and
  Models in Autom. and Robot.}} \bibinfo{pages}{30--35}.
\newblock


\bibitem[LeCun et~al\mbox{.}(2015)]%
        {004r}
\bibfield{author}{\bibinfo{person}{Yann LeCun}, \bibinfo{person}{Yoshua
  Bengio}, {and} \bibinfo{person}{Geoffrey Hinton}.}
  \bibinfo{year}{2015}\natexlab{}.
\newblock \showarticletitle{Deep Learning}.
\newblock \bibinfo{journal}{\emph{Nature}} \bibinfo{volume}{521},
  \bibinfo{number}{7553} (\bibinfo{year}{2015}), \bibinfo{pages}{436--444}.
\newblock
\showISSN{1476-4687}


\bibitem[LeCun et~al\mbox{.}(1989)]%
        {z13}
\bibfield{author}{\bibinfo{person}{Yann LeCun}, \bibinfo{person}{Bernhard
  Boser}, \bibinfo{person}{John~S Denker}, \bibinfo{person}{Donnie Henderson},
  \bibinfo{person}{Richard~E Howard}, {and} \bibinfo{person}{Wayne Hubbard}.}
  \bibinfo{year}{1989}\natexlab{}.
\newblock \showarticletitle{Backpropagation Applied to Handwritten Zip Code
  Recognition}.
\newblock \bibinfo{journal}{\emph{Neural computation}} \bibinfo{volume}{1},
  \bibinfo{number}{4} (\bibinfo{year}{1989}), \bibinfo{pages}{541--551}.
\newblock


\bibitem[Li et~al\mbox{.}(2020)]%
        {831r}
\bibfield{author}{\bibinfo{person}{Bailin Li}, \bibinfo{person}{Bowen Wu},
  \bibinfo{person}{Jiang Su}, \bibinfo{person}{Guangrun Wang}, {and}
  \bibinfo{person}{Liang Lin}.} \bibinfo{year}{2020}\natexlab{}.
\newblock \showarticletitle{Eagle{E}ye: Fast Sub-net Evaluation for Efficient
  Neural Network Pruning}. In \bibinfo{booktitle}{\emph{Proc. Eur. Conf.
  Comput. Vis.}} \bibinfo{pages}{639--654}.
\newblock
\showISSN{0302-9743}


\bibitem[Li et~al\mbox{.}(2021)]%
        {062r}
\bibfield{author}{\bibinfo{person}{Chaojian Li}, \bibinfo{person}{Zhongzhi Yu},
  \bibinfo{person}{Yonggan Fu}, \bibinfo{person}{Yongan Zhang},
  \bibinfo{person}{Yang Zhao}, \bibinfo{person}{Haoran You},
  \bibinfo{person}{Qixuan Yu}, \bibinfo{person}{Yue Wang},
  \bibinfo{person}{Cong Hao}, {and} \bibinfo{person}{Yingyan Lin}.}
  \bibinfo{year}{2021}\natexlab{}.
\newblock \showarticletitle{{HW-NAS}-{B}ench: Hardware-Aware Neural
  Architecture Search Benchmark}. In \bibinfo{booktitle}{\emph{Proc. Int. Conf.
  Learn. Represent.}} \bibinfo{pages}{https://arxiv.org/abs/2103.10584}.
\newblock


\bibitem[Li et~al\mbox{.}(2022)]%
        {665r}
\bibfield{author}{\bibinfo{person}{Qing Li}, \bibinfo{person}{Wei Zhang},
  \bibinfo{person}{Lin Zhao}, \bibinfo{person}{Xia Wu}, {and}
  \bibinfo{person}{Tianming Liu}.} \bibinfo{year}{2022}\natexlab{}.
\newblock \showarticletitle{Evolutional Neural Architecture Search for
  Optimization of Spatiotemporal Brain Network Decomposition}.
\newblock \bibinfo{journal}{\emph{IEEE. Trans. Biomed. Eng.}}
  \bibinfo{volume}{69}, \bibinfo{number}{2} (\bibinfo{year}{2022}),
  \bibinfo{pages}{624--634}.
\newblock


\bibitem[Li et~al\mbox{.}(2019)]%
        {504r}
\bibfield{author}{\bibinfo{person}{Youru Li}, \bibinfo{person}{Zhenfeng Zhu},
  \bibinfo{person}{Deqiang Kong}, \bibinfo{person}{Hua Han}, {and}
  \bibinfo{person}{Yao Zhao}.} \bibinfo{year}{2019}\natexlab{}.
\newblock \showarticletitle{{EA-LSTM}: Evolutionary Attention-Based {LSTM} for
  Time Series Prediction}.
\newblock \bibinfo{journal}{\emph{Knowl.-Based Syst.}}  \bibinfo{volume}{181}
  (\bibinfo{year}{2019}), \bibinfo{pages}{104785}.
\newblock
\showISSN{0950-7051}


\bibitem[Liu et~al\mbox{.}(2018c)]%
        {016r}
\bibfield{author}{\bibinfo{person}{Chenxi Liu}, \bibinfo{person}{Barret Zoph},
  \bibinfo{person}{Maxim Neumann}, \bibinfo{person}{Jonathon Shlens},
  \bibinfo{person}{Wei Hua}, \bibinfo{person}{Li-Jia Li}, \bibinfo{person}{Li
  Fei-Fei}, \bibinfo{person}{Alan Yuille}, \bibinfo{person}{Jonathan Huang},
  {and} \bibinfo{person}{Kevin Murphy}.} \bibinfo{year}{2018}\natexlab{c}.
\newblock \showarticletitle{Progressive Neural Architecture Search}. In
  \bibinfo{booktitle}{\emph{Proc. Eur. Conf. Comput. Vis.}}
  \bibinfo{pages}{19--34}.
\newblock
\showISSN{0302-9743}


\bibitem[Liu et~al\mbox{.}(2021a)]%
        {687r}
\bibfield{author}{\bibinfo{person}{Chia-Hsiang Liu}, \bibinfo{person}{Yu-Shin
  Han}, \bibinfo{person}{Yuan-Yao Sung}, \bibinfo{person}{Yi Lee},
  \bibinfo{person}{Hung-Yueh Chiang}, {and} \bibinfo{person}{Kai-Chiang Wu}.}
  \bibinfo{year}{2021}\natexlab{a}.
\newblock \showarticletitle{{FOX-NAS}: Fast, On-device and Explainable Neural
  Architecture Search}. In \bibinfo{booktitle}{\emph{Proc. IEEE Int. Conf.
  Comput. Vis.}} \bibinfo{pages}{789--797}.
\newblock


\bibitem[Liu et~al\mbox{.}(2018b)]%
        {694r}
\bibfield{author}{\bibinfo{person}{Hanxiao Liu}, \bibinfo{person}{Karen
  Simonyan}, {and} \bibinfo{person}{Yiming Yang}.}
  \bibinfo{year}{2018}\natexlab{b}.
\newblock \showarticletitle{{DARTS}: Differentiable Architecture Search}. In
  \bibinfo{booktitle}{\emph{Proc. Int. Conf. Learn. Represent.}}
  \bibinfo{pages}{https://arxiv.org/abs/1806.09055}.
\newblock


\bibitem[Liu et~al\mbox{.}(2019a)]%
        {577r}
\bibfield{author}{\bibinfo{person}{Peng Liu}, \bibinfo{person}{Mohammad D.~El
  Basha}, \bibinfo{person}{Yangjunyi Li}, \bibinfo{person}{Yao Xiao},
  \bibinfo{person}{Pina~C. Sanelli}, {and} \bibinfo{person}{Ruogu Fang}.}
  \bibinfo{year}{2019}\natexlab{a}.
\newblock \showarticletitle{Deep Evolutionary Networks with Expedited Genetic
  Algorithms for Medical Image Denoising}.
\newblock \bibinfo{journal}{\emph{Med. Image Anal.}}  \bibinfo{volume}{54}
  (\bibinfo{year}{2019}), \bibinfo{pages}{306--315}.
\newblock
\showISSN{1361-8415}


\bibitem[Liu and Guo(2021)]%
        {813r}
\bibfield{author}{\bibinfo{person}{Sicong Liu} {and} \bibinfo{person}{Bin
  Guo}.} \bibinfo{year}{2021}\natexlab{}.
\newblock \showarticletitle{Ada{S}pring: Context-adaptive and
  Runtime-evolutionary Deep Model Compression for Mobile Applications}. In
  \bibinfo{booktitle}{\emph{Proc. ACM Interact., Mobile, Wearable Ubiquitous
  Tech.}}, Vol.~\bibinfo{volume}{5}. \bibinfo{publisher}{ACM},
  \bibinfo{pages}{1--22}.
\newblock


\bibitem[Liu et~al\mbox{.}(2018a)]%
        {210r}
\bibfield{author}{\bibinfo{person}{Xiao-Ying Liu}, \bibinfo{person}{Yong
  Liang}, \bibinfo{person}{Sai Wang}, \bibinfo{person}{Zi-Yi Yang}, {and}
  \bibinfo{person}{Han-Shuo Ye}.} \bibinfo{year}{2018}\natexlab{a}.
\newblock \showarticletitle{A Hybrid Genetic Algorithm With Wrapper-Embedded
  Approaches for Feature Selection}.
\newblock \bibinfo{journal}{\emph{IEEE Access}}  \bibinfo{volume}{6}
  (\bibinfo{year}{2018}), \bibinfo{pages}{22863--22874}.
\newblock
\showISSN{2169-3536}


\bibitem[Liu et~al\mbox{.}(2021b)]%
        {037r}
\bibfield{author}{\bibinfo{person}{Yuqiao Liu}, \bibinfo{person}{Yanan Sun},
  \bibinfo{person}{Bing Xue}, \bibinfo{person}{Mengjie Zhang},
  \bibinfo{person}{Gary~G. Yen}, {and} \bibinfo{person}{Kay~Chen Tan}.}
  \bibinfo{year}{2021}\natexlab{b}.
\newblock \showarticletitle{A Survey on Evolutionary Neural Architecture
  Search}.
\newblock \bibinfo{journal}{\emph{IEEE Trans. Neural Netw. Learn. Syst.}}
  (\bibinfo{year}{2021}).
\newblock
\urldef\tempurl%
\url{https://doi.org/10.1109/TNNLS.2021.3100554}
\showDOI{\tempurl}


\bibitem[Liu et~al\mbox{.}(2019b)]%
        {768r}
\bibfield{author}{\bibinfo{person}{Zechun Liu}, \bibinfo{person}{Haoyuan Mu},
  \bibinfo{person}{Xiangyu Zhang}, \bibinfo{person}{Zichao Guo},
  \bibinfo{person}{Xin Yang}, \bibinfo{person}{K. Cheng}, {and}
  \bibinfo{person}{Jian Sun}.} \bibinfo{year}{2019}\natexlab{b}.
\newblock \showarticletitle{Meta{P}runing: Meta Learning for Automatic Neural
  Network Channel Pruning}. In \bibinfo{booktitle}{\emph{Proc. IEEE Int. Conf.
  Comput. Vis.}} \bibinfo{pages}{3295--3304}.
\newblock


\bibitem[Lomurno et~al\mbox{.}(2021)]%
        {686r}
\bibfield{author}{\bibinfo{person}{Eugenio Lomurno}, \bibinfo{person}{Stefano
  Samele}, \bibinfo{person}{Matteo Matteucci}, {and} \bibinfo{person}{Danilo
  Ardagna}.} \bibinfo{year}{2021}\natexlab{}.
\newblock \showarticletitle{Pareto-optimal Progressive Neural Architecture
  Search}. In \bibinfo{booktitle}{\emph{Proc. Genetic Evol. Comput. Conf.}}
  \bibinfo{pages}{1726--1734}.
\newblock


\bibitem[Londt et~al\mbox{.}(2021)]%
        {z15}
\bibfield{author}{\bibinfo{person}{Trevor Londt}, \bibinfo{person}{Xiaoying
  Gao}, {and} \bibinfo{person}{Peter Andreae}.}
  \bibinfo{year}{2021}\natexlab{}.
\newblock \showarticletitle{Evolving Character-level DenseNet Architectures
  Using Genetic Programming}. In \bibinfo{booktitle}{\emph{Proc. Int. Conf.
  Appl. Evol. Comput.}} \bibinfo{pages}{665--680}.
\newblock


\bibitem[Londt et~al\mbox{.}(2020)]%
        {z16}
\bibfield{author}{\bibinfo{person}{Trevor Londt}, \bibinfo{person}{Xiaoying
  Gao}, \bibinfo{person}{Bing Xue}, {and} \bibinfo{person}{Peter Andreae}.}
  \bibinfo{year}{2020}\natexlab{}.
\newblock \showarticletitle{Evolving Character-level Convolutional Neural
  Networks for Text Classification}.
\newblock \bibinfo{journal}{\emph{arXiv preprint arXiv:2012.02223}}
  (\bibinfo{year}{2020}).
\newblock


\bibitem[Loni et~al\mbox{.}(2020)]%
        {607r}
\bibfield{author}{\bibinfo{person}{Mohammad Loni}, \bibinfo{person}{Sima
  Sinaei}, {and} \bibinfo{person}{Ali Zoljodi}.}
  \bibinfo{year}{2020}\natexlab{}.
\newblock \showarticletitle{Deep{M}aker: A Multi-Objective Optimization
  Framework for Deep Neural Networks in Embedded Systems}.
\newblock \bibinfo{journal}{\emph{Microprocess. Microsyst.}}
  \bibinfo{volume}{73} (\bibinfo{year}{2020}), \bibinfo{pages}{102989}.
\newblock
\showISSN{0141-9331}


\bibitem[Lorenzo and Nalepa(2018)]%
        {662r}
\bibfield{author}{\bibinfo{person}{Pablo~Ribalta Lorenzo} {and}
  \bibinfo{person}{Jakub Nalepa}.} \bibinfo{year}{2018}\natexlab{}.
\newblock \showarticletitle{Memetic Evolution of Deep Neural Networks}. In
  \bibinfo{booktitle}{\emph{Proc. Genetic Evol. Comput. Conf.}}
  \bibinfo{pages}{505--512}.
\newblock


\bibitem[Lorenzo et~al\mbox{.}(2017)]%
        {682r}
\bibfield{author}{\bibinfo{person}{Pablo~Ribalta Lorenzo},
  \bibinfo{person}{Jakub Nalepa}, \bibinfo{person}{Luciano~S{\'a}nchez Ramos},
  {and} \bibinfo{person}{Jos{\'e} Ranilla}.} \bibinfo{year}{2017}\natexlab{}.
\newblock \showarticletitle{Hyper-parameter Selection in Deep Neural Networks
  Using Parallel Particle Swarm Optimization}. In
  \bibinfo{booktitle}{\emph{Proc. Genetic Evol. Comput. Conf.}}
  \bibinfo{pages}{1864--1871}.
\newblock


\bibitem[Lu et~al\mbox{.}(2021)]%
        {032r}
\bibfield{author}{\bibinfo{person}{Zhichao Lu}, \bibinfo{person}{Gautam
  Sreekumar}, \bibinfo{person}{Erik Goodman}, \bibinfo{person}{Wolfgang
  Banzhaf}, \bibinfo{person}{Kalyanmoy Deb}, {and}
  \bibinfo{person}{Vishnu~Naresh Boddeti}.} \bibinfo{year}{2021}\natexlab{}.
\newblock \showarticletitle{Neural Architecture Transfer}.
\newblock \bibinfo{journal}{\emph{IEEE IEEE Trans. Pattern Anal. Mach.
  Intell.}} \bibinfo{volume}{43}, \bibinfo{number}{9} (\bibinfo{year}{2021}),
  \bibinfo{pages}{2971--2989}.
\newblock
\showISSN{0162-8828}


\bibitem[Lu et~al\mbox{.}(2019)]%
        {611r}
\bibfield{author}{\bibinfo{person}{Zhichao Lu}, \bibinfo{person}{Ian Whalen},
  \bibinfo{person}{Vishnu~Naresh Boddeti}, \bibinfo{person}{Yashesh~D. Dhebar},
  {and} \bibinfo{person}{Kalyanmoy Deb}.} \bibinfo{year}{2019}\natexlab{}.
\newblock \showarticletitle{{NSGA-N}et: Neural Architecture Search using
  Multi-objective Genetic Algorithm}. In \bibinfo{booktitle}{\emph{Proc.
  Genetic Evol. Comput. Conf.}} \bibinfo{pages}{419--427}.
\newblock


\bibitem[Luo et~al\mbox{.}(2018)]%
        {127r}
\bibfield{author}{\bibinfo{person}{Renqian Luo}, \bibinfo{person}{Fei Tian},
  \bibinfo{person}{Tao Qin}, \bibinfo{person}{Enhong Chen}, {and}
  \bibinfo{person}{Tie-Yan Liu}.} \bibinfo{year}{2018}\natexlab{}.
\newblock \showarticletitle{Neural architecture optimization}. In
  \bibinfo{booktitle}{\emph{Proc. Adv. Neural Inf. Process. Syst.}},
  Vol.~\bibinfo{volume}{31}. \bibinfo{pages}{7827--7838}.
\newblock


\bibitem[Ma et~al\mbox{.}(2021b)]%
        {633r}
\bibfield{author}{\bibinfo{person}{Ailong Ma}, \bibinfo{person}{Yuting Wan},
  \bibinfo{person}{Yanfei Zhong}, \bibinfo{person}{Junjue Wang}, {and}
  \bibinfo{person}{Liang pei Zhang}.} \bibinfo{year}{2021}\natexlab{b}.
\newblock \showarticletitle{Scene{N}et: Remote Sensing Scene Classification
  Deep Learning Network Using Multi-Objective Neural Evolution Architecture
  Search}.
\newblock \bibinfo{journal}{\emph{ISPRS J. Photogramm. Remote Sens.}}
  \bibinfo{volume}{172} (\bibinfo{year}{2021}), \bibinfo{pages}{171--188}.
\newblock
\showISSN{0924-2716}


\bibitem[Ma et~al\mbox{.}(2022)]%
        {002r}
\bibfield{author}{\bibinfo{person}{Lianbo Ma}, \bibinfo{person}{Min Huang},
  \bibinfo{person}{Shengxiang Yang}, \bibinfo{person}{Rui Wang}, {and}
  \bibinfo{person}{Xingwei Wang}.} \bibinfo{year}{2022}\natexlab{}.
\newblock \showarticletitle{An Adaptive Localized Decision Variable Analysis
  Approach to Large-Scale Multiobjective and Many-Objective Optimization}.
\newblock \bibinfo{journal}{\emph{IEEE Trans. Cybern.}} \bibinfo{volume}{52},
  \bibinfo{number}{7} (\bibinfo{year}{2022}), \bibinfo{pages}{6684--6696}.
\newblock


\bibitem[Ma et~al\mbox{.}(2021a)]%
        {677r}
\bibfield{author}{\bibinfo{person}{Lianbo Ma}, \bibinfo{person}{Nan Li},
  \bibinfo{person}{Guo Yu}, \bibinfo{person}{Xiaoyu Geng}, \bibinfo{person}{Min
  Huang}, {and} \bibinfo{person}{Xingwei Wang}.}
  \bibinfo{year}{2021}\natexlab{a}.
\newblock \showarticletitle{How to Simplify Search: Classification-wise Pareto
  Evolution for One-shot Neural Architecture Search}.
\newblock \bibinfo{journal}{\emph{arXiv preprint arXiv:2109.07582}}
  (\bibinfo{year}{2021}).
\newblock


\bibitem[Ma et~al\mbox{.}(2021c)]%
        {205r}
\bibfield{author}{\bibinfo{person}{Wenping Ma}, \bibinfo{person}{Xiaobo Zhou},
  \bibinfo{person}{Hao Zhu}, \bibinfo{person}{Longwei Li}, {and}
  \bibinfo{person}{Licheng Jiao}.} \bibinfo{year}{2021}\natexlab{c}.
\newblock \showarticletitle{A Two-stage Hybrid Ant Colony Optimization for
  High-dimensional Feature Selection}.
\newblock \bibinfo{journal}{\emph{Pattern Recognit.}}  \bibinfo{volume}{116}
  (\bibinfo{year}{2021}), \bibinfo{pages}{107933}.
\newblock


\bibitem[Maniezzo(1994)]%
        {701r}
\bibfield{author}{\bibinfo{person}{V. Maniezzo}.}
  \bibinfo{year}{1994}\natexlab{}.
\newblock \showarticletitle{Genetic Evolution of the Topology and Weight
  Distribution of Neural Networks}.
\newblock \bibinfo{journal}{\emph{IEEE Trans. Neural. Netw.}}
  \bibinfo{volume}{5}, \bibinfo{number}{1} (\bibinfo{year}{1994}),
  \bibinfo{pages}{39--53}.
\newblock


\bibitem[Mauceri et~al\mbox{.}(2021)]%
        {445r}
\bibfield{author}{\bibinfo{person}{Stefano Mauceri}, \bibinfo{person}{James
  Sweeney}, \bibinfo{person}{Miguel Nicolau}, {and} \bibinfo{person}{James
  McDermott}.} \bibinfo{year}{2021}\natexlab{}.
\newblock \showarticletitle{Feature Extraction by Grammatical Evolution for
  One-class Time Series Classification}.
\newblock \bibinfo{journal}{\emph{Genet. Program. Evolvable Mach.}}
  \bibinfo{volume}{22}, \bibinfo{number}{3} (\bibinfo{year}{2021}),
  \bibinfo{pages}{267--295}.
\newblock
\showISSN{1389-2576}


\bibitem[Mazzawi et~al\mbox{.}(2019)]%
        {893r}
\bibfield{author}{\bibinfo{person}{Hanna Mazzawi}, \bibinfo{person}{Xavi
  Gonzalvo}, \bibinfo{person}{Aleksandar Kracun}, {and}
  \bibinfo{person}{Prashant Sridhar}.} \bibinfo{year}{2019}\natexlab{}.
\newblock \showarticletitle{Improving Keyword Spotting and Language
  Identification via Neural Architecture Search at Scale}. In
  \bibinfo{booktitle}{\emph{INTERSPEECH}}. \bibinfo{pages}{1278--1282}.
\newblock


\bibitem[Mei et~al\mbox{.}(2017)]%
        {267r}
\bibfield{author}{\bibinfo{person}{Yi Mei}, \bibinfo{person}{Su Nguyen},
  \bibinfo{person}{Bing Xue}, {and} \bibinfo{person}{Mengjie Zhang}.}
  \bibinfo{year}{2017}\natexlab{}.
\newblock \showarticletitle{An Efficient Feature Selection Algorithm for
  Evolving Job Shop Scheduling Rules With Genetic Programming}.
\newblock \bibinfo{journal}{\emph{IEEE Trans. Emerg. Topics Comput. Intell.}}
  \bibinfo{volume}{1}, \bibinfo{number}{5} (\bibinfo{year}{2017}),
  \bibinfo{pages}{339--353}.
\newblock


\bibitem[Miahi et~al\mbox{.}(2022)]%
        {077r}
\bibfield{author}{\bibinfo{person}{Erfan Miahi},
  \bibinfo{person}{Seyed~Abolghasem Mirroshandel}, {and}
  \bibinfo{person}{Alexis Nasr}.} \bibinfo{year}{2022}\natexlab{}.
\newblock \showarticletitle{Genetic Neural Architecture Search for Automatic
  Assessment of Human Sperm Images}.
\newblock \bibinfo{journal}{\emph{Expert Syst. Appl.}}  \bibinfo{volume}{188}
  (\bibinfo{year}{2022}), \bibinfo{pages}{115937}.
\newblock
\showISSN{0957-4174}


\bibitem[Miikkulainen et~al\mbox{.}(2019)]%
        {129r}
\bibfield{author}{\bibinfo{person}{Risto Miikkulainen}, \bibinfo{person}{Jason
  Liang}, \bibinfo{person}{Elliot Meyerson}, \bibinfo{person}{Aditya Rawal},
  \bibinfo{person}{Daniel Fink}, \bibinfo{person}{Olivier Francon},
  \bibinfo{person}{Bala Raju}, \bibinfo{person}{Hormoz Shahrzad},
  \bibinfo{person}{Arshak Navruzyan}, \bibinfo{person}{Nigel Duffy},
  {et~al\mbox{.}}} \bibinfo{year}{2019}\natexlab{}.
\newblock \showarticletitle{Evolving Deep Neural Networks}.
\newblock In \bibinfo{booktitle}{\emph{Artificial Intelligence in the Age Of
  Neural Networks and Brain Computing}}. \bibinfo{publisher}{Elsevier},
  \bibinfo{pages}{293--312}.
\newblock


\bibitem[Mirjalili et~al\mbox{.}(2019)]%
        {048r}
\bibfield{author}{\bibinfo{person}{Seyedali Mirjalili}, \bibinfo{person}{Hossam
  Faris}, {and} \bibinfo{person}{Ibrahim Aljarah}.}
  \bibinfo{year}{2019}\natexlab{}.
\newblock \bibinfo{booktitle}{\emph{Evolutionary Machine Learning Techniques}}.
\newblock \bibinfo{publisher}{Springer}.
\newblock


\bibitem[Mo et~al\mbox{.}(2021)]%
        {115r}
\bibfield{author}{\bibinfo{person}{Hyunho Mo}, \bibinfo{person}{Leonardo~Lucio
  Custode}, {and} \bibinfo{person}{Giovanni Iacca}.}
  \bibinfo{year}{2021}\natexlab{}.
\newblock \showarticletitle{Evolutionary Neural Architecture Search for
  Remaining Useful Life Prediction}.
\newblock \bibinfo{journal}{\emph{Appl. Soft Comput.}}  \bibinfo{volume}{108}
  (\bibinfo{year}{2021}), \bibinfo{pages}{107474}.
\newblock
\showISSN{1568-4946}


\bibitem[Montana and Davis(1989)]%
        {471r}
\bibfield{author}{\bibinfo{person}{David~J Montana} {and}
  \bibinfo{person}{Lawrence Davis}.} \bibinfo{year}{1989}\natexlab{}.
\newblock \showarticletitle{Training Feedforward Neural Networks Using Genetic
  Algorithms}. In \bibinfo{booktitle}{\emph{Proc. of the Int. Joint Conf.
  Artif. Intell.}}, Vol.~\bibinfo{volume}{4}. \bibinfo{pages}{762--767}.
\newblock


\bibitem[Muni et~al\mbox{.}(2006)]%
        {188r}
\bibfield{author}{\bibinfo{person}{Durga~Prasad Muni},
  \bibinfo{person}{Nikhil~R Pal}, {and} \bibinfo{person}{Jyotirmay Das}.}
  \bibinfo{year}{2006}\natexlab{}.
\newblock \showarticletitle{Genetic Programming for Simultaneous Feature
  Selection and Classifier Design}.
\newblock \bibinfo{journal}{\emph{IEEE Trans. Syst. Man. Cybern.}}
  \bibinfo{volume}{36}, \bibinfo{number}{1} (\bibinfo{year}{2006}),
  \bibinfo{pages}{106--117}.
\newblock
\showISSN{1083-4419}


\bibitem[Murray and Chiang(2015)]%
        {904r}
\bibfield{author}{\bibinfo{person}{Kenton Murray} {and} \bibinfo{person}{David
  Chiang}.} \bibinfo{year}{2015}\natexlab{}.
\newblock \showarticletitle{Auto-Sizing Neural Networks: With Applications to
  N-Gram Language Models}. In \bibinfo{booktitle}{\emph{Proc. Conf. Empir.
  Methods Nat. Lang. Proc.}} \bibinfo{pages}{908--916}.
\newblock


\bibitem[Nekrasov et~al\mbox{.}(2020)]%
        {147r}
\bibfield{author}{\bibinfo{person}{Vladimir Nekrasov}, \bibinfo{person}{Chunhua
  Shen}, {and} \bibinfo{person}{Ian Reid}.} \bibinfo{year}{2020}\natexlab{}.
\newblock \showarticletitle{Template-Based Automatic Search of Compact Semantic
  Segmentation Architectures}. In \bibinfo{booktitle}{\emph{Proc. Winter Conf.
  Appl. Comput. Vis.}} \bibinfo{pages}{1980--1989}.
\newblock


\bibitem[Neshat et~al\mbox{.}(2020)]%
        {599r}
\bibfield{author}{\bibinfo{person}{Mehdi Neshat},
  \bibinfo{person}{Meysam~Majidi Nezhad}, \bibinfo{person}{Ehsan Abbasnejad},
  \bibinfo{person}{Lina~Bertling Tjernberg}, \bibinfo{person}{Davide~Astiaso
  Garcia}, \bibinfo{person}{Bradley Alexander}, {and} \bibinfo{person}{Markus
  Wagner}.} \bibinfo{year}{2020}\natexlab{}.
\newblock \showarticletitle{An Evolutionary Deep Learning Method for Short-term
  Wind Speed Prediction: A Case Study of the Lillgrund Offshore Wind Farm}.
\newblock \bibinfo{journal}{\emph{arXiv preprint arXiv:abs/2002.09106}}
  (\bibinfo{year}{2020}).
\newblock


\bibitem[Neshatian et~al\mbox{.}(2012)]%
        {z04}
\bibfield{author}{\bibinfo{person}{Kourosh Neshatian}, \bibinfo{person}{Mengjie
  Zhang}, {and} \bibinfo{person}{Peter Andreae}.}
  \bibinfo{year}{2012}\natexlab{}.
\newblock \showarticletitle{A Filter Approach to Multiple Feature Construction
  for Symbolic Learning Classifiers Using Genetic Programming}.
\newblock \bibinfo{journal}{\emph{IEEE Trans. Evol. Comput.}}
  \bibinfo{volume}{16}, \bibinfo{number}{5} (\bibinfo{year}{2012}),
  \bibinfo{pages}{645--661}.
\newblock


\bibitem[Neshatian et~al\mbox{.}(2007)]%
        {374r}
\bibfield{author}{\bibinfo{person}{Kourosh Neshatian}, \bibinfo{person}{Mengjie
  Zhang}, {and} \bibinfo{person}{Mark Johnston}.}
  \bibinfo{year}{2007}\natexlab{}.
\newblock \showarticletitle{Feature Construction and Dimension Reduction Using
  Genetic Programming}. In \bibinfo{booktitle}{\emph{Proc. Aust. Conf. Artif.
  Intell.}} \bibinfo{pages}{242--253}.
\newblock
\showISSN{0302-9743}


\bibitem[Nguyen et~al\mbox{.}(2014)]%
        {172r}
\bibfield{author}{\bibinfo{person}{Hoai~Bach Nguyen}, \bibinfo{person}{Bing
  Xue}, \bibinfo{person}{Ivy Liu}, {and} \bibinfo{person}{Mengjie Zhang}.}
  \bibinfo{year}{2014}\natexlab{}.
\newblock \showarticletitle{{PSO} and Statistical Clustering for Feature
  Selection: A New Representation}. In \bibinfo{booktitle}{\emph{Proc.
  Asia-Pacific Conf. Simulated Evol. Learn.}} \bibinfo{pages}{569--581}.
\newblock
\showISSN{0302-9743}


\bibitem[O'Boyle and Palmer(2008)]%
        {190r}
\bibfield{author}{\bibinfo{person}{Noel~M O'Boyle} {and}
  \bibinfo{person}{David~S Palmer}.} \bibinfo{year}{2008}\natexlab{}.
\newblock \showarticletitle{Simultaneous Feature Selection and Parameter
  Optimisation Using An Artificial Ant Colony: Case Study of Melting Point
  Prediction}.
\newblock \bibinfo{journal}{\emph{Chem. Cent. J.}} \bibinfo{volume}{2},
  \bibinfo{number}{1} (\bibinfo{year}{2008}), \bibinfo{pages}{1--15}.
\newblock
\showISSN{1752-153X}


\bibitem[Olson and Moore(2016)]%
        {789r}
\bibfield{author}{\bibinfo{person}{Randal~S. Olson} {and}
  \bibinfo{person}{Jason~H. Moore}.} \bibinfo{year}{2016}\natexlab{}.
\newblock \showarticletitle{TPOT: A Tree-based Pipeline Optimization Tool for
  Automating Machine Learning}. In \bibinfo{booktitle}{\emph{Proc. Int. Conf.
  Mach. Learn.}} \bibinfo{pages}{151--160}.
\newblock


\bibitem[O'Neill et~al\mbox{.}(2018)]%
        {648r}
\bibfield{author}{\bibinfo{person}{Damien O'Neill}, \bibinfo{person}{Bing Xue},
  {and} \bibinfo{person}{Mengjie Zhang}.} \bibinfo{year}{2018}\natexlab{}.
\newblock \showarticletitle{Co-evolution of Novel Tree-Like ANNs and Activation
  Functions: An Observational Study}. In \bibinfo{booktitle}{\emph{Proc. Aust.
  Conf. Artif. Intell.}} \bibinfo{pages}{616--629}.
\newblock
\showISSN{0302-9743}


\bibitem[Oong and Isa(2011)]%
        {515r}
\bibfield{author}{\bibinfo{person}{Tatt~Hee Oong} {and} \bibinfo{person}{Nor
  Ashidi~Mat Isa}.} \bibinfo{year}{2011}\natexlab{}.
\newblock \showarticletitle{Adaptive Evolutionary Artificial Neural Networks
  for Pattern Classification}.
\newblock \bibinfo{journal}{\emph{IEEE Trans. Neural Netw.}}
  \bibinfo{volume}{22}, \bibinfo{number}{11} (\bibinfo{year}{2011}),
  \bibinfo{pages}{1823--1836}.
\newblock


\bibitem[Ortego et~al\mbox{.}(2020)]%
        {587r}
\bibfield{author}{\bibinfo{person}{Patxi Ortego}, \bibinfo{person}{Alberto
  Diez-Olivan}, \bibinfo{person}{Javier~Del Ser}, {and}
  \bibinfo{person}{Fernando Veiga}.} \bibinfo{year}{2020}\natexlab{}.
\newblock \showarticletitle{Evolutionary {LSTM-FCN} Networks for Pattern
  Classification in Industrial Processes}.
\newblock \bibinfo{journal}{\emph{Swarm Evol. Comput.}}  \bibinfo{volume}{54}
  (\bibinfo{year}{2020}), \bibinfo{pages}{100650}.
\newblock
\showISSN{2210-6502}


\bibitem[Peng et~al\mbox{.}(2021)]%
        {434r}
\bibfield{author}{\bibinfo{person}{Bo Peng}, \bibinfo{person}{Shuting Wan},
  \bibinfo{person}{Ying Bi}, \bibinfo{person}{Bing Xue}, {and}
  \bibinfo{person}{Mengjie Zhang}.} \bibinfo{year}{2021}\natexlab{}.
\newblock \showarticletitle{Automatic Feature Extraction and Construction Using
  Genetic Programming for Rotating Machinery Fault Diagnosis}.
\newblock \bibinfo{journal}{\emph{IEEE Trans. Cybern.}} \bibinfo{volume}{51},
  \bibinfo{number}{10} (\bibinfo{year}{2021}), \bibinfo{pages}{4909--4923}.
\newblock
\showISSN{2168-2267}


\bibitem[Peng et~al\mbox{.}(2018)]%
        {z08}
\bibfield{author}{\bibinfo{person}{Yiming Peng}, \bibinfo{person}{Gang Chen},
  \bibinfo{person}{Harman Singh}, {and} \bibinfo{person}{Mengjie Zhang}.}
  \bibinfo{year}{2018}\natexlab{}.
\newblock \showarticletitle{{NEAT} for Large-scale Reinforcement Learning
  Through Evolutionary Feature Learning and Policy Gradient Search}. In
  \bibinfo{booktitle}{\emph{Proc. Genetic Evol. Comput. Conf.}}
  \bibinfo{pages}{490--497}.
\newblock


\bibitem[Phan et~al\mbox{.}(2020)]%
        {803r}
\bibfield{author}{\bibinfo{person}{Hai~T. Phan}, \bibinfo{person}{Zechun Liu},
  \bibinfo{person}{Dang~The Huynh}, \bibinfo{person}{Marios Savvides},
  \bibinfo{person}{Kwang-Ting Cheng}, {and} \bibinfo{person}{Zhiqiang Shen}.}
  \bibinfo{year}{2020}\natexlab{}.
\newblock \showarticletitle{Binarizing MobileNet via Evolution-Based
  Searching}. In \bibinfo{booktitle}{\emph{Proc. IEEE Conf. Comput. Vis.
  Pattern Recognit.}} \bibinfo{pages}{13417--13426}.
\newblock


\bibitem[Polino et~al\mbox{.}(2018)]%
        {804r}
\bibfield{author}{\bibinfo{person}{Antonio Polino}, \bibinfo{person}{Razvan
  Pascanu}, {and} \bibinfo{person}{Dan Alistarh}.}
  \bibinfo{year}{2018}\natexlab{}.
\newblock \showarticletitle{Model Compression via Distillation and
  Quantization}. In \bibinfo{booktitle}{\emph{Proc. Int. Conf. Learn.
  Represent.}} \bibinfo{pages}{https://arxiv.org/abs/1802.05668}.
\newblock


\bibitem[Poyatos et~al\mbox{.}(2022)]%
        {824r}
\bibfield{author}{\bibinfo{person}{Javier Poyatos}, \bibinfo{person}{Daniel
  Molina}, \bibinfo{person}{Aritz Martinez}, {et~al\mbox{.}}}
  \bibinfo{year}{2022}\natexlab{}.
\newblock \showarticletitle{Evo{P}rune{D}eep{TL}: An Evolutionary Pruning Model
  for Transfer Learning based Deep Neural Networks}.
\newblock \bibinfo{journal}{\emph{arXiv preprint arXiv:2202.03844}}
  (\bibinfo{year}{2022}).
\newblock


\bibitem[Radford et~al\mbox{.}(2018)]%
        {776r}
\bibfield{author}{\bibinfo{person}{Alec Radford}, \bibinfo{person}{Karthik
  Narasimhan}, \bibinfo{person}{Tim Salimans}, \bibinfo{person}{Ilya
  Sutskever}, {et~al\mbox{.}}} \bibinfo{year}{2018}\natexlab{}.
\newblock \showarticletitle{Improving Language Understanding by Generative
  Pre-training}.
\newblock  (\bibinfo{year}{2018}),
  \bibinfo{pages}{https://www.cs.ubc.ca/~amuham01/LING530/papers/radford2018improving.pdf}.
\newblock


\bibitem[Rapaport et~al\mbox{.}(2019)]%
        {558r}
\bibfield{author}{\bibinfo{person}{Elad Rapaport}, \bibinfo{person}{Oren
  Shriki}, {and} \bibinfo{person}{Rami Puzis}.}
  \bibinfo{year}{2019}\natexlab{}.
\newblock \showarticletitle{{EEGNAS}: Neural Architecture Search for
  Electroencephalography Data Analysis and Decoding}. In
  \bibinfo{booktitle}{\emph{Proc. Int. Joint Conf. Artif. Intell.}}
  \bibinfo{pages}{3--20}.
\newblock
\showISSN{1865-0929}


\bibitem[Rashid et~al\mbox{.}(2020)]%
        {197r}
\bibfield{author}{\bibinfo{person}{ANM~Bazlur Rashid},
  \bibinfo{person}{Mohiuddin Ahmed}, \bibinfo{person}{Leslie~F Sikos}, {and}
  \bibinfo{person}{Paul Haskell-Dowland}.} \bibinfo{year}{2020}\natexlab{}.
\newblock \showarticletitle{Cooperative Co-Evolution for Feature Selection in
  Big Data With Random Feature Grouping}.
\newblock \bibinfo{journal}{\emph{J. Big Data}} \bibinfo{volume}{7},
  \bibinfo{number}{1} (\bibinfo{year}{2020}), \bibinfo{pages}{1--42}.
\newblock
\showISSN{2196-1115}


\bibitem[Rawal and Miikkulainen(2018)]%
        {151r}
\bibfield{author}{\bibinfo{person}{Aditya Rawal} {and} \bibinfo{person}{Risto
  Miikkulainen}.} \bibinfo{year}{2018}\natexlab{}.
\newblock \showarticletitle{From Nodes to Networks: Evolving Recurrent Neural
  Networks}.
\newblock \bibinfo{journal}{\emph{arXiv preprint arXiv:1803.04439}}
  (\bibinfo{year}{2018}).
\newblock


\bibitem[Real et~al\mbox{.}(2019)]%
        {122r}
\bibfield{author}{\bibinfo{person}{Esteban Real}, \bibinfo{person}{Alok
  Aggarwal}, \bibinfo{person}{Yanping Huang}, {and} \bibinfo{person}{Quoc~V
  Le}.} \bibinfo{year}{2019}\natexlab{}.
\newblock \showarticletitle{Regularized Evolution for Image Classifier
  Architecture Search}. In \bibinfo{booktitle}{\emph{Proc. AAAI Conf. Artif.
  Intell.}}, Vol.~\bibinfo{volume}{33}. \bibinfo{pages}{4780--4789}.
\newblock
\showISSN{2374-3468}


\bibitem[Real et~al\mbox{.}(2017)]%
        {539r}
\bibfield{author}{\bibinfo{person}{Esteban Real}, \bibinfo{person}{Sherry
  Moore}, \bibinfo{person}{Andrew Selle}, \bibinfo{person}{Saurabh Saxena},
  \bibinfo{person}{Yutaka~Leon Suematsu}, \bibinfo{person}{Jie Tan},
  \bibinfo{person}{Quoc~V. Le}, {and} \bibinfo{person}{Alexey Kurakin}.}
  \bibinfo{year}{2017}\natexlab{}.
\newblock \showarticletitle{Large-Scale Evolution of Image Classifiers}. In
  \bibinfo{booktitle}{\emph{Proc. Int. Conf. Mach. Learn.}}
  \bibinfo{pages}{2902--2911}.
\newblock
\showISSN{2374-3468}


\bibitem[Refahi et~al\mbox{.}(2020)]%
        {413r}
\bibfield{author}{\bibinfo{person}{Mohammad~Saleh Refahi}, \bibinfo{person}{A
  Mir}, {and} \bibinfo{person}{Jalal~A Nasiri}.}
  \bibinfo{year}{2020}\natexlab{}.
\newblock \showarticletitle{A Novel Fusion Based on the Evolutionary Features
  for Protein Fold Recognition Using Support Vector Machines}.
\newblock \bibinfo{journal}{\emph{Sci. Rep.}} \bibinfo{volume}{10},
  \bibinfo{number}{1} (\bibinfo{year}{2020}), \bibinfo{pages}{1--13}.
\newblock


\bibitem[Ren et~al\mbox{.}(2021)]%
        {045r}
\bibfield{author}{\bibinfo{person}{Pengzhen Ren}, \bibinfo{person}{Yun Xiao},
  \bibinfo{person}{Xiaojun Chang}, \bibinfo{person}{Po-Yao Huang}, {and}
  \bibinfo{person}{Zhihui Li}.} \bibinfo{year}{2021}\natexlab{}.
\newblock \showarticletitle{A Comprehensive Survey of Neural Architecture
  Search: Challenges and Solutions}.
\newblock \bibinfo{journal}{\emph{ACM Comput. Surv.}} \bibinfo{volume}{54},
  \bibinfo{number}{4} (\bibinfo{year}{2021}), \bibinfo{pages}{1--34}.
\newblock
\showISSN{0360-0300}


\bibitem[Roberts and Claridge(2005)]%
        {378r}
\bibfield{author}{\bibinfo{person}{Mark~E. Roberts} {and} \bibinfo{person}{Ela
  Claridge}.} \bibinfo{year}{2005}\natexlab{}.
\newblock \showarticletitle{A Multistage Approach to Cooperatively Coevolving
  Feature Construction and Object Detection}. In
  \bibinfo{booktitle}{\emph{Proc. Appl. Evol. Comput.}}
  \bibinfo{pages}{396--406}.
\newblock
\showISSN{0302-9743}


\bibitem[Rostami and Neri(2016)]%
        {708r}
\bibfield{author}{\bibinfo{person}{Shahin Rostami} {and}
  \bibinfo{person}{Ferrante Neri}.} \bibinfo{year}{2016}\natexlab{}.
\newblock \showarticletitle{Covariance Matrix Adaptation Pareto Archived
  Evolution Strategy With Hypervolume-Sorted Adaptive Grid Algorithm}.
\newblock \bibinfo{journal}{\emph{Integr. Comput. Aided. Eng.}}
  \bibinfo{volume}{23}, \bibinfo{number}{4} (\bibinfo{year}{2016}),
  \bibinfo{pages}{313--329}.
\newblock
\showISSN{1069-2509}


\bibitem[Rumelhart et~al\mbox{.}(1985)]%
        {z11}
\bibfield{author}{\bibinfo{person}{David~E Rumelhart},
  \bibinfo{person}{Geoffrey~E Hinton}, {and} \bibinfo{person}{Ronald~J
  Williams}.} \bibinfo{year}{1985}\natexlab{}.
\newblock \bibinfo{booktitle}{\emph{Learning Internal Representations by Error
  Propagation}}.
\newblock \bibinfo{type}{{T}echnical {R}eport}.
  \bibinfo{institution}{California Univ San Diego La Jolla Inst for Cognitive
  Science}.
\newblock


\bibitem[Rumelhart et~al\mbox{.}(1986)]%
        {z12}
\bibfield{author}{\bibinfo{person}{David~E Rumelhart},
  \bibinfo{person}{Geoffrey~E Hinton}, {and} \bibinfo{person}{Ronald~J
  Williams}.} \bibinfo{year}{1986}\natexlab{}.
\newblock \showarticletitle{Learning Representations by Back-propagating
  Errors}.
\newblock \bibinfo{journal}{\emph{Nature}} \bibinfo{volume}{323},
  \bibinfo{number}{6088} (\bibinfo{year}{1986}), \bibinfo{pages}{533--536}.
\newblock


\bibitem[Rundo et~al\mbox{.}(2019)]%
        {075r}
\bibfield{author}{\bibinfo{person}{Leonardo Rundo}, \bibinfo{person}{Andrea
  Tangherloni}, \bibinfo{person}{Marco~S Nobile}, \bibinfo{person}{Carmelo
  Militello}, \bibinfo{person}{Daniela Besozzi}, \bibinfo{person}{Giancarlo
  Mauri}, {and} \bibinfo{person}{Paolo Cazzaniga}.}
  \bibinfo{year}{2019}\natexlab{}.
\newblock \showarticletitle{Med{GA}: A Novel Evolutionary Method for Image
  Enhancement in Medical Imaging Systems}.
\newblock \bibinfo{journal}{\emph{Expert Syst. Appl.}}  \bibinfo{volume}{119}
  (\bibinfo{year}{2019}), \bibinfo{pages}{387--399}.
\newblock
\showISSN{0957-4174}


\bibitem[Samala et~al\mbox{.}(2018)]%
        {818r}
\bibfield{author}{\bibinfo{person}{Ravi~K. Samala}, \bibinfo{person}{Heang-Ping
  Chan}, \bibinfo{person}{Lubomir~M. Hadjiiski}, \bibinfo{person}{Mark~A.
  Helvie}, \bibinfo{person}{Caleb~D. Richter}, {and} \bibinfo{person}{Kenny~H.
  Cha}.} \bibinfo{year}{2018}\natexlab{}.
\newblock \showarticletitle{Evolutionary Pruning of Transfer Learned Deep
  Convolutional Neural Network For Breast Cancer Diagnosis In Digital Breast
  Tomosynthesis.}
\newblock \bibinfo{journal}{\emph{Phys. Med. Biol.}} \bibinfo{volume}{63},
  \bibinfo{number}{9} (\bibinfo{year}{2018}), \bibinfo{pages}{095005}.
\newblock
\showISSN{1361-6560}


\bibitem[Santra et~al\mbox{.}(2021)]%
        {006r}
\bibfield{author}{\bibinfo{person}{Santanu Santra}, \bibinfo{person}{Jun-Wei
  Hsieh}, {and} \bibinfo{person}{Chi-Fang Lin}.}
  \bibinfo{year}{2021}\natexlab{}.
\newblock \showarticletitle{Gradient Descent Effects on Differential Neural
  Architecture Search: A Survey}.
\newblock \bibinfo{journal}{\emph{IEEE Access}}  \bibinfo{volume}{9}
  (\bibinfo{year}{2021}), \bibinfo{pages}{89602--89618}.
\newblock


\bibitem[Sapra and Pimentel(2020)]%
        {661r}
\bibfield{author}{\bibinfo{person}{Dolly Sapra} {and} \bibinfo{person}{Andy~D
  Pimentel}.} \bibinfo{year}{2020}\natexlab{}.
\newblock \showarticletitle{Constrained Evolutionary Piecemeal Training to
  Design Convolutional Neural Networks}. In \bibinfo{booktitle}{\emph{Proc.
  Int. Conf. Industr., Eng. and Other Appl. of App. Intell. Syst.}}
  \bibinfo{pages}{709--721}.
\newblock


\bibitem[Schorn et~al\mbox{.}(2020)]%
        {088r}
\bibfield{author}{\bibinfo{person}{Christoph Schorn}, \bibinfo{person}{Thomas
  Elsken}, \bibinfo{person}{Sebastian Vogel}, {and} \bibinfo{person}{Armin
  Runge}.} \bibinfo{year}{2020}\natexlab{}.
\newblock \showarticletitle{Automated Design Of Error-Resilient and
  Hardware-Efficient Deep Neural Networks}.
\newblock \bibinfo{journal}{\emph{Neural. Comput. Appl.}} \bibinfo{volume}{32},
  \bibinfo{number}{24} (\bibinfo{year}{2020}), \bibinfo{pages}{18327--18345}.
\newblock
\showISSN{0941-0643}


\bibitem[Sciuto et~al\mbox{.}(2020)]%
        {785r}
\bibfield{author}{\bibinfo{person}{Christian Sciuto}, \bibinfo{person}{Kaicheng
  Yu}, \bibinfo{person}{Martin Jaggi}, \bibinfo{person}{Claudiu~Cristian
  Musat}, {and} \bibinfo{person}{Mathieu Salzmann}.}
  \bibinfo{year}{2020}\natexlab{}.
\newblock \showarticletitle{Evaluating the Search Phase of Neural Architecture
  Search}. In \bibinfo{booktitle}{\emph{Proc. Int. Conf. Learn. Represent.}}
  \bibinfo{pages}{https://arxiv.org/abs/1902.08142}.
\newblock


\bibitem[Shafti and P{\'e}rez(2008)]%
        {370r}
\bibfield{author}{\bibinfo{person}{Leila~Shila Shafti} {and}
  \bibinfo{person}{E.~Islas P{\'e}rez}.} \bibinfo{year}{2008}\natexlab{}.
\newblock \showarticletitle{Data Reduction by Genetic Algorithms and
  Non-Algebraic Feature Construction: A Case Study}.
\newblock \bibinfo{journal}{\emph{Proc. Int. Conf. Hybri. Intell. Syst.}}
  (\bibinfo{year}{2008}), \bibinfo{pages}{573--578}.
\newblock


\bibitem[Shakya et~al\mbox{.}(2021)]%
        {444r}
\bibfield{author}{\bibinfo{person}{Pratistha Shakya}, \bibinfo{person}{Eamonn
  Kennedy}, \bibinfo{person}{Christopher Rose}, {and} \bibinfo{person}{Jacob~K.
  Rotein}.} \bibinfo{year}{2021}\natexlab{}.
\newblock \showarticletitle{High-Dimensional Time Series Feature Extraction for
  Low-Cost Machine Olfaction}.
\newblock \bibinfo{journal}{\emph{IEEE Sens. J.}} \bibinfo{volume}{21},
  \bibinfo{number}{3} (\bibinfo{year}{2021}), \bibinfo{pages}{2495--2504}.
\newblock
\showISSN{1424-8220}


\bibitem[Shang et~al\mbox{.}(2022)]%
        {832r}
\bibfield{author}{\bibinfo{person}{Haopu Shang}, \bibinfo{person}{Jia-Liang
  Wu}, \bibinfo{person}{Wenjing Hong}, {and} \bibinfo{person}{Chao Qian}.}
  \bibinfo{year}{2022}\natexlab{}.
\newblock \showarticletitle{Neural Network Pruning by Cooperative Coevolution}.
\newblock \bibinfo{journal}{\emph{arXiv preprint arXiv:2204.05639}}
  (\bibinfo{year}{2022}).
\newblock


\bibitem[Shen et~al\mbox{.}(2019)]%
        {644r}
\bibfield{author}{\bibinfo{person}{Mingzhu Shen}, \bibinfo{person}{Kai Han},
  \bibinfo{person}{Chunjing Xu}, {and} \bibinfo{person}{Yunhe Wang}.}
  \bibinfo{year}{2019}\natexlab{}.
\newblock \showarticletitle{Searching for Accurate Binary Neural
  Architectures}. In \bibinfo{booktitle}{\emph{Proc. IEEE Int. Conf. Comput.
  Vis.}} \bibinfo{pages}{2041--2044}.
\newblock


\bibitem[Siems et~al\mbox{.}(2020)]%
        {793r}
\bibfield{author}{\bibinfo{person}{Julien Siems}, \bibinfo{person}{Lucas
  Zimmer}, \bibinfo{person}{Arber Zela}, \bibinfo{person}{Jovita Lukasik},
  \bibinfo{person}{Margret Keuper}, {and} \bibinfo{person}{Frank Hutter}.}
  \bibinfo{year}{2020}\natexlab{}.
\newblock \showarticletitle{{NAS}-bench-301 and The Case for Surrogate
  Benchmarks for Neural Architecture Search}.
\newblock \bibinfo{journal}{\emph{arXiv preprint arXiv:2008.09777}}
  (\bibinfo{year}{2020}).
\newblock


\bibitem[Sikdar et~al\mbox{.}(2012)]%
        {779r}
\bibfield{author}{\bibinfo{person}{Utpal~Kumar Sikdar}, \bibinfo{person}{Asif
  Ekbal}, {and} \bibinfo{person}{Sriparna Saha}.}
  \bibinfo{year}{2012}\natexlab{}.
\newblock \showarticletitle{Differential Evolution Based Feature Selection and
  Classifier Ensemble for Named Entity Recognition}. In
  \bibinfo{booktitle}{\emph{Proc. COLING}}. \bibinfo{pages}{2475--2490}.
\newblock


\bibitem[Smith and Jin(2014)]%
        {709r}
\bibfield{author}{\bibinfo{person}{Christopher Smith} {and}
  \bibinfo{person}{Yaochu Jin}.} \bibinfo{year}{2014}\natexlab{}.
\newblock \showarticletitle{Evolutionary Multi-objective Generation of
  Recurrent Neural Network Ensembles for Time Series Prediction}.
\newblock \bibinfo{journal}{\emph{Neurocomputing}}  \bibinfo{volume}{143}
  (\bibinfo{year}{2014}), \bibinfo{pages}{302--311}.
\newblock
\showISSN{0925-2312}


\bibitem[Socha and Blum(2007)]%
        {491r}
\bibfield{author}{\bibinfo{person}{Krzysztof Socha} {and}
  \bibinfo{person}{Christian Blum}.} \bibinfo{year}{2007}\natexlab{}.
\newblock \showarticletitle{An Ant Colony Optimization Algorithm for Continuous
  Optimization: Application to Feed-forward Neural Network Training}.
\newblock \bibinfo{journal}{\emph{Neural. Comput. Appl.}} \bibinfo{volume}{16},
  \bibinfo{number}{3} (\bibinfo{year}{2007}), \bibinfo{pages}{235--247}.
\newblock
\showISSN{0941-0643}


\bibitem[Song et~al\mbox{.}(2020)]%
        {570r}
\bibfield{author}{\bibinfo{person}{Dehua Song}, \bibinfo{person}{Chang Xu},
  \bibinfo{person}{Xu Jia}, \bibinfo{person}{Yiyi Chen},
  \bibinfo{person}{Chunjing Xu}, {and} \bibinfo{person}{Yunhe Wang}.}
  \bibinfo{year}{2020}\natexlab{}.
\newblock \showarticletitle{Efficient Residual Dense Block Search for Image
  Super-Resolution}. In \bibinfo{booktitle}{\emph{Proc. AAAI Conf. Artif.
  Intell.}}, Vol.~\bibinfo{volume}{34}. \bibinfo{pages}{12007--12014}.
\newblock


\bibitem[Song(2021)]%
        {778r}
\bibfield{author}{\bibinfo{person}{Xin Song}.} \bibinfo{year}{2021}\natexlab{}.
\newblock \showarticletitle{Intelligent English Translation System Based on
  Evolutionary Multi-Objective Optimization Algorithm}.
\newblock \bibinfo{journal}{\emph{J. Intell. Fuzzy Syst.}}
  \bibinfo{volume}{40} (\bibinfo{year}{2021}), \bibinfo{pages}{6327--6337}.
\newblock
\showISSN{1064-1246}


\bibitem[Stanley and Miikkulainen(2002)]%
        {035r}
\bibfield{author}{\bibinfo{person}{Kenneth~O Stanley} {and}
  \bibinfo{person}{Risto Miikkulainen}.} \bibinfo{year}{2002}\natexlab{}.
\newblock \showarticletitle{Evolving Neural Networks Through Augmenting
  Topologies}.
\newblock \bibinfo{journal}{\emph{Evol. Comput.}} \bibinfo{volume}{10},
  \bibinfo{number}{2} (\bibinfo{year}{2002}), \bibinfo{pages}{99--127}.
\newblock
\showISSN{1063-6560}


\bibitem[Sun et~al\mbox{.}(2021)]%
        {678r}
\bibfield{author}{\bibinfo{person}{Yanan Sun}, \bibinfo{person}{Xian Sun},
  \bibinfo{person}{Yuhan Fang}, \bibinfo{person}{Gary~G. Yen}, {and}
  \bibinfo{person}{Yuqiao Liu}.} \bibinfo{year}{2021}\natexlab{}.
\newblock \showarticletitle{A Novel Training Protocol for Performance
  Predictors of Evolutionary Neural Architecture Search Algorithms}.
\newblock \bibinfo{journal}{\emph{IEEE Trans. Evol. Comput.}}
  \bibinfo{volume}{25}, \bibinfo{number}{3} (\bibinfo{year}{2021}),
  \bibinfo{pages}{524--536}.
\newblock


\bibitem[Sun et~al\mbox{.}(2019a)]%
        {123r}
\bibfield{author}{\bibinfo{person}{Yanan Sun}, \bibinfo{person}{Handing Wang},
  \bibinfo{person}{Bing Xue}, \bibinfo{person}{Yaochu Jin},
  \bibinfo{person}{Gary~G Yen}, {and} \bibinfo{person}{Mengjie Zhang}.}
  \bibinfo{year}{2019}\natexlab{a}.
\newblock \showarticletitle{Surrogate-Assisted Evolutionary Deep Learning Using
  an End-To-End Random Forest-Based Performance Predictor}.
\newblock \bibinfo{journal}{\emph{IEEE Trans. Evol. Comput.}}
  \bibinfo{volume}{24}, \bibinfo{number}{2} (\bibinfo{year}{2019}),
  \bibinfo{pages}{350--364}.
\newblock
\showISSN{1089-778X}


\bibitem[Sun et~al\mbox{.}(2019b)]%
        {104r}
\bibfield{author}{\bibinfo{person}{Yanan Sun}, \bibinfo{person}{Bing Xue},
  \bibinfo{person}{Mengjie Zhang}, {and} \bibinfo{person}{Gary~G Yen}.}
  \bibinfo{year}{2019}\natexlab{b}.
\newblock \showarticletitle{Completely Automated {CNN} Architecture Design
  Based on Blocks}.
\newblock \bibinfo{journal}{\emph{IEEE Trans. Neural Netw. Learn. Syst.}}
  \bibinfo{volume}{31}, \bibinfo{number}{4} (\bibinfo{year}{2019}),
  \bibinfo{pages}{1242--1254}.
\newblock


\bibitem[Sun et~al\mbox{.}(2019c)]%
        {029r}
\bibfield{author}{\bibinfo{person}{Yanan Sun}, \bibinfo{person}{Bing Xue},
  \bibinfo{person}{Mengjie Zhang}, {and} \bibinfo{person}{Gary~G Yen}.}
  \bibinfo{year}{2019}\natexlab{c}.
\newblock \showarticletitle{Evolving Deep Convolutional Neural Networks for
  Image Classification}.
\newblock \bibinfo{journal}{\emph{IEEE Trans. Evol. Comput.}}
  \bibinfo{volume}{24}, \bibinfo{number}{2} (\bibinfo{year}{2019}),
  \bibinfo{pages}{394--407}.
\newblock
\showISSN{1089-778X}


\bibitem[Sun et~al\mbox{.}(2019d)]%
        {663r}
\bibfield{author}{\bibinfo{person}{Yanan Sun}, \bibinfo{person}{Bing Xue},
  \bibinfo{person}{Mengjie Zhang}, {and} \bibinfo{person}{Gary~G. Yen}.}
  \bibinfo{year}{2019}\natexlab{d}.
\newblock \showarticletitle{A Particle Swarm Optimization-Based Flexible
  Convolutional Autoencoder for Image Classification}.
\newblock \bibinfo{journal}{\emph{IEEE Trans. Neural Netw. Learn. Syst.}}
  \bibinfo{volume}{30}, \bibinfo{number}{8} (\bibinfo{year}{2019}),
  \bibinfo{pages}{2295--2309}.
\newblock


\bibitem[Sun et~al\mbox{.}(2020)]%
        {036r}
\bibfield{author}{\bibinfo{person}{Yanan Sun}, \bibinfo{person}{Bing Xue},
  \bibinfo{person}{Mengjie Zhang}, \bibinfo{person}{Gary~G Yen}, {and}
  \bibinfo{person}{Jiancheng Lv}.} \bibinfo{year}{2020}\natexlab{}.
\newblock \showarticletitle{Automatically Designing {CNN} Architectures Using
  the Genetic Algorithm for Image Classification}.
\newblock \bibinfo{journal}{\emph{IEEE Trans. Cybern.}} \bibinfo{volume}{50},
  \bibinfo{number}{9} (\bibinfo{year}{2020}), \bibinfo{pages}{3840--3854}.
\newblock
\showISSN{2168-2267}


\bibitem[Swaminathan et~al\mbox{.}(2020)]%
        {844r}
\bibfield{author}{\bibinfo{person}{Sridhar Swaminathan},
  \bibinfo{person}{Deepak Garg}, \bibinfo{person}{Rajkumar Kannan}, {and}
  \bibinfo{person}{Fr{\'e}d{\'e}ric Andr{\`e}s}.}
  \bibinfo{year}{2020}\natexlab{}.
\newblock \showarticletitle{Sparse Low Rank Factorization for Deep Neural
  Network compression}.
\newblock \bibinfo{journal}{\emph{Neurocomputing}}  \bibinfo{volume}{398}
  (\bibinfo{year}{2020}), \bibinfo{pages}{185--196}.
\newblock
\showISSN{0925-2312}


\bibitem[Tanaka et~al\mbox{.}(2016)]%
        {890r}
\bibfield{author}{\bibinfo{person}{Tomohiro Tanaka}, \bibinfo{person}{Takafumi
  Moriya}, {and} \bibinfo{person}{Takahiro Shinozaki}.}
  \bibinfo{year}{2016}\natexlab{}.
\newblock \showarticletitle{Evolutionary Optimization of Long Short-Term Memory
  Neural Network Language Model}.
\newblock \bibinfo{journal}{\emph{J. Acoust. Soc. Am.}} \bibinfo{volume}{140},
  \bibinfo{number}{4} (\bibinfo{year}{2016}), \bibinfo{pages}{3062--3062}.
\newblock
\showISSN{0001-4966}


\bibitem[Tang et~al\mbox{.}(2019)]%
        {769r}
\bibfield{author}{\bibinfo{person}{Yajiao Tang}, \bibinfo{person}{Junkai Ji},
  \bibinfo{person}{Yulin Zhu}, \bibinfo{person}{Shangce Gao},
  \bibinfo{person}{Zheng Tang}, {and} \bibinfo{person}{Yuki Todo}.}
  \bibinfo{year}{2019}\natexlab{}.
\newblock \showarticletitle{A Differential Evolution-Oriented Pruning Neural
  Network Model for Bankruptcy Prediction}. In
  \bibinfo{booktitle}{\emph{Complexity}}, Vol.~\bibinfo{volume}{2019}.
  \bibinfo{pages}{8682124:1--8682124:21}.
\newblock
\showISSN{1076-2787}


\bibitem[Tariq et~al\mbox{.}(2018)]%
        {384r}
\bibfield{author}{\bibinfo{person}{Hassan Tariq}, \bibinfo{person}{Elf
  Eldridge}, {and} \bibinfo{person}{Ian Welch}.}
  \bibinfo{year}{2018}\natexlab{}.
\newblock \showarticletitle{An Efficient Approach for Feature Construction of
  High-Dimensional Microarray Data By Random Projections}.
\newblock \bibinfo{journal}{\emph{PLoS ONE}} \bibinfo{volume}{13},
  \bibinfo{number}{4} (\bibinfo{year}{2018}), \bibinfo{pages}{e0196385}.
\newblock
\showISSN{1932-6203}


\bibitem[Telikani et~al\mbox{.}(2021)]%
        {041r}
\bibfield{author}{\bibinfo{person}{Akbar Telikani}, \bibinfo{person}{Amirhessam
  Tahmassebi}, \bibinfo{person}{Wolfgang Banzhaf}, {and}
  \bibinfo{person}{Amir~H Gandomi}.} \bibinfo{year}{2021}\natexlab{}.
\newblock \showarticletitle{Evolutionary Machine Learning: A Survey}.
\newblock \bibinfo{journal}{\emph{ACM Comput. Surv.}} \bibinfo{volume}{54},
  \bibinfo{number}{8} (\bibinfo{year}{2021}), \bibinfo{pages}{1--35}.
\newblock
\showISSN{0360-0300}


\bibitem[Teller and Veloso(1996)]%
        {368r}
\bibfield{author}{\bibinfo{person}{Astro Teller} {and} \bibinfo{person}{Manuela
  Veloso}.} \bibinfo{year}{1996}\natexlab{}.
\newblock \showarticletitle{{PADO}: A New Learning Architecture for Object
  Recognition}.
\newblock \bibinfo{journal}{\emph{Symbolic visual learn.}}
  (\bibinfo{year}{1996}), \bibinfo{pages}{81--116}.
\newblock


\bibitem[Tian et~al\mbox{.}(2019)]%
        {549r}
\bibfield{author}{\bibinfo{person}{Haiman Tian}, \bibinfo{person}{ShuChing
  Chen}, \bibinfo{person}{MeiLing Shyu}, {and} \bibinfo{person}{Stuart~Harvey
  Rubin}.} \bibinfo{year}{2019}\natexlab{}.
\newblock \showarticletitle{Automated Neural Network Construction with
  Similarity Sensitive Evolutionary Algorithms}. In
  \bibinfo{booktitle}{\emph{Proc. IEEE Int. Conf. Inf. Reuse Integr. Data
  Sci.}} \bibinfo{pages}{283--290}.
\newblock


\bibitem[Tran et~al\mbox{.}(2018)]%
        {z21}
\bibfield{author}{\bibinfo{person}{Binh Tran}, \bibinfo{person}{Bing Xue},
  {and} \bibinfo{person}{Mengjie Zhang}.} \bibinfo{year}{2018}\natexlab{}.
\newblock \showarticletitle{A New Representation in PSO for
  Discretization-Based Feature Selection}.
\newblock \bibinfo{journal}{\emph{IEEE Trans. Cybern.}} \bibinfo{volume}{48},
  \bibinfo{number}{6} (\bibinfo{year}{2018}), \bibinfo{pages}{1733--1746}.
\newblock


\bibitem[Tran et~al\mbox{.}(2019)]%
        {279r}
\bibfield{author}{\bibinfo{person}{Binh Tran}, \bibinfo{person}{Bing Xue},
  {and} \bibinfo{person}{Mengjie Zhang}.} \bibinfo{year}{2019}\natexlab{}.
\newblock \showarticletitle{Genetic Programming for Multiple-Feature
  Construction on High-Dimensional Classification}.
\newblock \bibinfo{journal}{\emph{Pattern Recognit.}}  \bibinfo{volume}{93}
  (\bibinfo{year}{2019}), \bibinfo{pages}{404--417}.
\newblock
\showISSN{0031-3203}


\bibitem[Tran et~al\mbox{.}(2016)]%
        {365r}
\bibfield{author}{\bibinfo{person}{Binh Tran}, \bibinfo{person}{Mengjie Zhang},
  {and} \bibinfo{person}{Bing Xue}.} \bibinfo{year}{2016}\natexlab{}.
\newblock \showarticletitle{Multiple Feature Construction in Classification on
  High-Dimensional Data Using GP}. In \bibinfo{booktitle}{\emph{IEEE Symposium
  Series on Computational Intelligence}}. \bibinfo{pages}{1--8}.
\newblock


\bibitem[Vafaie and De~Jong(1998)]%
        {369r}
\bibfield{author}{\bibinfo{person}{Haleh Vafaie} {and} \bibinfo{person}{Kenneth
  De~Jong}.} \bibinfo{year}{1998}\natexlab{}.
\newblock \showarticletitle{Feature Space Transformation Using Genetic
  Algorithms}.
\newblock \bibinfo{journal}{\emph{IEEE Intell. Syst. Appli.}}
  \bibinfo{volume}{13}, \bibinfo{number}{2} (\bibinfo{year}{1998}),
  \bibinfo{pages}{57--65}.
\newblock


\bibitem[Vargas-H{\'a}kim et~al\mbox{.}(2022)]%
        {061r}
\bibfield{author}{\bibinfo{person}{Gustavo~A Vargas-H{\'a}kim},
  \bibinfo{person}{Efr{\'e}n Mezura-Montes}, {and}
  \bibinfo{person}{H{\'e}ctor-Gabriel Acosta-Mesa}.}
  \bibinfo{year}{2022}\natexlab{}.
\newblock \showarticletitle{A Review on Convolutional Neural Networks Encodings
  for Neuroevolution}.
\newblock \bibinfo{journal}{\emph{IEEE Trans. Evol. Comput.}}
  \bibinfo{volume}{26}, \bibinfo{number}{1} (\bibinfo{year}{2022}),
  \bibinfo{pages}{12--27}.
\newblock


\bibitem[Vieira et~al\mbox{.}(2013)]%
        {170r}
\bibfield{author}{\bibinfo{person}{Susana~M Vieira},
  \bibinfo{person}{Lu{\'\i}s~F Mendon{\c{c}}a}, \bibinfo{person}{Goncalo~J
  Farinha}, {and} \bibinfo{person}{Jo{\~a}o~MC Sousa}.}
  \bibinfo{year}{2013}\natexlab{}.
\newblock \showarticletitle{Modified Binary {PSO} for Feature Selection Using
  {SVM} Applied to Mortality Prediction of Septic Patients}.
\newblock \bibinfo{journal}{\emph{Appl. Soft Comput.}} \bibinfo{volume}{13},
  \bibinfo{number}{8} (\bibinfo{year}{2013}), \bibinfo{pages}{3494--3504}.
\newblock
\showISSN{1568-4946}


\bibitem[Wang et~al\mbox{.}(2021b)]%
        {z18}
\bibfield{author}{\bibinfo{person}{Bin Wang}, \bibinfo{person}{Wenbin Pei},
  \bibinfo{person}{Bing Xue}, {and} \bibinfo{person}{Mengjie Zhang}.}
  \bibinfo{year}{2021}\natexlab{b}.
\newblock \showarticletitle{Evolving Local Interpretable Model-Agnostic
  Explanations for Deep Neural Networks in Image Classification}. In
  \bibinfo{booktitle}{\emph{Proc. Genetic Evol. Comput. Conf.}}
  \bibinfo{pages}{173--174}.
\newblock


\bibitem[Wang et~al\mbox{.}(2020b)]%
        {z10}
\bibfield{author}{\bibinfo{person}{Bin Wang}, \bibinfo{person}{Bing Xue}, {and}
  \bibinfo{person}{Mengjie Zhang}.} \bibinfo{year}{2020}\natexlab{b}.
\newblock \showarticletitle{Particle Swarm Optimization for Evolving Deep
  Convolutional Neural Networks for Image Classification: Single-and
  Multi-objective Approaches}.
\newblock In \bibinfo{booktitle}{\emph{Deep Neural Evolution}}.
  \bibinfo{publisher}{Springer}, \bibinfo{pages}{155--184}.
\newblock


\bibitem[Wang et~al\mbox{.}(2020c)]%
        {117r}
\bibfield{author}{\bibinfo{person}{Bin Wang}, \bibinfo{person}{Bing Xue}, {and}
  \bibinfo{person}{Mengjie Zhang}.} \bibinfo{year}{2020}\natexlab{c}.
\newblock \showarticletitle{Particle Swarm Optimization for Evolving Deep
  Neural Networks for Image Classification By Evolving and Stacking
  Transferable Blocks}. In \bibinfo{booktitle}{\emph{Proc. IEEE Congr. Evol.
  Comput.}} \bibinfo{pages}{1--8}.
\newblock


\bibitem[Wang et~al\mbox{.}(2021c)]%
        {126r}
\bibfield{author}{\bibinfo{person}{Bin Wang}, \bibinfo{person}{Bing Xue}, {and}
  \bibinfo{person}{Mengjie Zhang}.} \bibinfo{year}{2021}\natexlab{c}.
\newblock \showarticletitle{Surrogate-Assisted Particle Swarm Optimization for
  Evolving Variable-Length Transferable Blocks for Image Classification}.
\newblock \bibinfo{journal}{\emph{IEEE Trans. Neural Netw. Learn. Syst.}}
  \bibinfo{volume}{33}, \bibinfo{number}{8} (\bibinfo{year}{2021}),
  \bibinfo{pages}{3727--3740}.
\newblock


\bibitem[Wang et~al\mbox{.}(2020a)]%
        {798r}
\bibfield{author}{\bibinfo{person}{Hanrui Wang}, \bibinfo{person}{Zhanghao Wu},
  \bibinfo{person}{Zhijian Liu}, \bibinfo{person}{Han Cai},
  \bibinfo{person}{Ligeng Zhu}, \bibinfo{person}{Chuang Gan}, {and}
  \bibinfo{person}{Song Han}.} \bibinfo{year}{2020}\natexlab{a}.
\newblock \showarticletitle{{HAT}: Hardware-Aware Transformers for Efficient
  Natural Language Processing}. In \bibinfo{booktitle}{\emph{Proc. Assoc.
  Comput. Linguist.}} \bibinfo{pages}{7675--7688}.
\newblock


\bibitem[Wang et~al\mbox{.}(2020d)]%
        {181r}
\bibfield{author}{\bibinfo{person}{Xiao-han Wang}, \bibinfo{person}{Yong
  Zhang}, {and} \bibinfo{person}{Xiao-yan Sun}.}
  \bibinfo{year}{2020}\natexlab{d}.
\newblock \showarticletitle{Multi-Objective Feature Selection Based on
  Artificial Bee Colony: An Acceleration Approach With Variable Sample Size}.
\newblock \bibinfo{journal}{\emph{Appl. Soft Comput.}}  \bibinfo{volume}{88}
  (\bibinfo{year}{2020}), \bibinfo{pages}{106041}.
\newblock
\showISSN{1568-4946}


\bibitem[Wang et~al\mbox{.}(2018)]%
        {811r}
\bibfield{author}{\bibinfo{person}{Yunhe Wang}, \bibinfo{person}{Chang Xu},
  \bibinfo{person}{Jiayan Qiu}, \bibinfo{person}{Chao Xu}, {and}
  \bibinfo{person}{Dacheng Tao}.} \bibinfo{year}{2018}\natexlab{}.
\newblock \showarticletitle{Towards Evolutionary Compression}. In
  \bibinfo{booktitle}{\emph{Proc. ACM SIGKDD Int. Conf. Knowl. Discov. \& Data
  Min.}} \bibinfo{pages}{2476--2485}.
\newblock


\bibitem[Wang et~al\mbox{.}(2021a)]%
        {068r}
\bibfield{author}{\bibinfo{person}{Zhehui Wang}, \bibinfo{person}{Tao Luo},
  \bibinfo{person}{Miqing Li}, \bibinfo{person}{Joey~Tianyi Zhou},
  \bibinfo{person}{Rick Siow~Mong Goh}, {and} \bibinfo{person}{Liangli Zhen}.}
  \bibinfo{year}{2021}\natexlab{a}.
\newblock \showarticletitle{Evolutionary Multi-Objective Model Compression for
  Deep Neural Networks}.
\newblock \bibinfo{journal}{\emph{IEEE Comput. Intell. Mag.}}
  \bibinfo{volume}{16}, \bibinfo{number}{3} (\bibinfo{year}{2021}),
  \bibinfo{pages}{10--21}.
\newblock
\showISSN{1556-603X}


\bibitem[Wen and Xu(2011)]%
        {195r}
\bibfield{author}{\bibinfo{person}{Yun Wen} {and} \bibinfo{person}{Hua Xu}.}
  \bibinfo{year}{2011}\natexlab{}.
\newblock \showarticletitle{A Cooperative Coevolution-Based Pittsburgh Learning
  Classifier System Embedded With Memetic Feature Selection}. In
  \bibinfo{booktitle}{\emph{Proc. IEEE Congr. Evol. Comput.}}
  \bibinfo{pages}{2415--2422}.
\newblock


\bibitem[White et~al\mbox{.}(2021)]%
        {692r}
\bibfield{author}{\bibinfo{person}{Colin White}, \bibinfo{person}{Willie
  Neiswanger}, {and} \bibinfo{person}{Yash Savani}.}
  \bibinfo{year}{2021}\natexlab{}.
\newblock \showarticletitle{{BANANAS}: Bayesian Optimization with Neural
  Architectures for Neural Architecture Search}. In
  \bibinfo{booktitle}{\emph{Proc. AAAI Conf. Artif. Intell.}},
  Vol.~\bibinfo{volume}{35}. \bibinfo{pages}{10293--10301}.
\newblock


\bibitem[Winata et~al\mbox{.}(2019)]%
        {845r}
\bibfield{author}{\bibinfo{person}{Genta~Indra Winata}, \bibinfo{person}{Andrea
  Madotto}, \bibinfo{person}{Jamin Shin}, \bibinfo{person}{Elham~J Barezi},
  {and} \bibinfo{person}{Pascale Fung}.} \bibinfo{year}{2019}\natexlab{}.
\newblock \showarticletitle{On the Effectiveness of Low-rank Matrix
  Factorization for {LSTM} Model Compression}.
\newblock \bibinfo{journal}{\emph{arXiv preprint arXiv:1908.09982}}
  (\bibinfo{year}{2019}).
\newblock


\bibitem[Wu et~al\mbox{.}(2021b)]%
        {528r}
\bibfield{author}{\bibinfo{person}{Min Wu}, \bibinfo{person}{Wanjuan Su},
  \bibinfo{person}{Luefeng Chen}, {and} \bibinfo{person}{Zhentao Liu}.}
  \bibinfo{year}{2021}\natexlab{b}.
\newblock \showarticletitle{Weight-Adapted Convolution Neural Network for
  Facial Expression Recognition in Human-Robot Interaction}.
\newblock \bibinfo{journal}{\emph{IEEE Trans. Syst. Man Cybern.}}
  \bibinfo{volume}{51}, \bibinfo{number}{3} (\bibinfo{year}{2021}),
  \bibinfo{pages}{1473--1484}.
\newblock
\showISSN{2168-2216}


\bibitem[Wu et~al\mbox{.}(2021a)]%
        {814r}
\bibfield{author}{\bibinfo{person}{Tao Wu}, \bibinfo{person}{Xiaoyang Li},
  \bibinfo{person}{Deyun Zhou}, \bibinfo{person}{Na Li}, {and}
  \bibinfo{person}{Jiao Shi}.} \bibinfo{year}{2021}\natexlab{a}.
\newblock \showarticletitle{Differential Evolution Based Layer-Wise Weight
  Pruning for Compressing Deep Neural Networks}.
\newblock \bibinfo{journal}{\emph{Sens.}} \bibinfo{volume}{21},
  \bibinfo{number}{3} (\bibinfo{year}{2021}), \bibinfo{pages}{880}.
\newblock
\showISSN{1424-8220}


\bibitem[Wu et~al\mbox{.}(2019)]%
        {829r}
\bibfield{author}{\bibinfo{person}{Tao Wu}, \bibinfo{person}{Jiao Shi},
  \bibinfo{person}{Deyun Zhou}, \bibinfo{person}{Yu Lei}, {and}
  \bibinfo{person}{Maoguo Gong}.} \bibinfo{year}{2019}\natexlab{}.
\newblock \showarticletitle{A Multi-objective Particle Swarm Optimization for
  Neural Networks Pruning}. In \bibinfo{booktitle}{\emph{Proc. IEEE Congr.
  Evol. Comput.}} \bibinfo{pages}{570--577}.
\newblock


\bibitem[Wu et~al\mbox{.}(2020)]%
        {799r}
\bibfield{author}{\bibinfo{person}{Xiang Wu}, \bibinfo{person}{Ran He},
  \bibinfo{person}{Yibo Hu}, {and} \bibinfo{person}{Zhenan Sun}.}
  \bibinfo{year}{2020}\natexlab{}.
\newblock \showarticletitle{Learning an Evolutionary Embedding via Massive
  Knowledge Distillation}.
\newblock \bibinfo{journal}{\emph{Int. J. Comput. Vis.}} \bibinfo{volume}{128},
  \bibinfo{number}{8} (\bibinfo{year}{2020}), \bibinfo{pages}{1--18}.
\newblock
\showISSN{0920-5691}


\bibitem[Xie et~al\mbox{.}(2022a)]%
        {781r}
\bibfield{author}{\bibinfo{person}{Lingxi Xie}, \bibinfo{person}{Xin Chen},
  \bibinfo{person}{Kaifeng Bi}, \bibinfo{person}{Longhui Wei},
  \bibinfo{person}{Yuhui Xu}, \bibinfo{person}{Zhengsu Chen},
  \bibinfo{person}{Lanfei Wang}, \bibinfo{person}{Anxiang Xiao},
  \bibinfo{person}{Jianlong Chang}, \bibinfo{person}{Xiaopeng Zhang}, {and}
  \bibinfo{person}{Qi Tian}.} \bibinfo{year}{2022}\natexlab{a}.
\newblock \showarticletitle{Weight-Sharing Neural Architecture Search: A Battle
  to Shrink the Optimization Gap}.
\newblock \bibinfo{journal}{\emph{ACM Comput. Surv.}} \bibinfo{volume}{54},
  \bibinfo{number}{9} (\bibinfo{year}{2022}), \bibinfo{pages}{1--37}.
\newblock
\showISSN{0360-0300}


\bibitem[Xie and Yuille(2017)]%
        {137r}
\bibfield{author}{\bibinfo{person}{Lingxi Xie} {and} \bibinfo{person}{Alan
  Yuille}.} \bibinfo{year}{2017}\natexlab{}.
\newblock \showarticletitle{Genetic {CNN}}. In \bibinfo{booktitle}{\emph{Proc.
  IEEE Int. Conf. Comput. Vis.}} \bibinfo{pages}{1379--1388}.
\newblock


\bibitem[Xie et~al\mbox{.}(2022b)]%
        {z17}
\bibfield{author}{\bibinfo{person}{Xiangning Xie}, \bibinfo{person}{Yuqiao
  Liu}, \bibinfo{person}{Yanan Sun}, \bibinfo{person}{Gary~G. Yen},
  \bibinfo{person}{Bing Xue}, {and} \bibinfo{person}{Mengjie Zhang}.}
  \bibinfo{year}{2022}\natexlab{b}.
\newblock \showarticletitle{Bench{ENAS}: A Benchmarking Platform for
  Evolutionary Neural Architecture Search}.
\newblock \bibinfo{journal}{\emph{IEEE Trans. Evol. Comput.}}
  (\bibinfo{year}{2022}).
\newblock
\urldef\tempurl%
\url{https://doi.org/10.1109/TEVC.2022.3147526}
\showDOI{\tempurl}


\bibitem[Xu et~al\mbox{.}(2021)]%
        {838r}
\bibfield{author}{\bibinfo{person}{Ke Xu}, \bibinfo{person}{Dezheng Zhang},
  \bibinfo{person}{Jianjing An}, \bibinfo{person}{Li Liu},
  \bibinfo{person}{Lingzhi Liu}, {and} \bibinfo{person}{Dong Wang}.}
  \bibinfo{year}{2021}\natexlab{}.
\newblock \showarticletitle{Gen{E}xp: Multi-objective Pruning for Deep Neural
  Network based on Genetic Algorithm}.
\newblock \bibinfo{journal}{\emph{Neurocomputing}}  \bibinfo{volume}{451}
  (\bibinfo{year}{2021}), \bibinfo{pages}{81--94}.
\newblock
\showISSN{0925-2312}


\bibitem[Xue et~al\mbox{.}(2012)]%
        {091r}
\bibfield{author}{\bibinfo{person}{Bing Xue}, \bibinfo{person}{Mengjie Zhang},
  {and} \bibinfo{person}{Will~N Browne}.} \bibinfo{year}{2012}\natexlab{}.
\newblock \showarticletitle{Multi-Objective Particle Swarm Optimization ({PSO})
  for Feature Selection}. In \bibinfo{booktitle}{\emph{Proc. Genetic Evol.
  Comput. Conf.}} \bibinfo{pages}{81--88}.
\newblock


\bibitem[Xue et~al\mbox{.}(2015)]%
        {033r}
\bibfield{author}{\bibinfo{person}{Bing Xue}, \bibinfo{person}{Mengjie Zhang},
  \bibinfo{person}{Will~N Browne}, {and} \bibinfo{person}{Xin Yao}.}
  \bibinfo{year}{2015}\natexlab{}.
\newblock \showarticletitle{A Survey on Evolutionary Computation Approaches to
  Feature Selection}.
\newblock \bibinfo{journal}{\emph{IEEE Trans. Evol. Comput.}}
  \bibinfo{volume}{20}, \bibinfo{number}{4} (\bibinfo{year}{2015}),
  \bibinfo{pages}{606--626}.
\newblock
\showISSN{1089-778X}


\bibitem[Xue et~al\mbox{.}(2013)]%
        {064r}
\bibfield{author}{\bibinfo{person}{Bing Xue}, \bibinfo{person}{Mengjie Zhang},
  \bibinfo{person}{Yan Dai}, {and} \bibinfo{person}{Will~N Browne}.}
  \bibinfo{year}{2013}\natexlab{}.
\newblock \showarticletitle{{PSO} for Feature Construction and Binary
  Classification}. In \bibinfo{booktitle}{\emph{Proc. Genetic Evol. Comput.
  Conf.}} \bibinfo{pages}{137--144}.
\newblock


\bibitem[Yang et~al\mbox{.}(2019)]%
        {839r}
\bibfield{author}{\bibinfo{person}{Chuanguang Yang}, \bibinfo{person}{Zhulin
  An}, \bibinfo{person}{Chao Li}, \bibinfo{person}{Boyu Diao}, {and}
  \bibinfo{person}{Yongjun Xu}.} \bibinfo{year}{2019}\natexlab{}.
\newblock \showarticletitle{Multi-objective Pruning for {CNN}s Using Genetic
  Algorithm}. In \bibinfo{booktitle}{\emph{Proc. Int. Conf. Artif. Neural
  Netw.}} \bibinfo{pages}{299--305}.
\newblock


\bibitem[Yang et~al\mbox{.}(2021)]%
        {513r}
\bibfield{author}{\bibinfo{person}{Shangshang Yang}, \bibinfo{person}{Ye Tian},
  \bibinfo{person}{Cheng He}, \bibinfo{person}{Xingyi Zhang},
  \bibinfo{person}{Kay~Chen Tan}, {and} \bibinfo{person}{Yaochu Jin}.}
  \bibinfo{year}{2021}\natexlab{}.
\newblock \showarticletitle{A Gradient-Guided Evolutionary Approach to Training
  Deep Neural Networks}.
\newblock \bibinfo{journal}{\emph{IEEE Trans. Neural Netw. Learn. Syst.}}
  (\bibinfo{year}{2021}).
\newblock
\urldef\tempurl%
\url{https://doi.org/10.1109/TNNLS.2021.3061630}
\showDOI{\tempurl}


\bibitem[Yang et~al\mbox{.}(2020)]%
        {610r}
\bibfield{author}{\bibinfo{person}{Zhaohui Yang}, \bibinfo{person}{Yunhe Wang},
  \bibinfo{person}{Xinghao Chen}, \bibinfo{person}{Boxin Shi}, {and}
  \bibinfo{person}{Chao Xu}.} \bibinfo{year}{2020}\natexlab{}.
\newblock \showarticletitle{{CARS}: Continuous Evolution for Efficient Neural
  Architecture Search}. In \bibinfo{booktitle}{\emph{Proc. IEEE Conf. Comput.
  Vis. Pattern Recognit.}} \bibinfo{pages}{1826--1835}.
\newblock


\bibitem[Yao et~al\mbox{.}(2018)]%
        {024r}
\bibfield{author}{\bibinfo{person}{Quanming Yao}, \bibinfo{person}{Mengshuo
  Wang}, \bibinfo{person}{Yuqiang Chen}, \bibinfo{person}{Wenyuan Dai}, {and}
  \bibinfo{person}{Yu-Feng Li}.} \bibinfo{year}{2018}\natexlab{}.
\newblock \showarticletitle{Taking Human out of Learning Applications: A Survey
  on Automated Machine Learning}.
\newblock \bibinfo{journal}{\emph{arXiv preprint arXiv:1810.13306}}
  (\bibinfo{year}{2018}).
\newblock


\bibitem[Yao(1993)]%
        {049r}
\bibfield{author}{\bibinfo{person}{Xin Yao}.} \bibinfo{year}{1993}\natexlab{}.
\newblock \showarticletitle{A Review of Evolutionary Artificial Neural
  Networks}.
\newblock \bibinfo{journal}{\emph{Int. J. Intell. Syst.}} \bibinfo{volume}{8},
  \bibinfo{number}{4} (\bibinfo{year}{1993}), \bibinfo{pages}{539--567}.
\newblock
\showISSN{0884-8173}


\bibitem[Yao and Liu(1996)]%
        {717r}
\bibfield{author}{\bibinfo{person}{Xin Yao} {and} \bibinfo{person}{Yong Liu}.}
  \bibinfo{year}{1996}\natexlab{}.
\newblock \showarticletitle{Ensemble Structure of Evolutionary Artificial
  Neural Networks}. In \bibinfo{booktitle}{\emph{Proc. Genetic Evol. Comput.
  Conf.}} \bibinfo{pages}{659--664}.
\newblock
\showISSN{1083-4419}


\bibitem[Ying et~al\mbox{.}(2019)]%
        {Ying2019NASBench101TR}
\bibfield{author}{\bibinfo{person}{Chris Ying}, \bibinfo{person}{Aaron Klein},
  \bibinfo{person}{Esteban Real}, \bibinfo{person}{Eric Christiansen},
  \bibinfo{person}{Kevin~P. Murphy}, {and} \bibinfo{person}{Frank Hutter}.}
  \bibinfo{year}{2019}\natexlab{}.
\newblock \showarticletitle{{NAS}-Bench-101: Towards Reproducible Neural
  Architecture Search}. In \bibinfo{booktitle}{\emph{Proc. Int. Conf. Learn.
  Represent.}} \bibinfo{pages}{https://arxiv.org/abs/1902.09635}.
\newblock


\bibitem[Young et~al\mbox{.}(2018)]%
        {775r}
\bibfield{author}{\bibinfo{person}{Tom Young}, \bibinfo{person}{Devamanyu
  Hazarika}, \bibinfo{person}{Soujanya Poria}, {and} \bibinfo{person}{Erik
  Cambria}.} \bibinfo{year}{2018}\natexlab{}.
\newblock \showarticletitle{Recent Trends in Deep Learning Based Natural
  Language Processing [Review Article]}.
\newblock \bibinfo{journal}{\emph{IEEE Comput. Intell. Mag.}}
  \bibinfo{volume}{13}, \bibinfo{number}{3} (\bibinfo{year}{2018}),
  \bibinfo{pages}{55--75}.
\newblock


\bibitem[Yu et~al\mbox{.}(2009)]%
        {193r}
\bibfield{author}{\bibinfo{person}{Hualong Yu}, \bibinfo{person}{Guochang Gu},
  \bibinfo{person}{Haibo Liu}, \bibinfo{person}{Jing Shen}, {and}
  \bibinfo{person}{Jing Zhao}.} \bibinfo{year}{2009}\natexlab{}.
\newblock \showarticletitle{A modified Ant Colony Optimization Algorithm for
  Tumor Marker Gene Selection}.
\newblock \bibinfo{journal}{\emph{Genomics, Proteomics \& Bioinformatics}}
  \bibinfo{volume}{7}, \bibinfo{number}{4} (\bibinfo{year}{2009}),
  \bibinfo{pages}{200--208}.
\newblock
\showISSN{1672-0229}


\bibitem[Z.-Flores et~al\mbox{.}(2020)]%
        {435r}
\bibfield{author}{\bibinfo{person}{Emigdio Z.-Flores},
  \bibinfo{person}{Leonardo Trujillo}, \bibinfo{person}{Pierrick Legrand},
  {and} \bibinfo{person}{Fr{\'e}d{\'e}rique Fa{\"i}ta-A{\"i}nseba}.}
  \bibinfo{year}{2020}\natexlab{}.
\newblock \showarticletitle{{EEG} Feature Extraction Using Genetic Programming
  for the Classification of Mental States}.
\newblock \bibinfo{journal}{\emph{Algorithms}} \bibinfo{volume}{13},
  \bibinfo{number}{9} (\bibinfo{year}{2020}), \bibinfo{pages}{221}.
\newblock
\showISSN{1999-4893}


\bibitem[Zemouri et~al\mbox{.}(2019)]%
        {771r}
\bibfield{author}{\bibinfo{person}{Ryad~A. Zemouri}, \bibinfo{person}{N. Omri},
  \bibinfo{person}{Farhat Fnaiech}, \bibinfo{person}{Noureddine Zerhouni},
  {and} \bibinfo{person}{Nader Fnaiech}.} \bibinfo{year}{2019}\natexlab{}.
\newblock \showarticletitle{A New Growing Pruning Deep Learning Neural Network
  Algorithm (GP-DLNN)}.
\newblock \bibinfo{journal}{\emph{Neural. Comput. Appl.}}  \bibinfo{volume}{32}
  (\bibinfo{year}{2019}), \bibinfo{pages}{1--17}.
\newblock
\showISSN{0941-0643}


\bibitem[Zhan et~al\mbox{.}(2021)]%
        {885r}
\bibfield{author}{\bibinfo{person}{Zheng Zhan}, \bibinfo{person}{Yifan Gong},
  \bibinfo{person}{Pu Zhao}, \bibinfo{person}{Geng Yuan}, \bibinfo{person}{Wei
  Niu}, \bibinfo{person}{Yushu Wu}, \bibinfo{person}{Tianyun Zhang},
  \bibinfo{person}{Malith Jayaweera}, \bibinfo{person}{David~R. Kaeli},
  \bibinfo{person}{Bin Ren}, \bibinfo{person}{Xue Lin}, {and}
  \bibinfo{person}{Yanzhi Wang}.} \bibinfo{year}{2021}\natexlab{}.
\newblock \showarticletitle{Achieving on-Mobile Real-Time Super-Resolution with
  Neural Architecture and Pruning Search}. In \bibinfo{booktitle}{\emph{Proc.
  IEEE Int. Conf. Comput. Vis.}} \bibinfo{pages}{4801--4811}.
\newblock


\bibitem[Zhang and M{\"u}hlenbein(1995)]%
        {727r}
\bibfield{author}{\bibinfo{person}{Byoung-Tak Zhang} {and}
  \bibinfo{person}{Heinz M{\"u}hlenbein}.} \bibinfo{year}{1995}\natexlab{}.
\newblock \showarticletitle{Balancing Accuracy and Parsimony in Genetic
  Programming}.
\newblock \bibinfo{journal}{\emph{Evol. Comput.}} \bibinfo{volume}{3},
  \bibinfo{number}{1} (\bibinfo{year}{1995}), \bibinfo{pages}{17--38}.
\newblock
\showISSN{1063-6560}


\bibitem[Zhang et~al\mbox{.}(2022b)]%
        {649r}
\bibfield{author}{\bibinfo{person}{Di Zhang}, \bibinfo{person}{Yichen Zhou},
  \bibinfo{person}{Jiaqi Zhao}, {and} \bibinfo{person}{Yong Zhou}.}
  \bibinfo{year}{2022}\natexlab{b}.
\newblock \showarticletitle{Co-evolution-based Parameter Learning for Remote
  Sensing Scene Classification}.
\newblock \bibinfo{journal}{\emph{Int. J. Wavelets Multiresolut. Inf.
  Process.}} \bibinfo{volume}{20}, \bibinfo{number}{2} (\bibinfo{year}{2022}),
  \bibinfo{pages}{2150046}.
\newblock


\bibitem[Zhang et~al\mbox{.}(2020b)]%
        {703r}
\bibfield{author}{\bibinfo{person}{Haoling Zhang},
  \bibinfo{person}{Chao-Han~Huck Yang}, \bibinfo{person}{Hector Zenil},
  \bibinfo{person}{Narsis~Aftab Kiani}, \bibinfo{person}{Yue Shen}, {and}
  \bibinfo{person}{Jesper~N. Tegner}.} \bibinfo{year}{2020}\natexlab{b}.
\newblock \showarticletitle{Evolving Neural Networks through a Reverse Encoding
  Tree}. In \bibinfo{booktitle}{\emph{Proc. IEEE Congr. Evol. Comput.}}
  \bibinfo{pages}{1--10}.
\newblock


\bibitem[Zhang and Gouza(2018)]%
        {531r}
\bibfield{author}{\bibinfo{person}{Jiawei Zhang} {and}
  \bibinfo{person}{Fisher~B Gouza}.} \bibinfo{year}{2018}\natexlab{}.
\newblock \showarticletitle{{GADAM}: Genetic-evolutionary {ADAM} for Deep
  Neural Network optimization}.
\newblock \bibinfo{journal}{\emph{arXiv preprint arXiv:1805.07500}}
  (\bibinfo{year}{2018}).
\newblock


\bibitem[Zhang et~al\mbox{.}(2011)]%
        {046r}
\bibfield{author}{\bibinfo{person}{Jun Zhang}, \bibinfo{person}{Zhi-hui Zhan},
  \bibinfo{person}{Ying Lin}, \bibinfo{person}{Ni Chen},
  \bibinfo{person}{Yue-jiao Gong}, \bibinfo{person}{Jing-hui Zhong},
  \bibinfo{person}{Henry~SH Chung}, \bibinfo{person}{Yun Li}, {and}
  \bibinfo{person}{Yu-hui Shi}.} \bibinfo{year}{2011}\natexlab{}.
\newblock \showarticletitle{Evolutionary Computation Meets Machine Learning: A
  Survey}.
\newblock \bibinfo{journal}{\emph{IEEE Comput. Intell. Mag.}}
  \bibinfo{volume}{6}, \bibinfo{number}{4} (\bibinfo{year}{2011}),
  \bibinfo{pages}{68--75}.
\newblock
\showISSN{1556-603X}


\bibitem[Zhang et~al\mbox{.}(2021a)]%
        {840r}
\bibfield{author}{\bibinfo{person}{Kaiyu Zhang}, \bibinfo{person}{Jinglong
  Chen}, \bibinfo{person}{Shuilong He}, \bibinfo{person}{Enyong Xu},
  \bibinfo{person}{Fudong Li}, {and} \bibinfo{person}{Zitong Zhou}.}
  \bibinfo{year}{2021}\natexlab{a}.
\newblock \showarticletitle{Differentiable Neural Architecture Search Augmented
  with Pruning and Multi-objective Optimization for Time-efficient Intelligent
  Fault Diagnosis of Machinery}.
\newblock \bibinfo{journal}{\emph{Mech. Syst. Signal Process.}}
  \bibinfo{volume}{158} (\bibinfo{year}{2021}), \bibinfo{pages}{107773}.
\newblock
\showISSN{0888-3270}


\bibitem[Zhang et~al\mbox{.}(2022a)]%
        {800r}
\bibfield{author}{\bibinfo{person}{Kangkai Zhang}, \bibinfo{person}{Chunhui
  Zhang}, \bibinfo{person}{Shikun Li}, \bibinfo{person}{Dan Zeng}, {and}
  \bibinfo{person}{Shiming Ge}.} \bibinfo{year}{2022}\natexlab{a}.
\newblock \showarticletitle{Student Network Learning via Evolutionary Knowledge
  Distillation}.
\newblock \bibinfo{journal}{\emph{IEEE Trans. Circuits. Syst. Video Technol.}}
  \bibinfo{volume}{32}, \bibinfo{number}{4} (\bibinfo{year}{2022}),
  \bibinfo{pages}{2251--2263}.
\newblock


\bibitem[Zhang(2018)]%
        {z03}
\bibfield{author}{\bibinfo{person}{Mengjie Zhang}.}
  \bibinfo{year}{2018}\natexlab{}.
\newblock \showarticletitle{Evolutionary Deep Learning for Image Analysis}.
\newblock  (\bibinfo{year}{2018}),
  \bibinfo{pages}{https://ieeetv.ieee.org/mengjie--zhang--evolutionary--deep--learning--for--image--analysis}.
\newblock


\bibitem[Zhang and Cagnoni(2020)]%
        {z02}
\bibfield{author}{\bibinfo{person}{Mengjie Zhang} {and}
  \bibinfo{person}{Stefano Cagnoni}.} \bibinfo{year}{2020}\natexlab{}.
\newblock \showarticletitle{Evolutionary Computation and Evolutionary Deep
  Learning for Image Analysis, Signal Processing and Pattern Recognition}. In
  \bibinfo{booktitle}{\emph{Proc. Genetic Evol. Comput. Conf.}}
  \bibinfo{pages}{1221--1257}.
\newblock


\bibitem[Zhang et~al\mbox{.}(2020a)]%
        {783r}
\bibfield{author}{\bibinfo{person}{Miao Zhang}, \bibinfo{person}{Huiqi Li},
  \bibinfo{person}{Shirui Pan}, \bibinfo{person}{Xiaojun Chang}, {and}
  \bibinfo{person}{Steven~W. Su}.} \bibinfo{year}{2020}\natexlab{a}.
\newblock \showarticletitle{Overcoming Multi-Model Forgetting in One-Shot NAS
  With Diversity Maximization}. In \bibinfo{booktitle}{\emph{Proc. IEEE Conf.
  Comput. Vis. Pattern Recognit.}} \bibinfo{pages}{7806--7815}.
\newblock


\bibitem[Zhang and Smart(2004)]%
        {z06}
\bibfield{author}{\bibinfo{person}{Mengjie Zhang} {and} \bibinfo{person}{Will
  Smart}.} \bibinfo{year}{2004}\natexlab{}.
\newblock \showarticletitle{Genetic Programming with Gradient Descent Search
  for Multiclass Object Classification}. In \bibinfo{booktitle}{\emph{Proc.
  Eur. Conf. Genetic Program}}. \bibinfo{pages}{399--408}.
\newblock


\bibitem[Zhang et~al\mbox{.}(2017)]%
        {209r}
\bibfield{author}{\bibinfo{person}{Yong Zhang}, \bibinfo{person}{Dun-wei Gong},
  \bibinfo{person}{Xiao-yan Sun}, {and} \bibinfo{person}{Yi-nan Guo}.}
  \bibinfo{year}{2017}\natexlab{}.
\newblock \showarticletitle{A {PSO}-Based Multi-Objective Multi-Label Feature
  Selection Method in Classification}.
\newblock \bibinfo{journal}{\emph{Sci. Rep.}} \bibinfo{volume}{7},
  \bibinfo{number}{1} (\bibinfo{year}{2017}), \bibinfo{pages}{1--12}.
\newblock
\showISSN{2045-2322}


\bibitem[Zhang and Rockett(2005)]%
        {454r}
\bibfield{author}{\bibinfo{person}{Yang Zhang} {and} \bibinfo{person}{Peter~I.
  Rockett}.} \bibinfo{year}{2005}\natexlab{}.
\newblock \showarticletitle{Evolving Optimal Feature Extraction Using
  Multi-objective Genetic Programming: A Methodology and Preliminary Study on
  Edge Detection}. In \bibinfo{booktitle}{\emph{Proc. Genetic Evol. Comput.
  Conf.}} \bibinfo{pages}{795--802}.
\newblock


\bibitem[Zhang and Rockett(2011)]%
        {427r}
\bibfield{author}{\bibinfo{person}{Yang Zhang} {and} \bibinfo{person}{Peter~I.
  Rockett}.} \bibinfo{year}{2011}\natexlab{}.
\newblock \showarticletitle{A Generic Optimising Feature Extraction Method
  Using Multiobjective Genetic Programming}.
\newblock \bibinfo{journal}{\emph{Appl. Soft Comput.}} \bibinfo{volume}{11},
  \bibinfo{number}{1} (\bibinfo{year}{2011}), \bibinfo{pages}{1087--1097}.
\newblock
\showISSN{1568-4946}


\bibitem[Zhang et~al\mbox{.}(2021b)]%
        {828r}
\bibfield{author}{\bibinfo{person}{Yidan Zhang}, \bibinfo{person}{Youheng
  Zhen}, \bibinfo{person}{Zhenan He}, {and} \bibinfo{person}{Gray~G. Yen}.}
  \bibinfo{year}{2021}\natexlab{b}.
\newblock \showarticletitle{Improvement of Efficiency in Evolutionary Pruning}.
  In \bibinfo{booktitle}{\emph{Proc. Int. Joint Conf. Neural Netw.}}
  \bibinfo{pages}{1--8}.
\newblock


\bibitem[Zhao et~al\mbox{.}(2006)]%
        {420r}
\bibfield{author}{\bibinfo{person}{Qijun Zhao}, \bibinfo{person}{David Zhang},
  {and} \bibinfo{person}{Hongtao Lu}.} \bibinfo{year}{2006}\natexlab{}.
\newblock \showarticletitle{A Direct Evolutionary Feature Extraction Algorithm
  for Classifying High Dimensional Data}. In \bibinfo{booktitle}{\emph{Proc.
  AAAI Conf. Artif. Intell.}}, Vol.~\bibinfo{volume}{1}.
  \bibinfo{pages}{561--566}.
\newblock


\bibitem[Zhao et~al\mbox{.}(2009)]%
        {422r}
\bibfield{author}{\bibinfo{person}{Qijun Zhao}, \bibinfo{person}{David~Dian
  Zhang}, \bibinfo{person}{Lei Zhang}, {and} \bibinfo{person}{Hongtao Lu}.}
  \bibinfo{year}{2009}\natexlab{}.
\newblock \showarticletitle{Evolutionary Discriminant Feature Extraction with
  Application to Face Recognition}.
\newblock \bibinfo{journal}{\emph{EURASIP J. Adv. Signal. Process.}}
  \bibinfo{volume}{2009} (\bibinfo{year}{2009}), \bibinfo{pages}{1--12}.
\newblock
\showISSN{1687-6180}


\bibitem[Zhao et~al\mbox{.}(2007)]%
        {421r}
\bibfield{author}{\bibinfo{person}{Tianwen Zhao}, \bibinfo{person}{Qijun Zhao},
  \bibinfo{person}{Hongtao Lu}, {and} \bibinfo{person}{David~Dian Zhang}.}
  \bibinfo{year}{2007}\natexlab{}.
\newblock \showarticletitle{Bagging Evolutionary Feature Extraction Algorithm
  for Classification}. In \bibinfo{booktitle}{\emph{Proc. Int. Conf. Neural
  Comput.}}, Vol.~\bibinfo{volume}{3}. \bibinfo{pages}{540--545}.
\newblock


\bibitem[Zhou et~al\mbox{.}(2021a)]%
        {031r}
\bibfield{author}{\bibinfo{person}{Xun Zhou}, \bibinfo{person}{A.~K. Qin},
  \bibinfo{person}{Maoguo Gong}, {and} \bibinfo{person}{Kay~Chen Tan}.}
  \bibinfo{year}{2021}\natexlab{a}.
\newblock \showarticletitle{A Survey on Evolutionary Construction of Deep
  Neural Networks}.
\newblock \bibinfo{journal}{\emph{IEEE Trans. Evol. Comput.}}
  \bibinfo{volume}{25}, \bibinfo{number}{5} (\bibinfo{year}{2021}),
  \bibinfo{pages}{894--912}.
\newblock


\bibitem[Zhou et~al\mbox{.}(2020)]%
        {772r}
\bibfield{author}{\bibinfo{person}{Yao Zhou}, \bibinfo{person}{Gary~G. Yen},
  {and} \bibinfo{person}{Zhang Yi}.} \bibinfo{year}{2020}\natexlab{}.
\newblock \showarticletitle{Evolutionary Compression of Deep Neural Networks
  for Biomedical Image Segmentation}.
\newblock \bibinfo{journal}{\emph{IEEE Trans. Neural Netw. Learn. Syst.}}
  \bibinfo{volume}{31}, \bibinfo{number}{8} (\bibinfo{year}{2020}),
  \bibinfo{pages}{2916--2929}.
\newblock


\bibitem[Zhou et~al\mbox{.}(2021b)]%
        {805r}
\bibfield{author}{\bibinfo{person}{Yao Zhou}, \bibinfo{person}{Gary~G. Yen},
  {and} \bibinfo{person}{Zhang Yi}.} \bibinfo{year}{2021}\natexlab{b}.
\newblock \showarticletitle{Evolutionary Shallowing Deep Neural Networks at
  Block Levels}.
\newblock \bibinfo{journal}{\emph{IEEE Trans. Neural Netw. Learn. Syst.}}
  (\bibinfo{year}{2021}).
\newblock
\urldef\tempurl%
\url{https://doi.org/10.1109/TNNLS.2021.3059529}
\showDOI{\tempurl}


\bibitem[Zhou et~al\mbox{.}(2021c)]%
        {827r}
\bibfield{author}{\bibinfo{person}{Yao Zhou}, \bibinfo{person}{Gary~G. Yen},
  {and} \bibinfo{person}{Zhang Yi}.} \bibinfo{year}{2021}\natexlab{c}.
\newblock \showarticletitle{A Knee-Guided Evolutionary Algorithm for
  Compressing Deep Neural Networks}.
\newblock \bibinfo{journal}{\emph{IEEE Trans. Cybern.}} \bibinfo{volume}{51},
  \bibinfo{number}{3} (\bibinfo{year}{2021}), \bibinfo{pages}{1626--1638}.
\newblock
\showISSN{2168-2267}


\bibitem[Zhu et~al\mbox{.}(2019)]%
        {545r}
\bibfield{author}{\bibinfo{person}{Hui Zhu}, \bibinfo{person}{Zhulin An},
  \bibinfo{person}{Chuanguang Yang}, \bibinfo{person}{Kaiqiang Xu}, {and}
  \bibinfo{person}{Yongjun Xu}.} \bibinfo{year}{2019}\natexlab{}.
\newblock \showarticletitle{{EENA}: Efficient Evolution of Neural
  Architecture}. In \bibinfo{booktitle}{\emph{Proc. IEEE Int. Conf. Comput.
  Vis.}} \bibinfo{pages}{1891--1899}.
\newblock


\bibitem[Zhu and Jin(2022)]%
        {612r}
\bibfield{author}{\bibinfo{person}{Hangyu Zhu} {and} \bibinfo{person}{Yaochu
  Jin}.} \bibinfo{year}{2022}\natexlab{}.
\newblock \showarticletitle{Real-Time Federated Evolutionary Neural
  Architecture Search}.
\newblock \bibinfo{journal}{\emph{IEEE Trans. Evol. Comput.}}
  \bibinfo{volume}{26}, \bibinfo{number}{2} (\bibinfo{year}{2022}),
  \bibinfo{pages}{364--378}.
\newblock


\bibitem[Zhu et~al\mbox{.}(2007)]%
        {780r}
\bibfield{author}{\bibinfo{person}{Zexuan Zhu}, \bibinfo{person}{Y. Ong}, {and}
  \bibinfo{person}{Manoranjan Dash}.} \bibinfo{year}{2007}\natexlab{}.
\newblock \showarticletitle{Markov Blanket-Embedded Genetic Algorithm for Gene
  Selection}.
\newblock \bibinfo{journal}{\emph{Pattern Recognit.}} \bibinfo{volume}{40},
  \bibinfo{number}{11} (\bibinfo{year}{2007}), \bibinfo{pages}{3236--3248}.
\newblock
\showISSN{0031-3203}


\bibitem[Zoph et~al\mbox{.}(2018)]%
        {027r}
\bibfield{author}{\bibinfo{person}{Barret Zoph}, \bibinfo{person}{Vijay
  Vasudevan}, \bibinfo{person}{Jonathon Shlens}, {and} \bibinfo{person}{Quoc~V
  Le}.} \bibinfo{year}{2018}\natexlab{}.
\newblock \showarticletitle{Learning Transferable Architectures for Scalable
  Image Recognition}. In \bibinfo{booktitle}{\emph{Proc. IEEE Conf. Comput.
  Vis. Pattern Recognit.}} \bibinfo{pages}{8697--8710}.
\newblock


\end{thebibliography}
	
\end{document}